\documentclass[10pt,dvipsnames]{article} % For LaTeX2e
\usepackage[preprint]{tmlr}
\usepackage{amsmath,amsfonts,bm}

\def\eqref#1{equation~\ref{#1}}
\def\1{\bm{1}}

\def\rvw{{\mathbf{w}}}

\DeclareMathAlphabet{\mathsfit}{\encodingdefault}{\sfdefault}{m}{sl}
\SetMathAlphabet{\mathsfit}{bold}{\encodingdefault}{\sfdefault}{bx}{n}

\def\sC{{\mathbb{C}}}

\newcommand{\E}{\mathbb{E}}
\newcommand{\Ls}{\mathcal{L}}
\newcommand{\R}{\mathbb{R}}

\usepackage{hyperref}
\usepackage{url}

\usepackage{booktabs}       % professional-quality tables
\usepackage{amsfonts}       % blackboard math symbols
\usepackage{nicefrac}       % compact symbols for 1/2, etc.
\usepackage{microtype}      % microtypography
\usepackage{graphicx}
\usepackage{multirow}
\usepackage{subcaption}
\usepackage{bm}
\usepackage[textsize=tiny, color=green!30]{todonotes}
\usepackage{amsmath}
\usepackage{amssymb}
\usepackage{cleveref}
\usepackage{wrapfig}
\usepackage{enumitem}
\usepackage{float}
\setlist[itemize]{noitemsep, topsep=0pt, leftmargin=*}
\usepackage{standalone}
\usepackage{comment}
\usepackage{tcolorbox}
\usepackage{mdframed}
\usepackage{listings}

\newcommand{\up}[1]{\textcolor{BrickRed}{${}^{\blacktriangle}$}}
\newcommand{\down}[1]{\textcolor{ForestGreen}{${}^{\blacktriangledown}$}}
\newcommand{\ctick}{\textcolor{PineGreen}{$\surd$}}

\usepackage{tikz}
\usetikzlibrary{
  positioning,
  fit,
  calc,
  shapes,
  arrows.meta,
  shapes.callouts,
  shapes.symbols,
  shapes.geometric,
  shapes.arrows,
  arrows,
  arrows.meta,
  quotes,
  shadows,
  decorations.pathreplacing
}

\definecolor{lightblue}{HTML}{dae8fc}
\definecolor{darkblue}{HTML}{6C8EBF}

\definecolor{lightorange}{HTML}{FFE6CC}
\definecolor{darkorange}{HTML}{D79B00}

\definecolor{lightred}{HTML}{F8CECC}
\definecolor{darkred}{HTML}{B85450}

\definecolor{lightpurple}{HTML}{e1d5e7}
\definecolor{darkpurple}{HTML}{9673a6}

\definecolor{lightgreen}{HTML}{d5e8d4}
\definecolor{darkgreen}{HTML}{82b366}

\definecolor{lightgray}{HTML}{F5F5F5}
\definecolor{darkgray}{HTML}{666666}
\definecolor{verylightgray}{HTML}{E6E6E6}

\definecolor{lightgreenblue}{HTML}{B0E3E6}
\definecolor{darkgreenblue}{HTML}{0E8088}

\definecolor{lightyellow}{HTML}{FFF2CC}
\definecolor{darkyellow}{HTML}{D6B656}

\usepackage{amsthm}
\newtheorem{remark}{Remark}
\usepackage[ruled, vlined, linesnumbered]{algorithm2e}

\newcommand{\bmark}{{{\sf \small TDBench}}\xspace}

\usepackage{bold-extra}
\newcommand{\ometh}{\texttt{\textbf{TDColER}}\xspace}

\newcommand{\ddsetcount}{226,890\xspace}
\newcommand{\dclfcount}{548,880\xspace}

\title{On Learning Representations for Tabular Data Distillation}

\author{\name Inwon Kang \email kangi@rpi.edu \\
  \addr Department of Computer Science\\
  Rensselear Polytechnic Institute
  \AND
  \name Parikshit Ram \email Parikshit.Ram@ibm.com \\
  \addr IBM Research \\
  Yorktown Heights, NY
  \AND
  \name Yi Zhou \email Yi.Zhou@ibm.com\\
  \addr IBM Research \\
  San Jose, CA 95120
  \AND
  \name Horst Samulowitz \email samulowitz@us.ibm.com\\
  \addr IBM Research \\
  Yorktown Heights, NY
  \AND
  \name Oshani Seneviratne \email senevo@rpi.edu\\
  \addr Department of Computer Science\\
  Renesselaer Polytechnic Institute}

\def\month{MM}  % Insert correct month for camera-ready version
\def\year{YYYY} % Insert correct year for camera-ready version
\def\openreview{\url{https://openreview.net/forum?id=XXXX}} % Insert correct link to OpenReview for camera-ready version

\begin{document}

\maketitle

\begin{abstract}
	Dataset distillation generates a small set of information-rich instances from a large dataset, resulting in reduced storage requirements, privacy or copyright risks, and computational costs for downstream modeling, though much of the research has focused on the image data modality. We study tabular data distillation, which brings in novel challenges such as the inherent feature heterogeneity and the common use of non-differentiable learning models (such as decision tree ensembles and nearest-neighbor predictors). To mitigate these challenges, we present \ometh, a tabular data distillation framework via column embeddings-based representation learning. To evaluate this framework, we also present a tabular data distillation benchmark, \bmark. Based on an elaborate evaluation on \bmark, resulting in \ddsetcount distilled datasets and \dclfcount models trained on them, we demonstrate that \ometh is able to boost the distilled data quality of off-the-shelf distillation schemes by 0.5-143\% across 7 different tabular learning models.
\end{abstract}

\section{Introduction}

\begin{comment}
{\tiny \begin{itemize}
	\item[\ctick] What is data distillation? Why is it important?
	\item[\ctick] Why is table distillation important? What is new/hard about it (heterogeneity, non-differentiable models)?
	\item[\ctick] Look into applications of distillation discussed in \citet{yu2023dataset}; list some also applicable to tables
	\item[\ctick] What do we study? Use of learned representations for table distillation
	\item[\ctick] What are our contributions? (i) well motivated table distillation pipeline making use of modern tools (column embeddings, gnn/trf architectures, downstream model agnostic), (ii) thorough eval, (iii) a benchmark for future evaluation
	\item (maybe) along with Fig 1, add fig of RND vs best kmeans in original vs best kmeans in ENC for XGB, LR, KNN (distill size vs regret)
\end{itemize}}
\end{comment}

Dataset distillation or dataset condensation is the process of creating a small set of extremely informative samples (usually synthetic) from a large dataset such that a model trained on this set will have predictive performance comparable to that of a model trained on the original large dataset~\citep{wangDatasetDistillation2020, yu2023dataset}. First, data distillation reduces data storage costs and can mitigate the privacy and copyright concerns involved in keeping around (and continuously utilizing) large amounts of raw data.
% \todo[inline,author=OS]{The above sentence probably needs a revision; "keeping" the raw data does not mitigate the privacy and copyright risks -- as the risk is only mitigated in the outputs of the data synthesis process.}
Furthermore, the reduction in the data size implies a lower computational cost of model training, especially when multiple models need to be trained on any given dataset. The above advantages of dataset distillation also facilitate various applications. Continual learning, where we need to learn new tasks while avoiding forgetting older tasks sequentially, often makes use of a ``replay buffer'' of old task data to be used while learning new tasks to mitigate forgetting of the older tasks~\citep{rolnickExperienceReplayContinual2019}.
% \todo[inline,author=OS]{Should we cite something for replay buffers? Maybe https://arxiv.org/abs/1811.11682}
Dataset distillation reduces the memory overhead of this replay buffer, allowing learning of a larger number of tasks without forgetting~\citep{tiwariGCRGradientCoreset2022,rosascoDistilledReplayOvercoming2022}.
\begin{wrapfigure}[13]{r}{0.55\textwidth}
	\vspace{-6px}
	\centering
	\includegraphics[width=0.55\textwidth]{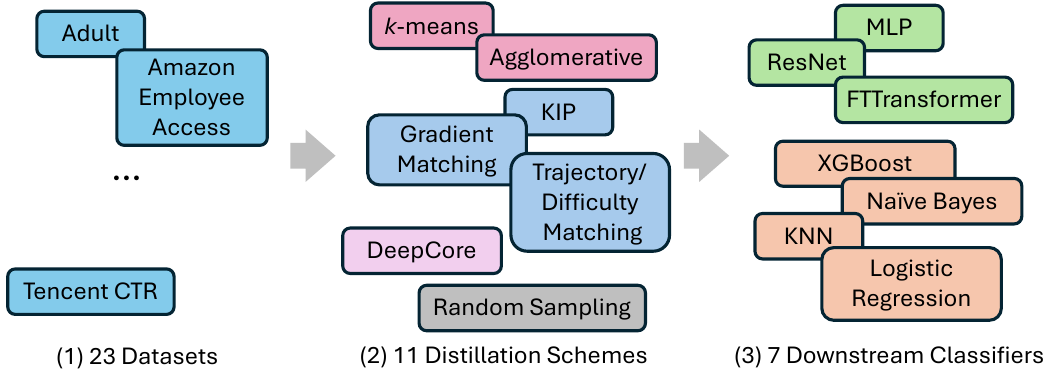}
	\vspace{-12px}
	\captionof{figure}{
		{\bf Overview of \bmark.}
		The benchmarking suite allows for flexible choice of datasets, distillation schemes, and downstream models that enables for modular evaluation of any new distillation method.
	}
	\label{fig:components_overview}
\end{wrapfigure}
In federated learning, we need to train a model using data spread across multiple clients without ever moving the data between clients and reducing the communication overhead. Dataset distillation generates compact yet private synthetic data from the client data that can be safely exchanged for communication-efficient model training~\citep{song2023federated, goetz2020federated, zhou2020distilled}.

While dataset distillation has been widely studied for image datasets~\citep{cui2022dc, yu2023dataset}, the equally important application to other data modalities is limited. The problem of tabular data distillation has received very little attention, though many real-world learning problems and applications involve tabular data~\citep{guoDeepFMFactorizationMachineBased2017,clementsSequentialDeepLearning2020,borisov2024deep}. Various image data distillation schemes have been proposed in the literature, but their application to tabular data is not straightforward. First, all image data distillation schemes rely on the choice of a {\em differentiable} ``backbone model.'' While differentiable neural network-based schemes are standard for images, a wide variety of non-differentiable models are used with tabular data, such as decision tree ensembles, nearest-neighbor models, and kernel machines. Second, almost all data distillation methods for images generate distilled data in the original pixel space. While pixels are homogeneous raw features of an image, the features in tabular data can be extremely heterogeneous, creating a mismatch between what the image data distillation methods are designed for and what we have as an inherent property of tabular data. Finally, it is standard to use vision-specific data augmentation schemes (such as rotation, reflection, cropping, and translation) to train the model on the distilled image data. Such standard augmentations are not available for tabular data, thus creating another discrepancy in the expected conditions for the problem.

\begin{figure}[t]
	\centering
	\begin{subfigure}{0.58\textwidth}
		\centering
		\includegraphics[width=\textwidth]{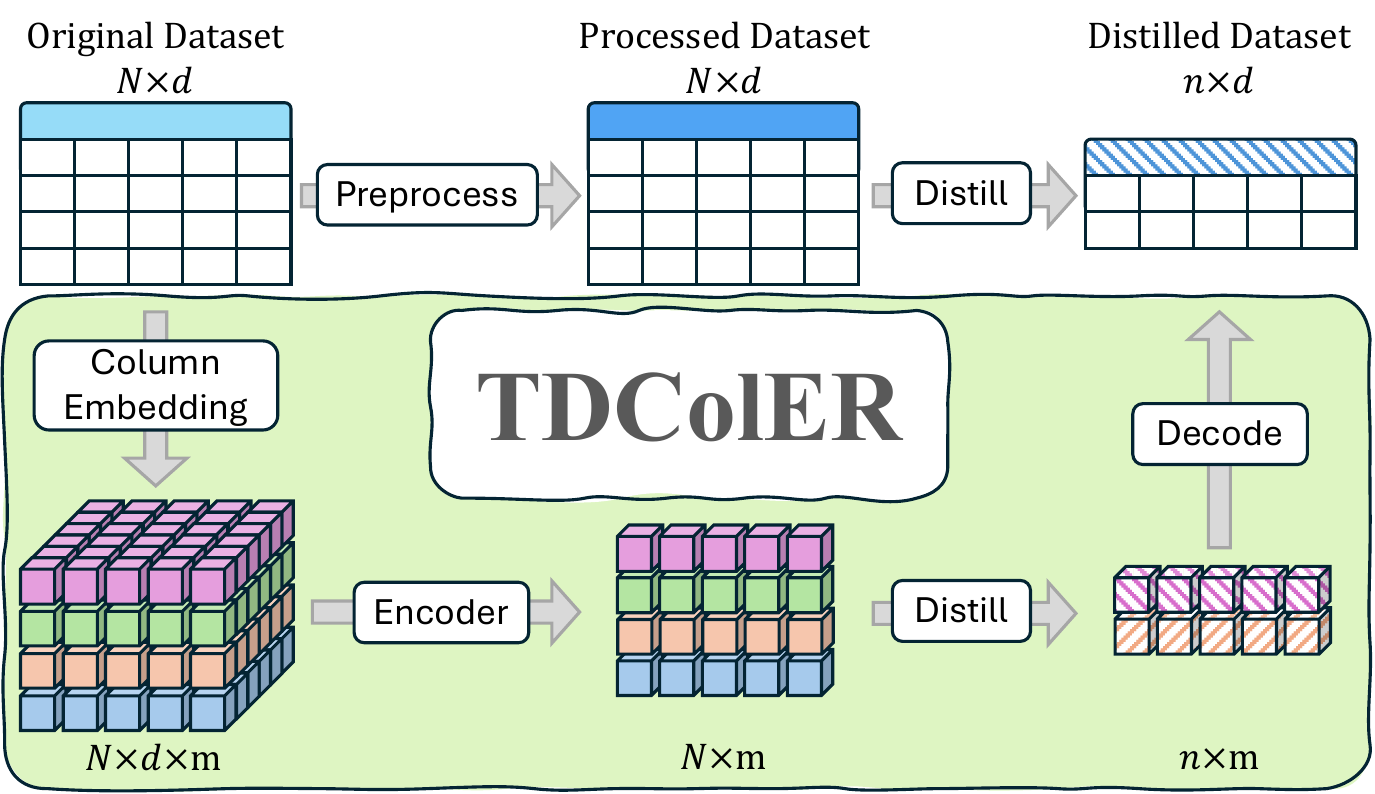}
		% \vspace{-12px}
		\caption{
			\textbf{Overview of \ometh.} The top describes a \textit{vanilla} distillation scheme that only uses standard preprocessing techniques before distillation.
			The highlighted box describes the proposed \ometh, which uses column embeddings after such preprocessing and encoder-decoder architectures to generate rich compact representations.
		}
		\label{fig:method-overview}
	\end{subfigure}
	~
	\begin{subfigure}{0.39\textwidth}
		\includegraphics[width=\textwidth]{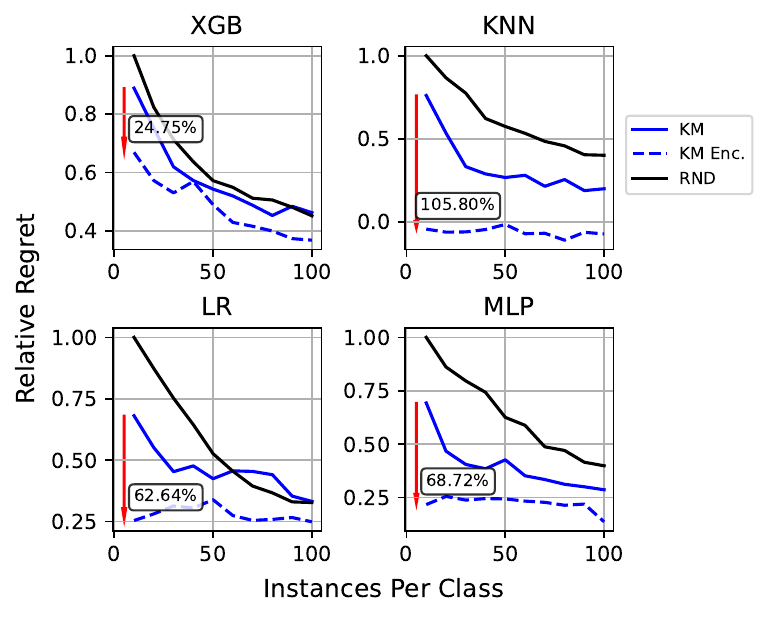}
		\vspace{-12px}
		\caption{
			Snapshot of downstream classifiers' performance increase when trained on data distilled $k$-means with and without \ometh.
			Throughout our experiments, we observe a performance increase from 0.5\% to as large as 143\%.
		}
		\label{fig:method-teaser}
	\end{subfigure}
	\vspace{-6px}
	\caption{Proposed approach -- \ometh: \textbf{T}abular \textbf{D}istillation Via \textbf{Col}umn \textbf{E}mbeddings based \textbf{R}epresentation Learning}
	\label{fig:method-overview-teaser}
	\vspace{-12px}
\end{figure}

\paragraph{Our contribution.}
In this paper, we study tabular dataset distillation and present a novel scheme to enhance the distilled data quality of multiple off-the-shelf data distillation schemes across various datasets, models, and distillation sizes. Specifically, we make the following contributions:
\begin{itemize}
	\item We propose {\bf T}abular {\bf D}istillation via {\bf Col}umn {\bf E}mbeddings based {\bf R}epresentation Learning or \ometh that can utilize modern neural-network architectures such as Transformers and graph neural networks to generate rich compact representations.
	      \ometh improves the quality of distilled data compared to existing distillation schemes.
	      \Cref{fig:method-overview} provides an overview of our proposed \ometh.
	      % \todo{add and point to figure visualizing tdcoler}
	      %	      \todo{name can also be TD-CERL instead of TDColER (OS: I like TDColER, as it's a bit more pronounceable and gives invokes "columns" and "ER", i.e., Entity-Relationships from DB literature.  We can use TD-CERL, but it may end up pronounced as "cereal" :))}
	\item We present \bmark, a {\bf T}abular {\bf D}istillation {\bf Bench}mark with 23 tabular datasets, 7 model classes, and 11 distillation schemes.
	      We present an overview of \bmark, an extensible and modular framework for measuring various aspects of data distillation on tabular data,  in \cref{fig:components_overview}.
	      % \todo{%
	      % - first part of \cref{fig:components_overview} should not have ``homogenizing'' anymore i think

	      % - also, can this figure now be shrunk horizontally to be half the textwidth?
	      % }
	      % \todo{%
	      %     - say more about benchmark
	      %     
	      %     - add and point to some version of \cref{fig:components_overview}; focus on things that are part of the benchmark, not part of tdcoler
	      %     
	      %     - fix caption of \cref{fig:components_overview}
	      % }
	\item With the elaborate evaluation of our proposed \ometh on \bmark, resulting in over \ddsetcount distilled datasets and \dclfcount model trainings, we show that, on aggregate across all datasets, \ometh improves upon direct application of off-the-shelf distillation method on tabular data by 0.5-143\% in terms of the distilled data quality across all models at the smallest distillation of 10 instances-per-class. \Cref{fig:method-teaser} presents a snapshot of our results.
	      % \todo{%
	      %  - fill in details
	      %
	      %  % - add fig with XGB, KNN, MLP, LR of regret vs IPC (instances-per-class) for kmeans + encoded space vs kmeans in original space vs RND
	      %
	      %  % - refer to this ``teaser results'' figure here
	      %
	      %  % - annotate fig with regret \%-improvement 
	      % }
	\item Based on our thorough evaluation, we present various insights regarding tabular dataset distillation, such as (i)~$k$-means clustering in the learned representations make for an extremely favorable distillation scheme, (ii)~ transformer-based tabular data representations obtain the highest distilled data quality on aggregate, while (iii)~graph neural network based tabular data representations perform slightly worse than transformers but are significantly more parameter efficient.
\end{itemize}

\subsection{Related Work}

Dataset distillation was introduced by \citet{wangDatasetDistillation2020} as a bilevel optimization problem~\citep{feng2024embarrassingly} and has been widely studied in the context of image data distillation.
Most methods can be categorized into approaches that match the original data by (i) backbone model performance, (ii) backbone model parameters, or (iii) backbone representation distributions~\citep{yu2023dataset}.
\citet{wangDatasetDistillation2020} minimized performance differences between the original and distilled data, while \citet{nguyenDatasetMetaLearningKernel2021} introduced kernel-induced points (KIP) using kernel ridge regression with a neural tangent kernel~\citep{jacotNeuralTangentKernel2018}.
Alternatively, methods have focused on parameter or gradient matching~\citep{zhao2021dataset,lee2022dataset,jiang2023delving,cazenavetteDatasetDistillationMatching2022b}.
Gradient matching~\citep{zhao2021dataset} aligns model gradients between original and synthetic data, while trajectory matching~\citep{cazenavetteDatasetDistillationMatching2022b} minimizes discrepancies between entire training trajectories.
Other approaches include distribution matching~\citep{zhao2023dataset}, which aligns per-class means, and cross-layer feature embedding matching~\citep{wangCAFE2022}.
However, the abovementioned methods rely on differentiable backbones, limiting cross-architecture generalization~\citep{cui2022dc,nguyenDatasetMetaLearningKernel2021}.
As a result, research has focused primarily on images, leaving \textit{tabular data distillation} largely unexplored~\citep{medvedev2021new}.
We address this gap by proposing a more general distillation framework.

Dataset distillation aligns with coreset selection~\citep{feldman2020core}, which aims to reduce data size, typically selecting real data instances (potentially risking privacy).
In contrast, distillation generates synthetic data beyond the real data manifold. Notably, coreset selection is a subset of dataset distillation, where the synthetic data lies on the real data manifold.
Generative modeling~\citep{goodfellow2020generative, kingma2013auto} is another related area, usually focused on generating realistic data.
In dataset distillation, the goal is to generate informative rather than realistic samples.
Recently, \citet{cazenavette2023generalizing} demonstrated how generative modeling can be used to seed the dataset distillation process, arguing that distillation methods should be applied to a latent representation instead of the pixel space.
This is aligned with our proposal, in which we demonstrate that distillation in the latent space is critical to obtaining meaningful distilled data quality with tabular datasets.
However, the proposed Generative Latent Distillation(GLaD) scheme is very focused on generative vision models, requiring a careful choice of the latent representation from within the model for trade-off in \textit{realistic} distilled data or \textit{expressivity}, thus limiting cross-architecture generalization.
% , trading off between generating realistic distilled data with expressivity (and thus limited cross-architecture generalization) and more flexible latent representations from earlier layers of a vision model.
% However, the proposed Generative Latent Distillation(GLaD) scheme is very focused on generative vision models. 
% It requires a careful choice of the latent representation from within the model, trading off between generating realistic distilled data with expressivity (and thus limited cross-architecture generalization) and more flexible latent representations from earlier layers of a vision model.

\citet{cui2022dc} benchmarked several distillation methods and found trajectory matching~\citep{cazenavetteDatasetDistillationMatching2022b} to be most effective, followed by KIP~\citep{nguyenDatasetMetaLearningKernel2021}.
Coreset methods, like $k$-means clustering, also outperformed many model-based distillation techniques, which we corroborate.
We focus on GM and KIP due to the high computational overhead of trajectory matching and omit data augmentation due to its limited applicability to tabular data.
As noted before, data augmentation is not standard with tabular data, and we do not consider it in our evaluation with \bmark.

\begin{comment}
The following literature then considered methods that leverage the parameter gradient of the target model instead of the parameters~\cite{cazenavetteDatasetDistillationMatching2022b,nguyenDatasetMetaLearningKernel2020,zhao2021dataset}.
~\citet{zhao2021dataset} propose matching the target model's gradient produced from the original and distilled dataset at each training iteration.
~\citet{nguyenDatasetMetaLearningKernel2020} introduce a kernel-inducing points (KIP) algorithm, which uses kernel ridge regression (KRR) with the neural tangent kernel (NTK)~\cite{jacotNeuralTangentKernel2018} of a wide neural network to approximate its training and train a synthetic (distilled) dataset.
Finally,~\citet{cui2022dc} perform a survey and experiment to benchmark six different data distillation methods, reporting that trajectory matching~\cite{cazenavetteDatasetDistillationMatching2022b} performs the best, with KIP~\cite{nguyenDatasetMetaLearningKernel2020} following the second.
The authors also consider \textit{k}-means algorithm as one of the baseline methods and note that it performs better than many other model-based distillation methods considered in the study, which is confirmed in our findings.
In this work, we consider gradient matching (GM) and KIP and in this work due to the fact that~\citet{cazenavetteDatasetDistillationMatching2022b} considers a heavily NN-reliant and \textit{model-centric} method while GM algorithm is not as heavily reliant on the target model and KIP treats the NN as a kernel machine.
\end{comment}

\section{Table Distillation}

Data distillation has been primarily studied in the context of images where each data point is composed of a homogeneous set of features -- pixels -- and the downstream models are neural networks.
The two main distinctions with tabular data distillation are: (i)~\textbf{Feature Heterogeneity}: Features in tabular data are usually heterogeneous and can have vastly different meanings, making it challenging to generate appropriate feature aggregations as usually done with neural networks. This is further exacerbated by the common presence of missing values.
\begin{wrapfigure}[9]{l}{0.56\textwidth}
	\vspace{-12px}
	%\begin{tcolorbox}[colback=PineGreen!10]
	\begin{mdframed}[linecolor=black!0,backgroundcolor=Cerulean!15]
		\begin{algorithm}[H]
			\caption{\small Distill original data $S$ with $N$ samples given a {\em preprocessor} $P: \R^r \times \sC^c \to \R^D$ and a {\em distiller} $F: \R^{N \times D} \times Y^N \to \R^{n \times D} \times Y^n$.}
			\label{alg:orig-distill}
			\SetAlgoCaptionSeparator{}% no separator, default colon
			\DontPrintSemicolon
			{\footnotesize
				%Given: Preprocessor $P: \R^r \times \sC^c \to \R^D$ \;
				%Given: Distiller $F: \R^{N \times D} \times Y^N \to \R^{n \times D} \times Y^n $ \;
				%{\bf Input}: $S = \{(x_i, y_i), i \in [N]: x_i \in \R^r \!\times\! \sC^c, y_i \in Y \}$ \;
				$\tilde{S} \gets \left\{(P(x), y)\, \forall (x, y) \in S \right\} $ \tcp*{Preprocess}
				$R \gets F(\tilde{S})$ \tcp*{Distill}
				\Return{$R$}
			}
		\end{algorithm}
		%\end{tcolorbox}
	\end{mdframed}
\end{wrapfigure}
(ii)~\textbf{Model Agnosticity}: For tabular data, the downstream model that will use the distilled data can be quite varied, with decision-tree-based models often being quite successful~\citep{grinsztajnWhyTreebasedModels2022}, while linear and nearest-neighbor models are used for interpretability.
Various increasing competitive neural-network-based models have also been developed for tabular data~\citep{borisov2024deep, gorishniyRevisitingDeepLearning2021, mcelfreshWhenNeuralNets2023, grinsztajnWhyTreebasedModels2022}.
%\todo{Cite tabular NN survey paper} and all the tabzilla etc benchmark papers here
%  Guo et al points to a knowledge distillation paper which seems irrelevant here}
However, in the most common cases, we cannot assume that the downstream model is differentiable and thus will be unable to perform a downstream model-specific distillation via the common bilevel formulation of the problem. The distillation has to be model-agnostic, which means that we have to retain as much of the information in the original data as possible since we do not know a priori what information the downstream model might leverage.

We will consider a classification dataset $S = \{(x_1, y_1), \ldots, (x_N, y_N)\}$ with $N$ samples, $r$ numerical features and $c$ categorical features, and $L$ labels, where each $x_i \in \R^r \times \sC^c$ and $y_i \in Y = \{1, \ldots, L\}$.
Following \cite{cui2022dc}, we only consider classification tasks in this work, but it should be noted that regression can be easily added into our framework.
% \todo{Might need a small footnote here to address why only classification. Maybe note that DC-bench only does classification. Regression relevant and can be addressed within our framework (but how?)}
Note that features may contain missing values.
After appropriate preprocessing steps to convert the categorical variables to numerical ones and imputing the missing values,
\footnote{For example, using data science tools such as \href{https://scikit-learn.org/stable/modules/generated/sklearn.preprocessing.OneHotEncoder.html}{\tt preprocessing.OneHotEncoder} and \href{https://scikit-learn.org/stable/modules/generated/sklearn.impute.SimpleImputer.html}{\tt impute.SimpleImputer} from the {\tt scikit-learn} machine learning toolkit.}
we can directly apply some existing distillation schemes such as KIP~\citep{nguyenDatasetMetaLearningKernel2021} or GM~\citep{zhao2021dataset}. This procedure is sketched in Algorithm~\ref{alg:orig-distill}.
% \todo{current placing of algo~\ref{alg:orig-distill} very unfortunate. need to fix once text is finalized}
% While this is an extremely simplistic procedure, we are explicitly discussing this as a baseline to highlight (in our empirical evaluations) the challenge of directly applying existing distillation schemes to tabular data, and the opportunity for improvements via tabular representation learning.
%\todo{need better way to say why we present this in so formal a form 
%
%{\color{blue} inwon: do we need to justify this? i feel like this is already OK?}
%}
%
\subsection{Representation Learning via Column Embedding}
\label{sec:repr-learning}
A key ingredient in the development of neural networks for tabular data is the use of column embeddings. First developed for categorical features, the idea is to learn an embedding for each of the categories in a categorical feature~\citep{guoEntityEmbeddingsCategorical2016}.
This embedding would replace the one-hot encoded numerical representation of the categories and be used in conjunction with the (appropriately scaled and imputed) numerical features in standard and custom feed-forward networks (FFNs)~\citep{borisov2024deep}.
Column embeddings for numerical data were developed to use more standard modern architectures such as graph neural networks (GNNs) and Transformers. As with categorical data, each numerical value in a numerical feature of the table would be converted into a learnable embedding. Thus, more precisely, a sample (row) in a table with $r$ numerical features and $c$ categorical features is now represented as a set of $(r+c)$ embeddings in $\R^m$ each of size $m$ (where $m$ is a user-specified hyperparameter), thus effectively as the $(m \times (r+c))$ matrix.
\footnote{While each feature can have column embeddings of different sizes, many neural network architectures require the column embedding size to match across all features.}

\paragraph{Encoder Architectures.}
Given the $\R^{m \times (r+c)}$ representation of a row (sample) using column embeddings, our goal is to learn a more compact yet faithful representation of a row.
One simple strategy is to concatenate all the $(r+c)$ column embeddings into a single vector in $\R^{m(r+c)}$ of size $m(r+c)$ and input it into an FFN which projects it down to a lower dimensionality (\cref{fig:enc:ffn}).
% \todo{point to \cref{fig:enc:ffn}}
%
However, one of our main motivations for using column embeddings is to leverage the capabilities of more modern architectures. For a given row, the $(r+c)$ column embeddings can be treated as initial token embeddings that are progressively updated through multiple Transformer blocks as described by~\citet{gorishniyRevisitingDeepLearning2021}.
% via the self-attention mechanism, layer normalization, and the nonlinear MLP transformation~\citep{gorishniyRevisitingDeepLearning2021}.
Using a dummy \texttt{[CLS]} token, the above process can create a $m$-dimensional representation of the row (\cref{fig:enc:tfblock}).
% \todo{add/point to \cref{fig:enc:tfblock};
% 
% 	might be good to have a version of \cref{fig:enc:tfblock} where we have the column embedding colors as in \cref{fig:enc:ffn,fig:enc:gnn}}
%
An alternate procedure is to represent a table as a bipartite graph between columns and rows (with column values and rows as vertices) and utilize the column embeddings as representations for the column vertices~\citep{wuOpenWorldFeatureExtrapolation2021}.
Then, the row embeddings are obtained by filling in representations for the row vertices via multiple rounds of message passing in a multi-layered GNN (\cref{fig:enc:gnn}).
% \todo{
% 	% cite FATE paper, 
% 	add/point to \cref{fig:enc:gnn}
% }
%
For our purposes, we consider all three architectures
-- FFN, Transformer and GNN -- as encoders that project the $\R^{m \times (r+c)}$ representation of row into an embedding in $\R^m$.
While categorical column embeddings are standard, there are multiple techniques for numerical column embeddings~\citep{gorishniyRevisitingDeepLearning2021,gorishniyEmbeddingsNumericalFeatures2022}. 
We discuss and ablate the effect these different schemes have in~\cref{apdx:sec:col_emb_effect}.
 % \todo{I think we are being preemptively defensive here since we have not even brought up binning yet, and we are already trying to defend it. I think we can say
 
 % ``While categorical column embeddings are standard, there are multiple techniques for numerical column embeddings (see discussion in Appendix xxx), and ablate the effect of numerical column embedding in \S xx.''}

% \todo{make footnote shorter and move longer discussion to appendix OR move discussion to main text}
\paragraph{Learning Objective.}
%
% While we have developed a compact representation for each sample (row) in a table using column embeddings, we still need an appropriate objective to optimize the representations.
% Since our goal is to retain as much information regarding the original data in the learned representation as possible, we try to reconstruct the original data from the learned representation as well as possible.
Our goal is to retain as much information regarding the original data in the learned representation as possible. 
{\em The need for high-fidelity learned representations is critical because we do not assume anything regarding the downstream model, which will be trained with the distilled data}.
Thus, we try to reconstruct the original data from the learned representation as well as possible.
Formally, given column embeddings $C: \R^r \times \sC^c \to \R^{m \times (r+c)}$, and an encoder $\phi: \R^{m \times (r+c)} \to \R^m$, we utilize a decoder $\psi: \R^m \to \R^r \times \sC^c$ to reconstruct the original data, and solve the following optimization problem:
\begin{equation} \label{eq:recon-loss-opt}
	\min_{C, \phi, \psi}
	\sum_{(x, y) \in S}
	\ell \left( x, \psi(\phi(C(x))) \right),
\end{equation}
where $\ell(\cdot, \cdot)$ is a reconstruction error (RE). Note that the above representation learning does not use the label information in the data $S$. This representation learning framework allows us to infuse class information in the representations while ensuring no loss of original information. Thus, after obtaining the column embeddings $C$, encoder $\phi$ and decoder $\psi$ by solving \cref{eq:recon-loss-opt}, we fine-tune the encoder by learning a classifier $f: \R^m \to Y$ on top of the learned representations while keeping the reconstruction loss low:
\begin{equation}\label{eq:sft-loss-opt}
	\min_{C, \phi, \psi, f} \sum_{(x, y) \in S} \ell(x, \psi(\phi(C(x))) + \alpha \Ls(y, f(\phi(C(x)))),
\end{equation}
where $\Ls(\cdot, \cdot)$ is the downstream learning loss function, and $\alpha > 0$ is a hyperparameter balancing the classification and reconstruction quality.
\Cref{apdx:subsec:ae-sft} discusses this procedure in more detail.
% \todo{do we really solve this \cref{eq:sft-loss-opt} to convergence, or just run it for a couple of epochs?
%
% 	\textcolor{blue}{inwon: we run for 50 epochs for each HPO instance, and almost all instances I observed show convergence. }
%
% 	pr: so why is it ``fine-tuning'' of encoder? if we are converging, why not directly solve \cref{eq:sft-loss-opt} and skip solving \cref{eq:recon-loss-opt}? Is the encoder learning rate small? What is the HPO for $\alpha$ optimizing for? reconstruction accuracy or prediction accuracy?
% }
%
%

\begin{wrapfigure}[13]{L}{0.57\textwidth}
	\vspace{-16px}
	\begin{mdframed}[linecolor=black!0,backgroundcolor=Cerulean!15]
		\begin{algorithm}[H]
			\caption{\small \ometh: Distill dataset $S$ with $N$ samples given {\em distiller} $F: \R^{N \times m} \times Y^N \to \R^{n \times m} \times Y^n$, and learnable {\em column embeddings} $C: \R^r \times \sC^c \to \R^{m \times (r+c)}$, {\em encoder} $\phi: \R^{m(r+c)} \to \R^m$, {\em decoder} $\psi: \R^m \to \R^r \times \sC^c$, {\em classifier} $f: \R^m \to Y$.}
			\label{alg:lrep-distill}
			\SetAlgoCaptionSeparator{}% no separator, default colon
			\DontPrintSemicolon
			{\footnotesize
				%Given: Recon. loss $\ell$, fine-tuning loss $\Ls$, reg. param $\gamma > 0$\;
				%Given: Learnable Column Emb $C: \R^r \times \sC^c \to \R^{m \times (r+c)}$ \;
				%Given: Learnable Enc $\phi: \R^{m \times (r+c)} \to \R^m$ \;
				%Given: Learnable Dec $\psi: \R^m \to \R^r \times \sC^c$ \;
				%Given: Learnable Class.: $f: \R^m \to Y$ \;
				%Given: Distiller $F: \R^{N \times m} \times Y^N \to \R^{n \times m} \times Y^n $ \;
				%Input: $S = \{(x_i, y_i), i \in [N]: x_i \in \R^r \times \sC^c, y_i \in Y \}$ \;
				%\tcp{Learn $C, \phi$ minimizing recon. loss}
				%$\hat{C}, \hat{\phi}, \hat{\psi} \gets \min\limits_{C, \phi, \psi} \sum\limits_{(x, y) \in S} \ell \left (x, \psi(\phi(C(x))) \right) $ \;
				$C, \phi, \psi \gets$ solve \cref{eq:recon-loss-opt}  \tcp*{minimize RE}
				%\tcp{Fine-tune $C, \phi$ with supervised loss}
				%$\hat{C}, \hat{\phi}, \hat{\psi}, \hat{f} \gets
				%\min\limits_{C, \phi, \psi, f} \sum\limits_{(x,y) \in S}
				%\ell  \left (x, \psi(\phi(C(x))) \right)
				%+ \gamma
				%\Ls  \left (y, f(\phi(C(x))) \right)
				%$ \;
				$C, \phi, \psi, f \gets$ solve \cref{eq:sft-loss-opt} \tcp*{fine-tune}
				$\tilde S \gets \{ (\phi(C(x)), y), (x, y) \in S \}$ \tcp*{Encode}
				$\tilde R \gets F(\tilde{S})$ \tcp*{Distill in latent space}
				$R \gets \{ (\psi(x), y), (x, y) \in \tilde{R} \}$ \tcp*{Decode}
				\Return{$R, \tilde{R}, C, \phi, \psi$}
			}
		\end{algorithm}
	\end{mdframed}
\end{wrapfigure}

\paragraph{Complete Distillation Pipeline.}
After the column embeddings $C$, encoder $\phi$ and decoder $\psi$ are learned (with \cref{eq:recon-loss-opt}) and fine-tuned (with \cref{eq:sft-loss-opt}), we convert the input features of the whole original dataset (with $N$ samples) into the learned representations in $\R^m$ using $C$ and $\phi$ and apply the aforementioned distillation schemes to this dataset ($N$ samples in $\R^m$) to get $n$ distilled samples in $\R^m$. At this point, we decode the distilled samples into the original representation using $\psi$. This whole pipeline is summarized in Algorithm~\ref{alg:lrep-distill}.
Note that the distillation with the learned representation in $\R^m$, and the availability of the decoder $\psi$, allows us to have two versions of the distilled data -- one in the learned representation ($\tilde{R}$ in Algorithm~\ref{alg:lrep-distill}, Line 4), and one in the original representation ($R$ in Algorithm~\ref{alg:lrep-distill}, Line 5). We can choose the appropriate distilled set based on the downstream application: If we require the distilled data to be obfuscated with no explicit correspondence to the original features, we can use $\tilde{R}$. In this setting, we are required to have the column embeddings $C$ and the encoder $\phi$ during inference with the downstream trained model to map the test points into the appropriate representation. If we require the distilled data and the model trained on it to be interpretable in terms of the original features, we should use the distilled set $R$ in the original representation. In this case, we do not need the column embeddings or the encoder during inference.

\begin{remark}
	Our contribution is a novel representation learning and distillation pipeline for model-agnostic tabular data distillation utilizing existing distillation schemes, column embeddings, and network architectures such as transformers and GNNs. In our thorough empirical evaluations, we will demonstrate the distilled data quality boost from this pipeline across multiple datasets and downstream models.
	%While we do not claim to present a novel data distillation scheme or a novel column embedding, we develop a novel pipeline for model-agnostic tabular data distillation utilizing existing distillation schemes, column embeddings and network architectures such as transformers and GNNs. In our thorough empirical evaluations, we will demonstrate the distilled data quality boost from this pipeline across multiple datasets and downstream models.
\end{remark}

\section{Evaluation Benchmark}
To thoroughly evaluate the various configurations of the proposed distillation pipeline, we establish a comprehensive benchmark suite with a varied set of datasets and downstream models, evaluating the pipeline at various levels of distillation sizes.
With 3 encoder architectures, 12 distillation schemes (including variants), 20+ datasets, 7 downstream models, 10 distillation sizes, 5 repetitions per distillation pipeline, and model training, we have generated over \ddsetcount distilled datasets and trained over \dclfcount individual downstream models~\footnote{
    The \bmark benchmarking suite (code provided in the supplement) can be extended to evaluate any new distillation method, tabular representation, and downstream model and compared against our current database of results (also provided in the supplement). The API requirements for each of these components in the distillation pipeline are described in \cref{apdx:sec:tdbench}, and the procedure to execute the benchmark suite can be found in \cref{apdx:sec:tdbench:workflow}, and the comparison using the current database of results can be found in \cref{apdx:sec:tdbench:reproduce}.
}%
% .\todo{Complete this section in the Appendix.}
% \todo{numbers in the above para dont match the ones in the contribution list in Intro}
%
\paragraph{Datasets.}
We consider 23 datasets from OpenML~\citep{OpenML2013} with the number of samples varying from 10,000 to over 110,000,
and number of features varying from 7 to 54.
Instead of investigating a few \textit{large} datasets, we choose to incorporate more datasets to generalize the findings across a wider range of datasets.
The datasets are chosen to be diverse in terms of the number of samples, features, and the type of features (numerical, categorical, or mixed).
There are 14/23 datasets with only numerical features, 2/23 with only categorical features, and 7/23 with both numerical and categorical features.
All these datasets correspond to binary classification problems.
Class imbalance is a common feature of tabular datasets~\citep{johnsonSurveyDeepLearning2019,thabtahDataImbalanceClassification2020}, and we focus on binary classification to carefully study the effect of class imbalance on the distilled data quality.
There are 9/23 almost perfectly balanced datasets and 10/23 datasets with a ratio of close to 1:2 between the smaller and larger classes, with the worst imbalance ratio smaller than 1:15.
Note that while we only consider binary classification datasets, the distillation pipelines are natively applicable to multi-class classification problems.
\paragraph{Distillation Methods.}
Given our aforementioned desiderata for model-agnosticity, we have the following existing distillation schemes available, which take as input the set $S$ of $N$ samples and output a set $R$ of $n \ll N$ distilled samples (further details regarding implementation of each distillation method is provided in~\cref{apdx:subsec:dd_params}):
% \todo{These details in the appendix needs to fleshed out better}
% \todo{Maybe add how each of these performed  on DC bench}
%
\begin{itemize}
	\item {\bf $k$-means Clustering (KM)} finds $n/L$ clusters for each of the $L$ classes to produce a total of $n$ distilled samples using Lloyd's $k$-means algorithm~\citep{lloyd1982least}.
	      % \todo{add citation or link to sklearn}
	      We consider two variations here by (i)~using the Euclidean center of each cluster to generate a synthetic sample or (ii)~choosing the closest \textit{real} point to the Euclidean center of each cluster. That is, $R$ comprises $n/L$ cluster centers (or closest real points) for each of the $L$ classes.
	\item {\bf Agglomerative Clustering (AG)}~\citep{mullnerModernHierarchicalAgglomerative2011} again generates $n/L$ clusters for each of the $L$ classes  is similar to $k$-means. We use the Ward linkage scheme with the Euclidean distance metric.
	      Similar to $k$-means, we generate (i)~synthetic samples by using the Euclidean center of a cluster or (ii)~real samples that are closest to the cluster centers.
	\item {\bf Kernel Induced Points (KIP)}~\citep{nguyenDatasetMetaLearningKernel2021}%
        % \footnote{
        %     The clustering-based distillation schemes and KIP are not explicitly tied to a specific model and thus satisfy our desiderata of model-agnosticity. In contrast, the Gradient Matching or GM distillation scheme heavily relies on the choice of the backbone model $M_\theta$ (as well as the learning algorithm parameters such as the learning rate), and there is no guarantee that the distilled samples $R$ would be useful for any other model.
        %     % Thus, this scheme is not model-agnostic. However, we consider GM to be representative of the model-specific distillation schemes for the sake of completeness of our evaluations.
        %     Thus, we consider GM to be representative of the non model-agnostic distillation schemes for the sake of completeness of our evaluations.
        %     For our table distillation, we choose $M_\theta$ to be a multi-layered perceptron with a single hidden layer. This will pose a mismatch when we evaluate the quality of the distilled data $R$ on standard tabular models such as decision tree ensembles and nearest-neighbor models, highlighting the need for model-agnosticity in tabular data distillation.
        % }%
	      ~uses the neural tangent kernel (NTK)~\citep{jacotNeuralTangentKernel2018} of a wide neural network and kernel ridge regression to produce a distilled set of samples.
	      Given the feature matrix $X \in \R^{N \times D}$ and the label vector $\mathbf y \in Y^N$, KIP learns the distilled feature matrix $\bar{X} \in \R^{n \times D}$ and label vector $\bar{\mathbf{y}} \in Y^n$ by solving the following  problem:
	      \begin{equation}
		      \min_{\bar{X}, \bar{\mathbf{y}}}
		      \Ls\left(
		      \mathbf{y}, K_{X \bar{X}}(K_{\bar{X} \bar{X}} + \lambda I)^{-1} \bar{\mathbf{y}}
		      \right),
		      \label{eq:kip_loss}
	      \end{equation}
	      where $\Ls$ is the downstream learning loss function, $K_{X \bar{X}} \in \R^{N \times n}$ is the NTK matrix between $X$ and $\bar{X}$, $K_{\bar{X} \bar{X}} \in \mathbb R^{n \times n}$ is the NTK matrix of $\bar{X}$ with itself, and $\lambda > 0$ is a regularization hyperparameter for the kernel ridge regression. Essentially, we are learning a set of synthetic samples such that the predictions made on the original dataset features using the distilled dataset via kernel ridge regression match the original labels.
	\item {\bf Gradient Matching (GM)}~\citep{zhao2021dataset} produces the distilled set $R$ for a given ``backbone model'' $M_\theta$ (parameterized by $\theta$) by directly optimizing for $R$ to induce model parameter gradients that are similar to the gradients obtained while training $M_\theta$ on the full dataset $S$.
	      Given a distance metric $D(\cdot, \cdot)$, and a distribution $P_{\theta_0}$ over the random model parameter initializations $\theta_0$, the distillation problem tries to minimize the distance between the model gradients computed on the full and distilled datasets over the $T$ steps of model learning as follows:
	      \begin{equation}
		      \min_{R} \E_{\theta_0 \sim P_{\theta_0}}\left[
			      \sum_{t=0}^{T-1} D\left (
			      \nabla_\theta \Ls (\theta_t; S),
			      \nabla_\theta \Ls (\theta_t; R)
			      \right) \right],
		      \label{eq:dc_loss}
	      \end{equation}
	      where $\Ls(\theta; S)$ is the loss of the model $M_\theta$ on the original full dataset $S$, $\Ls(\theta; R)$ is the loss of $M_\theta$ evaluated on $R$, and the model parameters $\theta_t$ are updated at $\theta_{t+1} \gets \theta_t - \eta_\theta \nabla_\theta \Ls(\theta_t; S)$ via gradient descent with a learning rate $\eta_\theta$ using the full original dataset.
\end{itemize}

We consider KIP and GM as representatives from previous data distillation literature that are \textit{model-agnostic} and \textit{model-centric}, respectively. \Cref{apdx:subsec:choices} further discusses our choice of distillation methods considered in this work. However, based on the weaker downstream performance of these methods in our initial experiments, we later included more recent methods from computer vision, such as trajectory-based matching~\cite{cazenavetteDatasetDistillationMatching2022b} and its evolved version difficultiy-based matching~\cite{guoLosslessDatasetDistillation2023}, and 4 representative NN-based coreset selection methods from DeepCore~\cite{guoDeepCoreComprehensiveLibrary2022}. All the above distillation schemes require the data to be preprocessed into a numerical form, and can be used in Algorithm~\ref{alg:orig-distill} to distill tables. But, as we will see, this is not a very useful scheme. Our evaluation of \ometh on \bmark will demonstrate how the performance of these distillation schemes are boosted via representation learning.

To study the ability of the distillation pipeline to generate really small but useful distilled datasets, we consider extremely small distilled datasets with 10-100 instances per class (IPC), corresponding to a distillation fraction of the order of 0.1-1.0\% on the smallest datasets, and 0.01-0.1\% for the largest datasets.
This is comparable to the compression ratio of 0.02-1\% used in~\citet{cui2022dc} and~\citet{cazenavette2023generalizing}.

\paragraph{Downstream models.}
\label{sec:downstream_models}
We consider 7 downstream models to evaluate the distilled data quality. We consider the Nearest-Neighbor Classifier (KNeighbors), Logistic Regression (LR), Gaussian Naive Bayes (GNB), and the Multi-Layered Perceptron (MLP) from the {\tt scikit-learn} library~\citep{scikit-learn}. We also consider the popular XGBoost ensemble of gradient-boosted decision trees (XGB)~\citep{chenXGBoostScalableTree2016a}. We include two recent neural network models for tabular data, the ResNet and the FTTransformer models~\citep{gorishniyRevisitingDeepLearning2021}. Since our distillation pipeline is deliberately model-agnostic, we train these models on the distilled data using the default hyperparameters of the corresponding libraries.
We also consider a hyperparameter optimization (HPO) use case using the distilled datasets in our evaluations, which can be found in Appendix~\ref{sec:results:best_overall}.

\paragraph{Evaluation metric.}
To have a standardized way to quantify the quality of the distilled data across different models and datasets, we use the notion of \textit{relative regret} which compares the model's balanced accuracy score when trained on the full, distilled and randomly sampled data points.
% Precisely, the {\em relative regret} is defined as $\nicefrac{(A_F - A)}{(A_F - A_{R_{10}})}$, where $A_F$ is the balanced accuracy of the model trained on the full training set (the upper watermark), $A_{R_{10}}$ is the balanced accuracy on the same test set when trained on 10 random samples per class averaged over 5 random repetitions (the lower watermark), and $A$ is the balanced accuracy of the model when trained on the distilled dataset over random 5 repetitions.
Precisely, the {\em relative regret} is defined as $\nicefrac{(A_F - A)}{(A_F - A_{R_{10}})}$, where $A_F$ is the balanced accuracy of the model trained on the full training set, $A_{R_{10}}$ is the balanced accuracy on the same test set when trained on 10 random samples per class averaged over 5 random repetitions, and $A$ is the balanced accuracy of the model when trained on the distilled dataset over random 5 repetitions. A relative regret of 1 matches the performance of random sampling at IPC=10, and a relative regret of 0 matches the performance of the model trained on the full dataset (which is usually the gold standard) -- lower relative regret implies higher distilled data quality~\footnote{For all the downstream models, the aggregate (median across all datasets) relative regret of random samples at IPC=10 (smallest distillation size) is 1.0 by definition, while the aggregate relative regret of random samples at IPC=100 (largest distillation size) is around 0.5, indicating that the benchmark is challenging enough with significant room for improvement.}.

\section{Results Analysis}
\label{sec:results}

\begin{figure}[t]
	\vspace{-12px}
	\centering
	\includegraphics[width=0.85\linewidth]{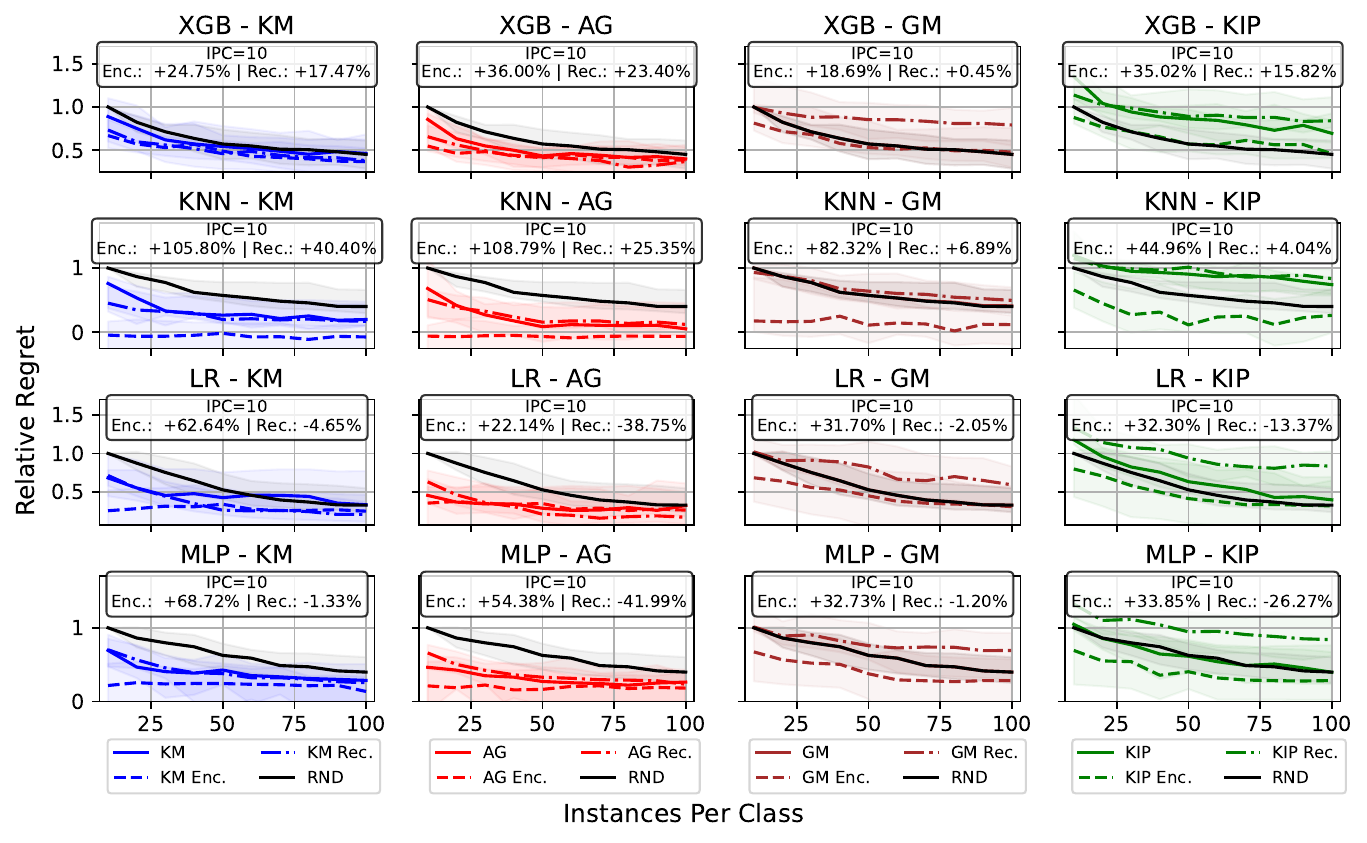}
	\vspace{-12px}
	\caption{
		Change in relative regret of downstream classifiers when trained on distilled data over IPC $\in [10, 100]$, aggregated over datasets and encoder architectures. Lower is better. Each column corresponds to a downstream classifier, and each row represents a representation scheme -- original, encoded (Enc.), and reconstructed (Rec.). Data distilled by clustering methods (AG, KM) in the encoded space show the best performance for all classifiers. In many cases, using the encoded representation as the final output yields a performance comparable to using the original representation. \Cref{fig:apdx:dm-per-clf-per-dm} shows a more detailed version of this plot that includes FTTransformer and ResNet.
	}
	\label{fig:results:clf-reg-per-dm-repr}
	\vspace{-12px}
\end{figure}

\begin{wrapfigure}[12]{r}{0.4\textwidth}
	\vspace{-14px}
	\centering
	\captionof{table}{
		Average rank and median relative regret of distillation pipelines that use the latent space of different encoder architectures evaluated at IPC=10, grouped over all datasets and classifiers.
	}
	\label{tab:results:encoder}
	\vspace{-8px}
	{\footnotesize
		\begin{tabular}{lcc}
			\toprule
			Encoder & Mean Rank          & Median R.R.        \\
			\midrule
			TF      & 4.1176             & 0.9439             \\
			FFN     & 4.3407             & 0.9746             \\
			GNN     & 4.2243             & 0.9695             \\
			TF*     & \textbf{2.3591}    & \textbf{0.6149}    \\
			FFN*    & 3.3652             & 0.8082             \\
			GNN*    & \underline{2.5931} & \underline{0.7135} \\
			\bottomrule
		\end{tabular}
	}
\end{wrapfigure}%
In this section, we present the analysis of the results obtained from our benchmarking experiments.
% We seek to answer the following two questions: 1). Which distillation method leads to best performance across datasets? 2). Which processing/transformations are the most helpful?
For the sake of brevity, we will use the following acronyms -- Instances Per Class: IPC, $k$-means: KM, agglomerative: AG, gradient matching: GM, kernel inducing points: KIP, feed-forward neural network: FFN, graph neural network: GNN, transformer: TF.
Additionally, the supervised-fine-tuned variant of the autoencoder will be marked with a *.
For example, the results of Algorithm~\ref{alg:lrep-distill} with a transformer architecture for $\phi$ as \textit{TF*}, whereas \textit{TF} denotes the version that skips line 2 of Algorithm~\ref{alg:lrep-distill} to highlight the importance of the supervised fine-tuning.

\paragraph{How beneficial are the learned representations for distillation?}
\label{sec:results:learned-reps-benefit}

As the first step of our analysis, we examine the performance difference between pipelines that use encoder's latent space and those that do not.
To fully understand the effect of our latent space projection step, we analyze our results from two angles: 1)~Is it better to distill in the latent or original space? 2)~If latent space is better, is it better to decode the data back to the original space or stay in the latent space?

\Cref{fig:results:clf-reg-per-dm-repr} shows the relative regret score of distillation methods under different data representation schemes.
We start by examining the downstream performance difference between pipelines that use the latent space to distill in vs. ones that do not (Algorithm~\ref{alg:lrep-distill} vs. Algorithm~\ref{alg:orig-distill}).
The results show %
% that three versions (Original, Encoded, Reconstructed) of clustering-based distillation methods lead to better downstream performance than random sampling and
that using the latent space is highly beneficial in most cases with lower IPC values.
This trend is most apparent in classifiers such as KNN (44.96-108.79\% improvement at IPC=10), Logistic Regression (22.14-62.64\% improvement) or MLP (32.73-68.72\% improvement), while XGBoost shows the least improvement from any of the distillation methods (15.82-36.00\% improvement).
$k$-means and agglomerative clustering also show a more apparent decrease in regret, while KIP and GM show noticeable improvements only when both the distillation and the final dataset are in the latent space.
\begin{wrapfigure}[18]{l}{0.45\textwidth}
	\vspace{-12px}
	\centering
	\includegraphics[width=0.43\textwidth]{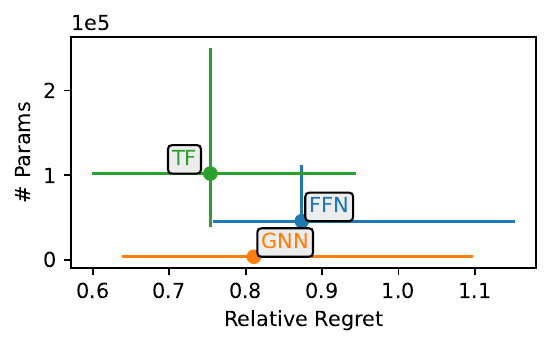}
	\vspace{-12px}
	\captionof{figure}{
		Scatterplot of encoder parameter size and downstream classifier regret at IPC=10 aggregated over datasets and classifiers.
		The dots represent the median values, and the error bars span the 25\% and 75\% percentile, respectively.
		Note that the encoder sizes for both SFT and base versions are the same for each dataset.
	}
	\label{fig:results:encoder-param-vs-reg}
\end{wrapfigure}
With this in mind, we examine the performance difference when training on the distilled data in the latent space or decoding to the original space before training the downstream classifier (using $\tilde{R}$ or $R$ from Algorithm~\ref{alg:lrep-distill}).
\Cref{fig:results:clf-reg-per-dm-repr} shows that training on the dataset in the latent space improves the downstream performance for all distillation pipelines -- in fact, it is the best performer for almost every instance over classifiers and distillation methods.
The change in performance is more apparent in KNN (40.92-65.40\%), Logistic Regression (33.75-67.29\%) and MLP (33.93-96.38\%), while XGBoost shows a more subtle change (7.28-19.20\%).
This leads us to conclude that \textbf{distillation methods benefit the most when both distilling \textit{and} downstream training in on the latent representations}.
It is also worth noting that decoding the distilled data from the latent space (Rec.) is also beneficial compared to random sampling in many cases.

\paragraph{How do different encoders compare?}

Having observed that using the latent space is beneficial, we now seek to identify which encoder architecture leads to the best performance.
\Cref{tab:results:encoder} shows the average rank of distillation pipelines that use the latent space of different encoder architectures.
Among the tested architectures and training objectives, the \textbf{transformer architecture with supervised fine-tuning} leads to the best downstream performance.
We find that \textbf{adding supervised fine-tuning improves the downstream performance of all encoders} in general.

% However, the downstream performance is not the sole goal of data distillation.
Another important aspect of data distillation is to improve downstream classifier efficiency providing a lightweight proxy.
Thus, it is important to examine the resources required in the distillation pipeline.
Specifically, one aspect of our distillation pipelines that can add an additional cost is the encoder.
In settings that require the data to be projected into latent space at inference time, the encoder can be considered part of the distilled data.
\Cref{fig:results:encoder-param-vs-reg} shows the parameter size of the different encoder architectures vs. the downstream classifier regret scores.
As noted before, the transformer architecture leads to the best downstream performance.
However, it is worth noting that \textit{GNN architecture has the smallest overall parameter size while providing the second-best performance.}
Further discussion on the parameter size analysis of each encoder architecture can be found in~\cref{apdx:sec:enc_param_discuss}.

\begin{table}[t]
	\centering
	\caption{
		Relative regret of pipelines that use different combinations of distill methods and encoders at IPC=10, aggregated over classifiers.
		The best value for each column is marked with \textbf{bold}, and the second best is marked with \underline{underline}.
		The best in each distillation method group is marked with \textit{italics}.
		On average, $k$-means with SFT transformer shows the best performance, but agglomerative clustering also shows comparable performance.
	}
	\label{tab:results:distill_method}
	\vspace{-6px}
	{\footnotesize\begin{tabular}{llcccccc}
	\toprule
	Distill Method       & Encoder & \multicolumn{6}{c}{Regret}                                                                                                                                                     \\
	                     &         & Min                           & Q1                          & Mean                      & Median                      & Q3                          & Max                      \\
	\midrule
	\multirow{3}{*}{KM}  & TF*     & \underline{\textit{-14.4491}} & \textit{\textbf{0.0733}}    & \textit{\textbf{-0.0464}} & \textit{\textbf{0.4056}}    & \textit{\underline{0.7379}} & \underline{1.1773}       \\
	                     & FFN*    & -11.9912                      & 0.2039                      & 0.1382                    & 0.6035                      & 0.8389                      & 1.5368                   \\
	                     & GNN*    & -12.1045                      & 0.0973                      & 0.1054                    & 0.5047                      & 0.7887                      & \textit{\textbf{1.0494}} \\
	\midrule
	\multirow{3}{*}{AG}  & TF*     & \textbf{\textit{-15.3965}}    & \textit{\underline{0.0810}} & \underline{\textit{0.0187}}& \textit{\underline{0.4135}} & \textit{\textbf{0.6507}}    & \textit{1.4982}          \\
	                     & FFN*    & -10.1288                      & 0.2483                      & 0.3695                    & 0.6230                      & 0.8823                      & 4.1191                   \\
	                     & GNN*    & -13.1881                      & 0.1397                      & 0.2245                    & 0.4793                      & 0.7595                      & 4.4801                   \\
	\midrule
	\multirow{3}{*}{KIP} & TF*     & -4.1619                       & \textit{0.5226}             & \textit{1.1124}           & \textit{0.9415}             & \textit{1.2966}             & 11.1034                  \\
	                     & FFN*    & \textit{-5.3973}              & 0.8053                      & 1.6363                    & 1.2502                      & 1.6434                      & 16.4137                  \\
	                     & GNN*    & -1.4649                       & 0.7403                      & 1.1957                    & 1.0136                      & 1.3329                      & \textit{10.5175}         \\
	\midrule
	\multirow{3}{*}{GM}  & TF*     & -3.8002                       & \textit{0.4105}             & \textit{0.7273}           & \textit{0.7952}             & 1.0564                      & \textit{4.9450}          \\
	                     & FFN*    & \textit{-4.3269}              & 0.5975                      & 1.2660                    & 0.9938                      & 1.3827                      & 16.5044                  \\
	                     & GNN*    & -1.4776                       & 0.4626                      & 0.8073                    & 0.8457                      & \textit{0.9779}             & 8.4566                   \\
	\bottomrule
\end{tabular}
}
\end{table}
\begin{figure}
	\centering
	\begin{subfigure}[t]{0.3\textwidth}
		\centering
		\includegraphics[width=\textwidth]{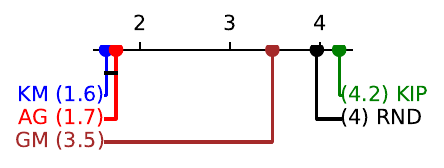}
		\vspace{-12px}
		\caption{IPC=10}
		\label{fig:results:distill_method_crit_n10}
	\end{subfigure}
	\begin{subfigure}[t]{0.3\textwidth}
		\centering
		\includegraphics[width=\textwidth]{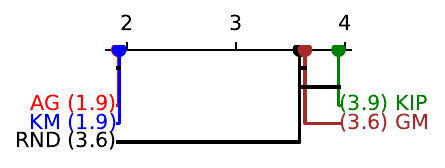}
		\vspace{-12px}
		\caption{IPC=50}
		\label{fig:results:distill_method_crit_n50}
	\end{subfigure}
	\begin{subfigure}[t]{0.3\textwidth}
		\centering
		\includegraphics[width=\textwidth]{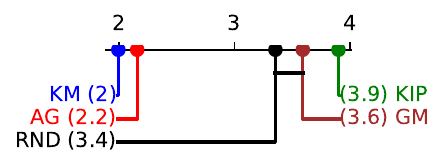}
		\vspace{-12px}
		\caption{IPC=100}
		\label{fig:results:distill_method_crit_n100}
	\end{subfigure}
	\vspace{-8px}
	\caption{
		Critical difference plot comparing ranks of distillation methods across datasets, encoders, and classifiers per IPC value.
		The x-axis denotes the average rank, and a black horizontal line connects groups of methods that are \textit{not significantly different} in the rank distribution.
		$k$-means and agglomerative are indistinguishable from each other in $\text{IPC} \in \{10, 50\}$, but $k$-means gains an edge in IPC=100.
	}
	\label{fig:results:distill_method_crit}
	% \vspace{-12px}
\end{figure}

\paragraph{Which distillation method leads to the best downstream performance?}
We now compare the most critical piece of the distillation pipeline -- the distillation method.%
We wish to understand which method leads to the best downstream performance across datasets, encoders, and classifier configurations.
To evaluate, we perform a Wilcoxon signed-rank test to identify groups that stand out from the rest, as shown in \cref{fig:results:distill_method_crit}.
The results show that \textbf{clustering-based methods ($k$-means, agglomerative) show the strongest performance across datasets and encoder configurations}, consistently placing in the top two ranks.
While both methods show similar performance, we find that $k$-means starts to outperform agglomerative as the IPC increases.

\begin{figure}
	\centering
	\begin{subfigure}[t]{0.49\textwidth}
		\centering
		\includegraphics[width=\textwidth]{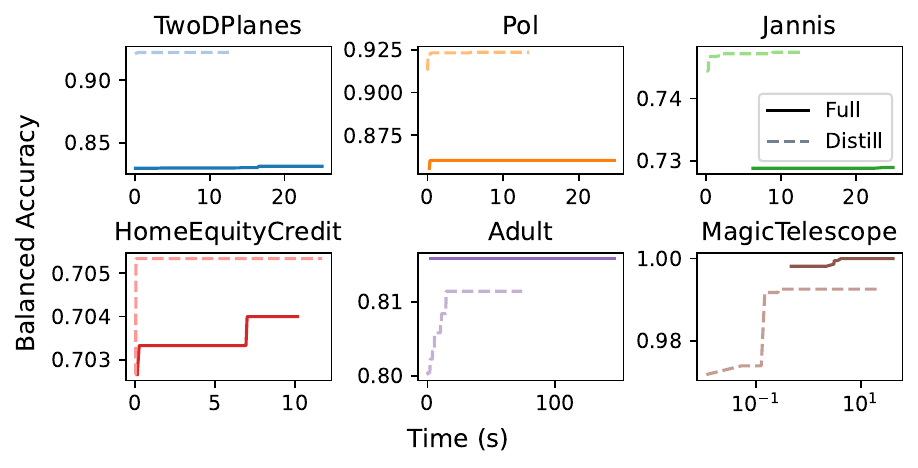}
		\vspace{-20px}
		\caption{Logistic Regression}
		\label{fig:results:hpo:lr}
	\end{subfigure}%
	\hfill%
	\begin{subfigure}[t]{0.49\textwidth}
		\centering
		\includegraphics[width=\textwidth]{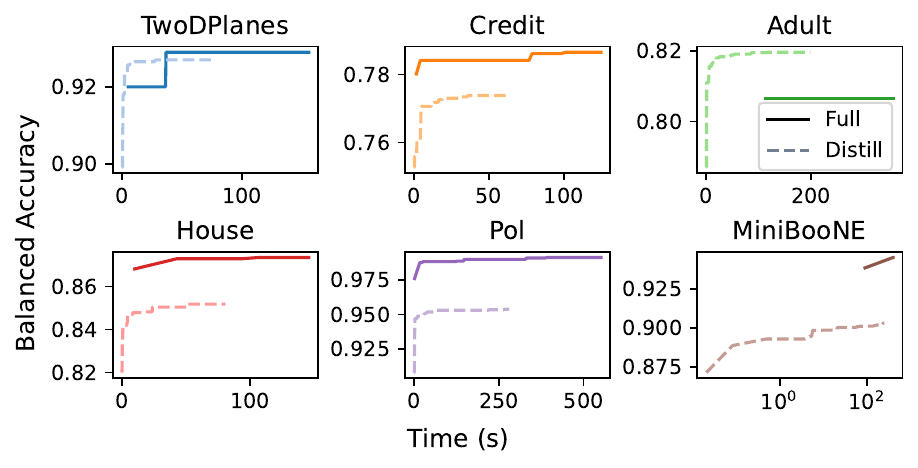}
		\vspace{-20px}
		\caption{XGBoost}
		\label{fig:results:hpo:xgb}
	\end{subfigure}%
	\vspace{-8px}
	\caption{
		Comparison of HPO validation performance vs. runtime when using full and distilled data.
		To better visualize the performance difference, we truncate the plot for the full data run at twice the runtime of the entire distilled run.
	}
	\label{fig:results:hpo}
	\vspace{-12px}
\end{figure}

\begin{wrapfigure}[13]{r}{0.4\textwidth}
	\centering
	\vspace{-14px}
	\captionof{table}{
		The best performers of each dataset are classifiers ranked by their appearance count at the top $3$ of each comparison at IPC=10.
		$k$-means stands out as the strongest performer in combination with a supervised-fine-tuned transformer encoder.
	}
	\label{tab:results:best_performer_by_count}
	\vspace{-6px}
	{\footnotesize
		\begin{tabular}{rlll}
			\toprule
			Count & Encoder & D.M. & Output \\
			\midrule
			67    & TF*     & KM   & Enc.   \\
			63    & GNN*    & KM   & Enc.   \\
			61    & GNN*    & AG   & Enc.   \\
			61    & TF*     & AG   & Enc.   \\
			42    & FFN*    & KM   & Enc.   \\
			\bottomrule
		\end{tabular}
	}
	% \vspace{-12px}
\end{wrapfigure}%
\paragraph{Which combination leads to best performance?}\label{sec:results:best_overall}
Our previous analysis has revealed that transformer encoders with SFT and clustering-based distillation methods perform best in their respective comparisons.
Now, we aim to identify which combination of encoder and distillation method leads to the best downstream performance.
We approach this question by examining the detailed statistics behind the combinations' performance and the top performers of each dataset, classifier, and $n$ combinations.
\Cref{tab:results:distill_method} shows detailed statistics about each distill method and encoder combination, while \cref{tab:results:best_performer_by_count} shows the count of the top 5 distillation pipelines that placed in the top 10 by performance in each comparison group.
% \todo{\cref{tab:results:distill_method} still need fix for best/second-best/best-in-group etc}
In line with our previous findings, the results show that \textbf{$k$-means clustering with supervised-fine-tuned transformer encoder leads to the best overall performance}.
All of the top performers are clustering-based methods, and all of them use the latent space, again confirming that \textbf{using the latent representation from the encoder greatly benefits distillation methods}.
In addition, the GNN encoder shows a comparative performance to that of the transformer encoder.
This is especially noteworthy, considering that GNN has the smallest parameter size among the encoder architectures.

We additionally run a smaller-scale HPO experiment to consider a use case for distilled data, as seen in ~\cref{fig:results:hpo}. Specifically, we consider a case where the validation and testing data is sampled from the original data, and the classifier is trained on either the full or distilled data. In general, we note that training on the distilled data gives comparable performance to training on the full data in a fraction of the time, consuming on average 21.84\% of the runtime and reaching 98.37\% of the performance.

\begin{figure}
	\centering
	\begin{subfigure}[t]{0.3\textwidth}
		\centering
		\includegraphics[width=\textwidth]{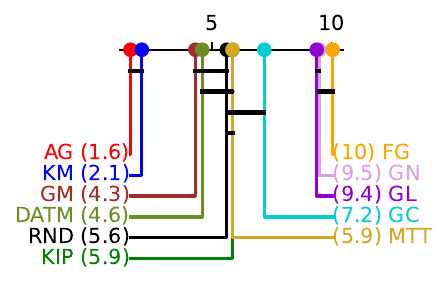}
		\vspace{-12px}
		\caption{IPC=10}
	\end{subfigure}
	\begin{subfigure}[t]{0.3\textwidth}
		\centering
		\includegraphics[width=\textwidth]{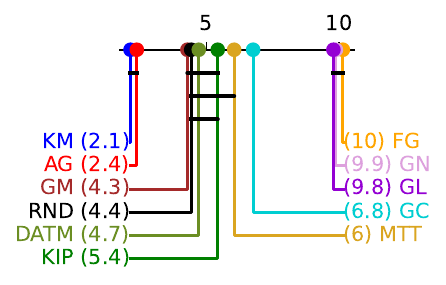}
		\vspace{-12px}
		\caption{IPC=50}
	\end{subfigure}
	\begin{subfigure}[t]{0.3\textwidth}
		\centering
		\includegraphics[width=\textwidth]{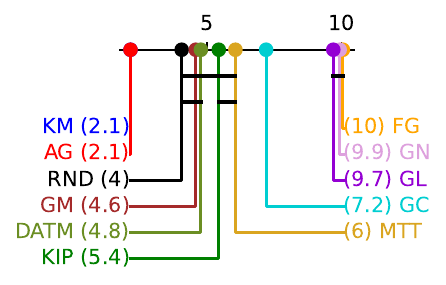}
		\vspace{-12px}
		\caption{IPC=100}
	\end{subfigure}
	\vspace{-8px}
	\caption{
		Critical difference plot comparing ranks of distillation methods across datasets per IPC value when applied with TF-SFT encoder for XGBoost classifier with additional baselines
		The x-axis denotes the average rank, and a black horizontal line connects groups of methods that are \textit{not significantly different} in the rank distribution.
		$k$-means and agglomerative are indistinguishable from each other in $\text{IPC} \in \{10, 50\}$, but $k$-means gains an edge in IPC=100.
		(FG: Forgetting, GN: GraNd, GL: Glister, GC: Graph Cut)
	}
	\label{fig:results:distill_method_crit_new}
	% \vspace{-12px}
\end{figure}

\paragraph{How do more recent data distillation methods compare?}
We conduct a further comparison of more recent distillation methods against the methods compared in~\cref{sec:results} to verify whether these methods will show superior performance. Specifically, we incorporate four representative NN-based coreset selection methods examined in DeepCore~\cite{guoDeepCoreComprehensiveLibrary2022} -- Forgetting~\cite{toneva2018empirical}, GraNd~\cite{paul2021deep}, Glister~\cite{killamsetty2021glister}, Graph Cut~\cite{iyer2013submodular}) and MTT~\cite{cazenavetteDatasetDistillationMatching2022b} and DATM~\cite{guoLosslessDatasetDistillation2023}. \Cref{fig:results:distill_method_crit_new} shows the updated critifcal different comparing the additional methods in the best-performing setting. The raw regret scores can be found in~\cref{apdx:tab:results:distill_method} of Appendix~\cref{apdx:sec:additional_distill_methods}. Consistent to our previous findings, we find that more recent distillation methods that rely on NNs do not fair well on non-differentiable downstream classifier (XGBoost), and that clustering methods still show dominance. It is also interesting to note that GM shows superior performance to MTT and DATM, suggesting that the latter two methods may actually be overfitting to the teacher network's architecture.

\paragraph{Does distillation preserve feature correlation?}

\begin{figure}
	\centering
	\includegraphics[width=0.8\textwidth]{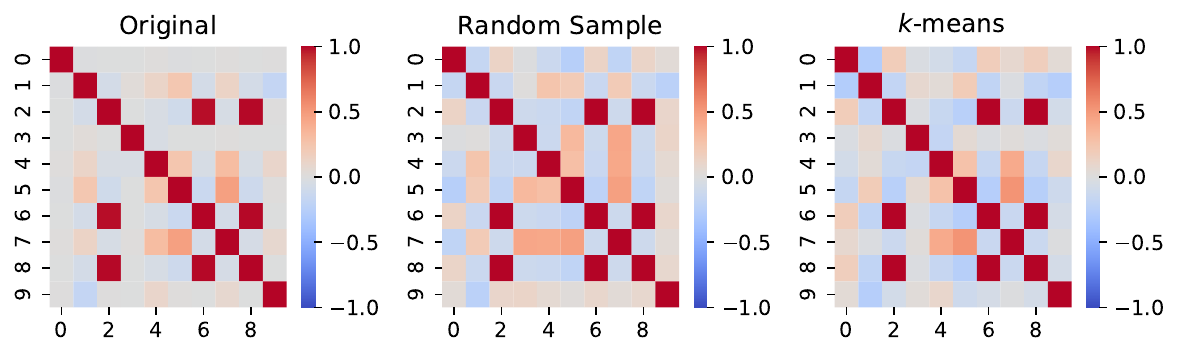}
	\caption{Feature correlation of the original, randomly sampled, and $k$-means distilled version of the Credit dataset@IPC=100.}
	\label{fig:results:feat_corr_credit}
	\vspace{-12px}
\end{figure}

We further investigate the presevation of feature correlation in the distilled data. \Cref{fig:results:feat_corr_credit} shows th feature correlation heatmaps for each version of the dataset. While the randomly sampled data also preserves most of the correlation, we observe that the dataset distilled with $k$-means is more similar (e.g. interaction between features 3 and 7 of Credit dataset) to the original dataset. We also observe this trend for other datasets, which can be seen in~\Cref{apdx:fig:results:feature_corr}.

\begin{wrapfigure}[16]{r}{0.4\textwidth}
	\vspace{-12px}
	\centering
	\includegraphics[width=0.4\textwidth]{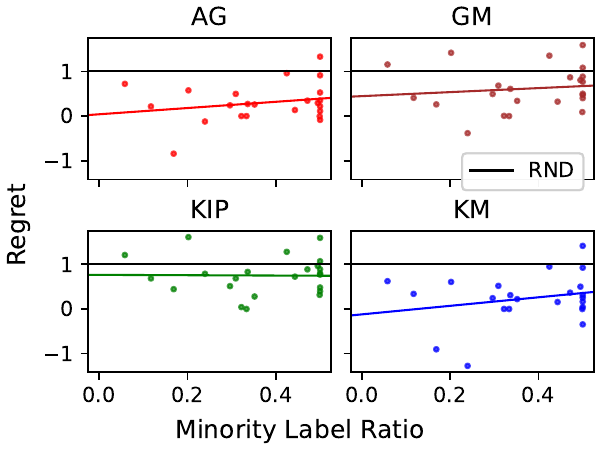}
	\vspace{-20px}
	\captionof{figure}{
		Average median relative regret of distillation methods aggregated over downstream classifiers and encoders at IPC=10 with a least-squares linear regression.
		Compared to KIP and GM, $k$-means and agglomerative show much stronger performance in imbalanced data.
	}
	\label{fig:results:class-imbal-regret}
	\vspace{-12px}
\end{wrapfigure}%
\paragraph{How does class imbalance affect performance?}
Finally, we examine the downstream performance of classifiers with respect to the label balance, or the imbalance, of the original dataset, shown \cref{fig:results:class-imbal-regret}.
Compared to other methods, including random sampling, clustering-based methods show impressive strength when distilling datasets with high label imbalance, highlighting their robustness under challenging data distributions. One possible explanation behind this phenomenon is that while NN-based distillation methods may prioritize the majority class due to the imbalance, the clustering methods are forced to place equal emphasis on all classes, preventing an overfitting on the majority class.

\section{Discussion}
\label{sec:conclusion}

% This work explored and benchmarked data distillation methods on tabular datasets, with a focus on the downstream classification performance of non-NN and NN ML classifiers trained on the distilled data.
% One main focus was to derive a distillation method that support a diversity of non-NN and NN ML classifiers as well as achieves competitive predictive performance on the distilled data.
This work introduced a tabular data distillation pipeline and evaluated it extensively leveraging various distillation methods, with a focus on supporting both non-NN and NN ML classifiers.
We introduce a novel framework, \ometh, that leverages latent representation of tabular data in distillation, and evaluate it thoroughly in our benchmark, \bmark, which included 23 datasets, 11 distillation algorithms, 3 autoencoder architectures, and 7 downstream classifiers, resulting in over \ddsetcount distilled datasets and \dclfcount downstream classifier instances.
Our results show that \ometh can induce superior performance in distillation methods on tabular data, improving the quality by 0.5-143\%.
We also show that $k$-means clustering and transformer autoencoder are a particularly strong combination for tabular data distillation.
We hope that this work will serve as a starting point for future research in tabular data distillation and plan to extend this benchmark further to incorporate new distillation pipelines.

\bibliography{references}
\bibliographystyle{tmlr}

\clearpage
\appendix
\section{Appendix}

\subsection{Datasets}\label{apdx:sec:datasets}

\begin{table}[t]
	\centering
	\caption{Dataset name and OpenML~\cite{OpenML2013} url}
	\label{apdx:tab:dataset_url}
	{\small
		\begin{tabular}[c]{ll}
			\toprule
			\textbf{Dataset Name}           & \textbf{Dataset URL}                 \\
			\midrule
			adult                           & \url{https://api.openml.org/d/1590}  \\
			Amazon\_employee\_access        & \url{https://api.openml.org/d/4135}  \\
			Bank\_marketing\_data\_set\_UCI & \url{https://api.openml.org/d/44234} \\
			credit                          & \url{https://api.openml.org/d/45027} \\
			default-of-credit-card-clients  & \url{https://api.openml.org/d/45020} \\
			Diabetes130US                   & \url{https://api.openml.org/d/45022} \\
			electrcity                      & \url{https://api.openml.org/d/151}   \\
			elevators                       & \url{https://api.openml.org/d/846}   \\
			higgs                           & \url{https://api.openml.org/d/23512} \\
			hcdr                            & \url{https://api.openml.org/d/45071} \\
			house\_16H                      & \url{https://api.openml.org/d/821}   \\
			jannis                          & \url{https://api.openml.org/d/45021} \\
			law-school-admission-bianry     & \url{https://api.openml.org/d/43890} \\
			MagicTelescope                  & \url{https://api.openml.org/d/1120}  \\
			Medical-Appointment-No-Shows    & \url{https://api.openml.org/d/43439} \\
			MiniBooNE                       & \url{https://api.openml.org/d/44088} \\
			numerai28.6                     & \url{https://api.openml.org/d/23517} \\
			nursery                         & \url{https://api.openml.org/d/959}   \\
			PhishingWebsites                & \url{https://api.openml.org/d/4534}  \\
			pol                             & \url{https://api.openml.org/d/722}   \\
			road-safety                     & \url{https://api.openml.org/d/44161} \\
			Click\_prediction\_small        & \url{https://api.openml.org/d/1220}  \\
			2dplanes                        & \url{https://api.openml.org/d/727}   \\
			\bottomrule
		\end{tabular}
	}
\end{table}

\begin{table}[t]
	\centering
	\caption{Metadata of each dataset seen in~\cref{apdx:tab:dataset_url}}
	\label{apdx:tab:dataset_info}
	{\small
		\begin{tabular}[c]{lllllll}
			\toprule
			\textbf{Dataset}                & \textbf{\# Instances} & \textbf{\# Features} & \textbf{\# Cont.} & \textbf{\# Cat.} & \textbf{\# Class 0} & \textbf{\# Class 1} \\
			\midrule
			2dplanes                        & 40,768                & 10                   & 10                & 0                & 20,420              & 20,348              \\
			Amazon\_employee\_access        & 32,769                & 9                    & 8                 & 1                & 1,897               & 30,872              \\
			Bank\_marketing\_data\_set\_UCI & 45,211                & 16                   & 7                 & 9                & 39,922              & 5,289               \\
			Click\_prediction\_small        & 39,948                & 11                   & 11                & 0                & 33,220              & 6,728               \\
			Diabetes130US                   & 71,090                & 7                    & 7                 & 0                & 35,545              & 35,545              \\
			MagicTelescope                  & 19,020                & 11                   & 11                & 0                & 12,332              & 6,688               \\
			Medical-Appointment-No-Shows    & 110,527               & 13                   & 10                & 3                & 88,208              & 22,319              \\
			MiniBooNE                       & 72,998                & 50                   & 50                & 0                & 36,499              & 36,499              \\
			PhishingWebsites                & 11,055                & 30                   & 0                 & 30               & 4,898               & 6,157               \\
			adult                           & 48,842                & 14                   & 6                 & 8                & 37,155              & 11,687              \\
			credit                          & 16,714                & 10                   & 10                & 0                & 8,357               & 8,357               \\
			default-of-credit-card-clients  & 13,272                & 20                   & 20                & 0                & 6,636               & 6,636               \\
			electrcity                      & 45,312                & 8                    & 7                 & 1                & 26,075              & 19,237              \\
			elevators                       & 16,599                & 18                   & 18                & 0                & 5,130               & 11,469              \\
			hcdr                            & 10,000                & 22                   & 22                & 0                & 5,000               & 5,000               \\
			higgs                           & 98,050                & 28                   & 28                & 0                & 46,223              & 51,827              \\
			house\_16H                      & 22,784                & 16                   & 16                & 0                & 6,744               & 16,040              \\
			jannis                          & 57,580                & 54                   & 54                & 0                & 28,790              & 28,790              \\
			law-school-admission-bianry     & 20,800                & 11                   & 6                 & 5                & 6,694               & 14,106              \\
			numerai28.6                     & 96,320                & 21                   & 21                & 0                & 47,662              & 48,658              \\
			nursery                         & 12,960                & 8                    & 0                 & 8                & 8,640               & 4,320               \\
			pol                             & 15,000                & 48                   & 48                & 0                & 5,041               & 9,959               \\
			road-safety                     & 111,762               & 32                   & 29                & 3                & 55,881              & 55,881              \\
			\bottomrule
		\end{tabular}
	}
\end{table}

\Cref{apdx:tab:dataset_url,apdx:tab:dataset_info} show the information about datasets used in our experiments along with their OpenML~\cite{OpenML2013} URLs.

\subsection{Hyperparameter optimization for encoders}
\label{apdx:subsec:enc_hpo}

\begin{table}
	\centering
	\captionof{table}{Hyperparameters tested for FFN encoder. }
	\label{apdx:tab:hpo:enc:ffn}
	{
		\small
		\begin{tabular}{lc}
			\toprule
			Hyperparameter          & Values                                                 \\
			\midrule
			\texttt{d\_hidden}      & $(100, 200)$                                           \\
			\texttt{n\_hidden}      & $[1, 4]$                                               \\
			\texttt{dropout}        & $(0, 0.2, 0.4)$                                        \\
			\texttt{d\_embedding}   & $(10, 20, 50, 100, 200)$                               \\
			\texttt{use\_embedding} & $($\texttt{True},\texttt{False}$)$                     \\
			\texttt{learning\_rate} & $10^{[-3,-1]}$                                         \\
			\texttt{weight\_decay}  & $10^{[-4,-1]}$                                         \\
			\texttt{lr\_scheduler}  & $($\texttt{None}, \texttt{Plateau}, \texttt{Cosine}$)$ \\
			\bottomrule
		\end{tabular}
	}
\end{table}%

\begin{table}
	\centering
	\captionof{table}{Hyperparameters tested for GNN encoder.}
	\label{apdx:tab:hpo:enc:gnn}
	{
		\small
		\begin{tabular}{lc}
			\hline
			\toprule
			Hyperparameter           & Values                                                 \\
			\midrule
			\texttt{graph\_layer}    & $($\texttt{graphsage}, \texttt{gcn}, \texttt{gat}$)$   \\
			\texttt{graph\_aggr}     & $($\texttt{mean}, \texttt{softmax}$)$                  \\
			\texttt{n\_graph}        & $[1, 15]$                                              \\
			\texttt{edge\_direction} & $($\texttt{bidirectional}, \texttt{multipass}$)$       \\
			\texttt{edge\_dropout}   & $(0, 0.2, 0.4)$                                        \\
			\texttt{learning\_rate}  & $10^{[-3,-1]}$                                         \\
			\texttt{weight\_decay}   & $10^{[-4,-1]}$                                         \\
			\texttt{lr\_scheduler}   & $($\texttt{None}, \texttt{Plateau}, \texttt{Cosine}$)$ \\
			\bottomrule
		\end{tabular}
	}
\end{table}%

\begin{table}
	\centering
	\captionof{table}{Hyperparameters tested for TF autoencoder.}
	\label{apdx:tab:hpo:enc:tf}
	{
		\small
		\begin{tabular}{lc}
			\hline
			\toprule
			Hyperparameter               & Values                                                 \\
			\midrule
			\texttt{n\_blocks}           & $[1, 10]$                                              \\
			\texttt{n\_attention\_heads} & $2^{[1, 4]}$                                           \\
			\texttt{d\_qkv}              & $2^{[0, 7]}$                                           \\
			\texttt{layer\_norm\_eps}    & $10^{[-5, -1]}$                                        \\
			\texttt{d\_mlp}              & $2^{[7, 11]}$                                          \\
			\texttt{d\_mlp\_hidden}      & $(100, 200)$                                           \\
			\texttt{n\_mlp\_hidden}      & $[1, 4]$                                               \\
			\texttt{dropout}             & $[0, 0.4]$                                             \\
			\texttt{learning\_rate}      & $10^{[-3,-1]}$                                         \\
			\texttt{weight\_decay}       & $10^{[-4,-1]}$                                         \\
			\texttt{lr\_scheduler}       & $($\texttt{None}, \texttt{Plateau}, \texttt{Cosine}$)$ \\
			\bottomrule
		\end{tabular}
	}
\end{table}

\begin{table}
	\centering
	\captionof{table}{Hyperparameters tested for decoders. The decoder architecture is kept the same for all encoders and optimized individually.}
	\label{apdx:tab:hpo:enc:dec}
	{
		\small
		\begin{tabular}{lc}
			\toprule
			Hyperparameter          & Values                                                 \\
			\midrule
			\texttt{d\_hidden}      & $(100, 200)$                                           \\
			\texttt{n\_hidden}      & $[1,4]$                                                \\
			\texttt{learning\_rate} & $10^{[-3,-1]}$                                         \\
			\texttt{weight\_decay}  & $10^{[-4,-1]}$                                         \\
			\texttt{lr\_scheduler}  & $($\texttt{None}, \texttt{Plateau}, \texttt{Cosine}$)$ \\
			\bottomrule
		\end{tabular}
	}
\end{table}

\begin{table}[b!]
	\centering
	\caption{Hyperparameters tested for classifier head in SFT.}
	\label{apdx:tab:hpo:enc:clf}
	{
		\small
		\begin{tabular}{lc}
			\hline
			\toprule
			Hyperparameter          & Values                                                 \\
			\midrule
			\texttt{d\_hidden}      & $\{100, 200\}$                                         \\
			\texttt{n\_hidden}      & $[1,3]$                                                \\
			\texttt{dropout}        & $\{0, 0.2, 0.4\}$                                      \\
			\texttt{alpha}          & $\{0.3, 0.5, 0.7\}$                                    \\
			\texttt{learning\_rate} & $10^{[-3,-1]}$                                         \\
			\texttt{weight\_decay}  & $10^{[-4,-1]}$                                         \\
			\texttt{lr\_scheduler}  & $($\texttt{None}, \texttt{Plateau}, \texttt{Cosine}$)$ \\
			\bottomrule
		\end{tabular}
	}
\end{table}

\Cref{apdx:tab:hpo:enc:ffn,apdx:tab:hpo:enc:gnn,apdx:tab:hpo:enc:tf,apdx:tab:hpo:enc:dec,apdx:tab:hpo:enc:clf} show the hyperparameters considered for different modules of the autoencoders.
We use $\{x,y,z\}$ to denote a set of variables and $[a,b]$ to denote an inclusive range of values.
% \todo{Might be useful to clarify that $(x,y,z)$ implies a set vs $[a,b]$ implies a range. Also if we want to denote sets, usually we use $\{x,y,z\}$ instead of $(x,y,z)$.}
We conduct HPO for each autoencoder + dataset pair using an implementation of hyperopt~\cite{bergstra2015hyperopt} from Ray Tune~\cite{liaw2018tune} with a maximum of 500 samples for each HPO run.
As noted in~\cref{sec:repr-learning}, we first train the vanilla autoencoders for each dataset using the encoder hyperparameters seen in~\cref{apdx:tab:hpo:enc:ffn,apdx:tab:hpo:enc:gnn,apdx:tab:hpo:enc:tf} and decoder parameters seen in~\cref{apdx:tab:hpo:enc:dec}. Once the vanilla autoencoders are trained, we then conduct an additional fine-tuning with a classifier head with hyperperameters seen in~\cref{apdx:tab:hpo:enc:clf} where $\alpha$ is used to balance the objective functions of the decoder and classifier heads.
% \todo{add citation for hyperopt}
% \todo{Might be useful to talk about how many hyperparameters we try for each pipeline and how we optimize autoencoder HPs first and then SFT HPs then. Is ``samples'' the number of HPs we tried.}

\subsection{Discussion on parameter size of autoencoders}
\label{apdx:sec:enc_param_discuss}

Here we expand on our parameter size of the encoder architectures of the autoencoders.
This is worth noting because if the distilled data is in the latent space, the encoder module is required to project any new data to the same space.
Thus, the encoder is considered to be a part of the distilled output.

We can characterize the parameter size of each encoder architecture given a $D$-dimensional binarized dataset with $c$ categorical features and $r$ continuous features that is projected to a $d$-dimensional latent space.

\noindent\textbf{FFN.}
We used an FFN architecture with an $M$-dimensional embedding layer followed by $H$ hidden layers that receive and output $W$-dimensional vectors. The parameter size of such an FFN is as follows:
\begin{equation}
	O(DM + (c+r)MW + HW^2 + Wd)
\end{equation}
The column embeddings are of size $O(DM)$, the input layer maps the concatenated $(c+r)M$-dimensional vector to hidden layer dimension $W$ with $(c+r)MW$ size. The hidden layers are of sizes $O(W^2)$ each for $H$  hidden layers. The output layer maps the $W$-dimensional hidden layer output to the desired $d$-dimensions.

\noindent\textbf{GNN.}
We use a GNN encoder with $H$ consecutive layers.
The dimension of the vectors passed between the graph layers are fixed to $d$, meaning that $M = d$.
Thus, each graph layer maintains a $d$ by $d$ matrix to handle a $d$-dimensional input vector and output a $d$-dimensional vector.
\begin{equation}
	O(Dd + Hd^2)
\end{equation}
The column embeddings are of size $O(Dd)$ since $M = d$. Each of the $H$ GNN layers is of size $O(d^2)$.

\noindent\textbf{Transformer.}
We consider an implementation of a transformer autoencoder inspired by the architecture of FT-Transformer~\cite{gorishniyRevisitingDeepLearning2021}.
The encoder has an $M$-dimensional embedding layer followed $H$ transformer blocks.
Each transformer block takes in a sequence of $M$-dimensional embeddings and oututs a single $d$-dimensional vector.
The block is composed of a multihead-attention module with $m$ heads and a FFN module to project the attention scores back to the input space.
% The implementation seen in~\citep{gorishniyRevisitingDeepLearning2021} uses
% Instead of using computing attention in the $$
We modify the architecture seen in~\citep{gorishniyRevisitingDeepLearning2021} by allowing the dimension of the attention head to be configurable
-- i.e. instead of using $M/m$ as the dimension of a single attention head, we allow the module to compute the attention in $d_{qkv}$.
This choice is motivated by the fact that our encoders were trained with a latent size of $16$, which may not be wide enough for the TF encoder.
We then project the resulting embedding in $d_{qkv}m$-dimension back to $M$-dimensionals with $W_o$.
Thus, each of $W_q$, $W_k$, $W_v$ and $W_o$ has $d_{qkv}mM$ parameters.
The MHA module is then followed by an FFN module which takes a $M$-dimensional vector and projects it back to $M$-dimensions with a $W$-dimensional hidden layer.
\begin{equation}
	O(H(4d_{qkv}Mm + 2MW))
\end{equation}

\subsection{Autoencoder implementation details}
\label{apdx:sec:ae_imp_detail}

\begin{figure}[t]
	\centering
	\begin{minipage}[b]{0.42\columnwidth}%
		\centering
		\definecolor{lightblue}{HTML}{dae8fc}
\definecolor{darkblue}{HTML}{6C8EBF}

\definecolor{lightorange}{HTML}{FFE6CC}
\definecolor{darkorange}{HTML}{D79B00}

\definecolor{lightred}{HTML}{F8CECC}
\definecolor{darkred}{HTML}{B85450}

\definecolor{lightgray}{HTML}{F5F5F5}
\definecolor{darkgray}{HTML}{666666}

\begin{tikzpicture}[%
		neuron/.style={
				circle,
				draw,
				minimum size=1cm
			},
		orange/.style={
				draw=darkorange,
				fill=lightorange,
			},
		blue/.style={
				draw=darkblue,
				fill=lightblue,
			},
		red/.style={
				draw=darkred,
				fill=lightred,
			},
		gray/.style={
				draw=darkgray,
				fill=lightgray,
			},
		embedding/.style={
				rectangle,
				draw,
				minimum height=1cm,
				minimum width=1cm,
				rounded corners,
			},
		shadowed/.style={
				drop shadow={
						shadow xshift=0.1cm,
						shadow yshift=-0.1cm,
						fill=black,
						opacity=0.2,
					}
			},
		brace/.style={
				decorate,
				decoration={
						brace,
						amplitude=0.4cm,
						mirror,
						% raise=0.6cm
					}
			},
		bigarrow/.style ={
				draw,
				single arrow,
				minimum height=0.7cm,
				minimum width=0.4cm,
				single arrow head extend=0.2cm
			}
	]

	\foreach \j in {1,...,3}{
			\foreach \i in {1,...,4}{
					\coordinate (e\j\i) at (-1-1*\i,1+-2*\j);
				}
		}

	\foreach \j in {4,...,7}{
			\foreach \i in {1,...,4}{
					\coordinate (e\j\i) at (-1-1*\i,-2*\j);
				}
		}

	\foreach \i in {1,...,4}{
			\coordinate (n1\i) at (0,-2*\i+2);
		}

	\foreach \i in {5,...,8}{
			\coordinate (n1\i) at (0,-2*\i+2);
		}

	\foreach \i in {1,...,6}{
			\coordinate (n2\i) at (2.5,-2*\i);
			% \coordinate (n3\i) at (5,-2*\i);
		}

	\foreach \i in {1,...,4}{
			\coordinate (n3\i) at (5.0,-2*\i-2);
		}

	\foreach \i in {1,...,4}{
			\coordinate (e8\i) at (6.0+1*\i,-7);
		}

	\foreach \i in {1,...,8}{
			\foreach \j in {1,...,6}{
					\draw[darkgray] (n1\i) -- (n2\j);
				}
		}

	\foreach \i in {1,...,6}{
			\foreach \j in {1,...,4}{
					\draw[darkgray] (n2\i) -- (n3\j);
				}
		}

	\foreach \j in {1,...,3}{
			\ifthenelse{
				\j=2
			}{
				\foreach \i in {4,...,1}{
						\node[embedding,orange,shadowed] at (e\j\i) {};
					}
			}{
				\foreach \i in {4,...,1}{
						\node[embedding,gray,shadowed] at (e\j\i) {};
					}
			}
		}

	\foreach \j in {4,...,7}{
			\ifthenelse{
				\j=4
			}{
				\foreach \i in {4,...,1}{
						\node[embedding,blue,shadowed] at (e\j\i) {};
					}
			}{
				\foreach \i in {4,...,1}{
						\node[embedding,gray,shadowed] at (e\j\i) {};
					}
			}
		}

	\foreach \i in {1,...,4}{
			\node[neuron,orange,shadowed] at (n1\i) {};
		}

	\foreach \i in {5,...,8}{
			\node[neuron,blue,shadowed] at (n1\i) {};
		}

	\foreach \i in {1,...,6}{
			\node[neuron,gray,shadowed] at (n2\i) {};
		}

	\foreach \i in {1,...,4}{
			\node[neuron,red,shadowed] at (n3\i) {};
		}

	\foreach \i in {1,...,4}{
			\node[embedding,red,shadowed] at (e8\i) {};
		}

	\draw[darkgray,brace] ($(e11) + (0.5, 0.7)$) -- ($(e14) + (-0.5, 0.7)$) node[midway,yshift=1cm]{\huge Feature Embeddings};
	\draw[darkgray,brace] ($(e14) + (-0.7, 0.5)$) -- ($(e34) + (-0.7, -0.5)$) node[midway,xshift=-2.0cm]{\huge Feature 1};
	\draw[darkgray,brace] ($(e44) + (-0.7, 0.5)$) -- ($(e74) + (-0.7, -0.5)$) node[midway,xshift=-2.0cm]{\huge Feature 2};
	\draw[darkgray,brace] ($(e84) + (0.5, 0.7)$) -- ($(e81) + (-0.5, 0.7)$) node[midway,yshift=1cm]{\huge Row Embedding};

	\node[bigarrow] at ($(e21) + (1, 0)$) {};
	\node[bigarrow] at ($(e41) + (1, 0)$) {};
	\node[bigarrow] at ($(e81) + (-1.1, 0)$) {};

  \node[rectangle] at ($(e71)+(0,-0.5)$) {};

\end{tikzpicture}
		\captionof{figure}{FFN encoder.}  \label{fig:enc:ffn}
	\end{minipage}
	\hfill
	\begin{minipage}[b]{0.56\columnwidth}%
		\centering
		\definecolor{lightblue}{HTML}{dae8fc}
\definecolor{darkblue}{HTML}{6c8ebf}

\definecolor{lightorange}{HTML}{ffe6cc}
\definecolor{darkorange}{HTML}{d79b00}

\definecolor{lightred}{HTML}{f8cecc}
\definecolor{darkred}{HTML}{b85450}

\definecolor{lightpurple}{HTML}{e1d5e7}
\definecolor{darkpurple}{HTML}{9673a6}

\definecolor{lightgreen}{HTML}{d5e8d4}
\definecolor{darkgreen}{HTML}{82b366}

\definecolor{lightgray}{HTML}{f5f5f5}
\definecolor{darkgray}{HTML}{666666}

\begin{tikzpicture}[%
		neuron/.style={
				circle,
				draw,
				minimum size=1cm
			},
		graphneuron/.style={
				ellipse,
				draw,
				minimum height=0.7cm,
				minimum width=1cm,
			},
		orange/.style={
				draw=darkorange,
				fill=lightorange,
			},
		blue/.style={
				draw=darkblue,
				fill=lightblue,
			},
		red/.style={
				draw=darkred,
				fill=lightred,
			},
		purple/.style={
				draw=darkpurple,
				fill=lightpurple,
			},
		green/.style={
				draw=darkgreen,
				fill=lightgreen,
			},
		gray/.style={
				draw=darkgray,
				fill=lightgray,
			},
		embedding/.style={
				rectangle,
				draw,
				minimum height=1cm,
				minimum width=1cm,
				rounded corners,
			},
		shadowed/.style={
				drop shadow={
						shadow xshift=0.1cm,
						shadow yshift=-0.1cm,
						fill=black,
						opacity=0.2,
					}
			},
		brace/.style={
				decorate,
				decoration={
						brace,
						amplitude=0.4cm,
						mirror,
						% raise=0.6cm
					}
			},
		bigarrow/.style ={
				draw,
				single arrow,
				minimum height=0.7cm,
				minimum width=0.4cm,
				single arrow head extend=0.2cm
			},
		doublearrow/.style={
				>=triangle 45,
				<->
			},
		graphlayer/.style={
				draw,
				trapezium,
				trapezium angle=80,
				minimum width=8cm,
				minimum height=4cm,
				trapezium stretches=true,
				rounded corners
			},
	]

	\foreach \j in {1,...,7}{
			\foreach \i in {1,...,4}{
					\coordinate (e\j\i) at (-1-1*\i,1+-2*\j);
				}
		}

	\foreach \j in {1,...,3}{
			\foreach \i in {1,...,4}{
					\coordinate (r\j\i) at (5-1*\i,1-4*\j);
				}
		}

	\foreach \j in {1,...,4}{
			\coordinate (g\j) at (8,-12+2*\j);

			\coordinate (n\j1) at (9,-11.9+2*\j);
			\coordinate (n\j2) at (11.8,-10.9+2*\j);
			\coordinate (n\j3) at (10.9,-11.8+2*\j);

			\coordinate (n\j4) at (9.9,-10.7+2*\j);
			\coordinate (n\j5) at (8.8,-13.1+2*\j);
			\coordinate (n\j6) at (13.5,-10.8+2*\j);
			\coordinate (n\j7) at (13.3,-11.9+2*\j);

			\coordinate (n\j8) at (10.3,-13.1+2*\j);
			\coordinate (n\j9) at (12.2,-12.9+2*\j);
			\coordinate (n\j10) at (14.1,-13.1+2*\j);
		}

	\foreach \j in {1,...,7}{
			\foreach \i in {4,...,1}{
					\ifthenelse{
						\j<4
					}{
						\node[embedding,orange,shadowed] at (e\j\i) {};
					}{
						\node[embedding,blue,shadowed] at (e\j\i) {};
					}
				}
		}

	\foreach \i in {4,...,1}{
			\node[embedding,red] at (r1\i) {};
		}
	\foreach \i in {4,...,1}{
			\node[embedding,green] at (r2\i) {};
		}
	\foreach \i in {4,...,1}{
			\node[embedding,purple] at (r3\i) {};
		}

	\draw[darkgray,brace] ($(e11) + (0.5, 0.7)$) -- ($(e14) + (-0.5, 0.7)$) node[midway,yshift=1cm]{\huge Feature Embeddings};
	\draw[darkgray,brace] ($(r11) + (0.5, 0.7)$) -- ($(r14) + (-0.5, 0.7)$) node[midway,yshift=1cm]{\huge Row Embeddings};
	\draw[darkgray,brace] ($(e14) + (-0.7, 0.5)$) -- ($(e34) + (-0.7, -0.5)$) node[midway,xshift=-2.0cm]{\huge Feature 1};
	\draw[darkgray,brace] ($(e44) + (-0.7, 0.5)$) -- ($(e74) + (-0.7, -0.5)$) node[midway,xshift=-2.0cm]{\huge Feature 2};
	\draw[darkgray,brace] ($(r11) + (0.7, -0.5)$) -- ($(r11) + (0.7, 0.5)$) node[midway,xshift=1.5cm]{\huge Row 1};
	\draw[darkgray,brace] ($(r21) + (0.7, -0.5)$) -- ($(r21) + (0.7, 0.5)$) node[midway,xshift=1.5cm]{\huge Row 2};
	\draw[darkgray,brace] ($(r31) + (0.7, -0.5)$) -- ($(r31) + (0.7, 0.5)$) node[midway,xshift=1.5cm]{\huge Row 3};

	\draw[doublearrow,darkgray] ($(e11)+(0.6,0)$) -- ($(r14)+(-0.6,0)$);
	\draw[doublearrow,darkgray] ($(e21)+(0.6,0)$) -- ($(r24)+(-0.6,0)$);
	\draw[doublearrow,darkgray] ($(e31)+(0.6,0)$) -- ($(r34)+(-0.6,0)$);

	\draw[doublearrow,darkgray] ($(e41)+(0.6,0)$) -- ($(r14)+(-0.6,0)$);
	\draw[doublearrow,darkgray] ($(e61)+(0.6,0)$) -- ($(r24)+(-0.6,0)$);
	\draw[doublearrow,darkgray] ($(e71)+(0.6,0)$) -- ($(r34)+(-0.6,0)$);

  \node[bigarrow,gray,shape border rotate=270,fill=white] at ($(g2) + (8, 1)$) {};

	\node[graphlayer,gray,opacity=0.8,anchor=west] at (g1) {};
	\draw[darkgray] (n11) -- (n18);
	\draw[darkgray] (n12) -- (n19);
	\draw[darkgray] (n13) -- (n110);
	\draw[darkgray] (n14) -- (n18);
	\draw[darkgray] (n16) -- (n19);
	\draw[darkgray] (n17) -- (n110);

	\node[graphneuron,orange,shadowed] at (n11) {};
	\node[graphneuron,orange,shadowed] at (n12) {};
	\node[graphneuron,orange,shadowed] at (n13) {};

	\node[graphneuron,blue,shadowed] at (n14) {};
	\node[graphneuron,blue,shadowed] at (n15) {};
	\node[graphneuron,blue,shadowed] at (n16) {};
	\node[graphneuron,blue,shadowed] at (n17) {};

	\node[graphneuron,red,fill=lightred] at (n18) {};
	\node[graphneuron,green,fill=lightgreen] at (n19) {};
	\node[graphneuron,purple,fill=lightpurple] at (n110) {};

	\node[graphlayer,gray,opacity=0.8,anchor=west] at (g2) {};
	\draw[darkgray] (n21) -- (n28);
	\draw[darkgray] (n22) -- (n29);
	\draw[darkgray] (n23) -- (n210);
	\draw[darkgray] (n24) -- (n28);
	\draw[darkgray] (n26) -- (n29);
	\draw[darkgray] (n27) -- (n210);

	\node[graphneuron,orange,shadowed] at (n21) {};
	\node[graphneuron,orange,shadowed] at (n22) {};
	\node[graphneuron,orange,shadowed] at (n23) {};

	\node[graphneuron,blue,shadowed] at (n24) {};
	\node[graphneuron,blue,shadowed] at (n25) {};
	\node[graphneuron,blue,shadowed] at (n26) {};
	\node[graphneuron,blue,shadowed] at (n27) {};

	\node[graphneuron,red,fill=lightred!80] at (n28) {};
	\node[graphneuron,green,fill=lightgreen!80] at (n29) {};
	\node[graphneuron,purple,fill=lightpurple!80] at (n210) {};

	\node[graphlayer,gray,opacity=0.8,anchor=west] at (g3) {};
	\draw[darkgray] (n31) -- (n38);
	\draw[darkgray] (n32) -- (n39);
	\draw[darkgray] (n33) -- (n310);
	\draw[darkgray] (n34) -- (n38);
	\draw[darkgray] (n36) -- (n39);
	\draw[darkgray] (n37) -- (n310);

	\node[graphneuron,orange,shadowed] at (n31) {};
	\node[graphneuron,orange,shadowed] at (n32) {};
	\node[graphneuron,orange,shadowed] at (n33) {};

	\node[graphneuron,blue,shadowed] at (n34) {};
	\node[graphneuron,blue,shadowed] at (n35) {};
	\node[graphneuron,blue,shadowed] at (n36) {};
	\node[graphneuron,blue,shadowed] at (n37) {};

	\node[graphneuron,red,fill=lightred!60] at (n38) {};
	\node[graphneuron,green,fill=lightgreen!60] at (n39) {};
	\node[graphneuron,purple,fill=lightpurple!60] at (n310) {};

	\node[graphlayer,gray,opacity=0.8,anchor=west] at (g4) {};
	\draw[darkgray] (n41) -- (n48);
	\draw[darkgray] (n42) -- (n49);
	\draw[darkgray] (n43) -- (n410);
	\draw[darkgray] (n44) -- (n48);
	\draw[darkgray] (n46) -- (n49);
	\draw[darkgray] (n47) -- (n410);

	\node[graphneuron,orange,shadowed] at (n41) {};
	\node[graphneuron,orange,shadowed] at (n42) {};
	\node[graphneuron,orange,shadowed] at (n43) {};

	\node[graphneuron,blue,shadowed] at (n44) {};
	\node[graphneuron,blue,shadowed] at (n45) {};
	\node[graphneuron,blue,shadowed] at (n46) {};
	\node[graphneuron,blue,shadowed] at (n47) {};

	\node[graphneuron,red,fill=lightred!40] at (n48) {};
	\node[graphneuron,green,fill=lightgreen!40] at (n49) {};
	\node[graphneuron,purple,fill=lightpurple!40] at (n410) {};

  \node[rectangle] at ($(e71)+(0,-0.5)$) {};
\end{tikzpicture}
		\captionof{figure}{GNN encoder.}  \label{fig:enc:gnn}
	\end{minipage}
	% \caption{Latent space encoders $\phi:\{0,1\}^D \to \mathbb R^d$}
	% \label{fig:enc}
\end{figure}

\begin{figure}[t]
	\centering
	\begin{minipage}[b]{0.49\columnwidth}%
		\centering
		\definecolor{lightblue}{HTML}{dae8fc}
\definecolor{darkblue}{HTML}{6c8ebf}

\definecolor{lightorange}{HTML}{ffe6cc}
\definecolor{darkorange}{HTML}{d79b00}

\definecolor{lightred}{HTML}{f8cecc}
\definecolor{darkred}{HTML}{b85450}

\definecolor{lightpurple}{HTML}{e1d5e7}
\definecolor{darkpurple}{HTML}{9673a6}

\definecolor{lightgreen}{HTML}{d5e8d4}
\definecolor{darkgreen}{HTML}{82b366}

\definecolor{lightgray}{HTML}{f5f5f5}
\definecolor{darkgray}{HTML}{666666}

\begin{tikzpicture}[%
		box/.style={
				draw,
				rectangle,
				rounded corners,
			},
		trap/.style={
				draw,
				trapezium,
				trapezium angle=120,
				trapezium stretches=true,
				rounded corners=0.1cm
			},
		down/.style={
				trapezium angle=120,
			},
		up/.style={
				trapezium angle=80,
			},
		small/.style={
				minimum width=2cm,
				minimum height=1cm,
			},
		medium/.style={
				minimum width=6cm,
				minimum height=1cm,
			},
		large/.style={
				minimum width=8cm,
				minimum height=4cm,
			},
		orange/.style={
				draw=darkorange,
				fill=lightorange,
			},
		blue/.style={
				draw=darkblue,
				fill=lightblue,
			},
		red/.style={
				draw=darkred,
				fill=lightred,
			},
		purple/.style={
				draw=darkpurple,
				fill=lightpurple,
			},
		green/.style={
				draw=darkgreen,
				fill=lightgreen,
			},
		gray/.style={
				draw=darkgray,
				fill=lightgray,
			},
		shadowed/.style={
				drop shadow={
						shadow xshift=0.1cm,
						shadow yshift=-0.1cm,
						fill=black,
						opacity=0.2,
					}
			},
		arrow/.style={
				>=triangle 45,
				->
			},
		doublearrow/.style={
				>=triangle 45,
				<->
			},
		brace/.style={
				decorate,
				decoration={
						brace,
						amplitude=0.1cm,
						mirror,
						% raise=0.6cm
					}
			},
	]

	\foreach \i in {0,...,3}{
			\coordinate (SDP \i) at (0-0.1*\i,0-0.1*\i);
			\coordinate (Wq \i) at (-2-0.1*\i,-2.0-0.1*\i);
			\coordinate (Wk \i) at (0-0.1*\i,-2.0-0.1*\i);
			\coordinate (Wv \i) at (2-0.1*\i,-2.0-0.1*\i);
			% \coordinate (Wo \i) at (0-0.1*\i,2-0.1*\i);
		}

	\coordinate (Wo) at ($(SDP 3) + (0.15,2.0)$);

	\foreach \i in {0,...,3}{
			\node[box,medium, orange,shadowed] at (SDP \i) {\large Scaled Dot Product};
			\node[trap,down,small,blue,shadowed] at (Wq \i) {\large $W_q$};
			\node[trap,down,small,green,shadowed] at (Wk \i) {\large $W_k$};
			\node[trap,down,small,red,shadowed] at (Wv \i) {\large $W_v$};
			% \node[trap,up,small,purple,] at (Wo \i) {\large $W_o$};
			\draw[arrow,gray] ($(Wq \i) + (0,0.5)$) -- ($(Wq \i) + (0, 1.4)$);
			\draw[arrow,gray] ($(Wk \i) + (0,0.5)$) -- ($(Wk \i) + (0, 1.4)$);
			\draw[arrow,gray] ($(Wv \i) + (0,0.5)$) -- ($(Wv \i) + (0, 1.4)$);
		}

	% \draw[brace] ($(SDP 0) + (0.00,0.65)$) -- ($(SDP 3) + (-0.15,0.45)$) ;

	% \draw[rounded corners=0.1cm]
	% ($(SDP 0) + (0.00,0.50)$)
	% -|($(SDP 0) + (0.00,0.55)$)
	% -- ($(SDP 3) + (-0.00,0.55)$)
	% |-($(SDP 3) + (0.00,0.50)$);

	\draw[rounded corners=0.05cm]
	($(SDP 3) + (-0.00,0.45)$)
	-- ($(SDP 3) + (-0.00,0.60)$)
	-- ($(SDP 0) + (0.00,0.60)$)
	-- ($(SDP 0) + (0.00,0.50)$)
node[midway,yshift=0.15cm,xshift=-0.8cm]{concat}
	;

  \draw[arrow,gray] ($(Wo) + (0, -1.1)$) -- ($(Wo) + (0, -0.6)$);

	\node[trap,up,medium,purple,minimum width=6.4cm] at (Wo) {\large $W_o$};

	% \foreach \i in {0,...,3}{
	% }
	% \foreach \i in {0,...,3}{
	% }

	% \node[box,small, green] at (0,-2.5) {};
	% \node[box,small, red] at (3,-2.5) {};
	% \node[box,small, blue] at (-1,-3) {};

\end{tikzpicture}
		\captionof{figure}{Modified MHA component.}  \label{fig:enc:mha}
	\end{minipage}
	\hfill
	\begin{minipage}[b]{0.49\columnwidth}%
		\centering
		\definecolor{lightblue}{HTML}{dae8fc}
\definecolor{darkblue}{HTML}{6c8ebf}

\definecolor{lightorange}{HTML}{ffe6cc}
\definecolor{darkorange}{HTML}{d79b00}

\definecolor{lightred}{HTML}{f8cecc}
\definecolor{darkred}{HTML}{b85450}

\definecolor{lightpurple}{HTML}{e1d5e7}
\definecolor{darkpurple}{HTML}{9673a6}

\definecolor{lightgreen}{HTML}{d5e8d4}
\definecolor{darkgreen}{HTML}{82b366}

\definecolor{lightgray}{HTML}{f5f5f5}
\definecolor{darkgray}{HTML}{666666}
\definecolor{verylightgray}{HTML}{E6E6E6}

\definecolor{lightgreenblue}{HTML}{B0E3E6}
\definecolor{darkgreenblue}{HTML}{0E8088}

\definecolor{lightyellow}{HTML}{FFF2CC}
\definecolor{darkyellow}{HTML}{D6B656}

\begin{tikzpicture}[%
		box/.style={
				draw,
				rectangle,
				rounded corners,
			},
		small/.style={
				minimum width=3cm,
				minimum height=0.4cm,
			},
		medium/.style={
				minimum width=4cm,
				minimum height=0.8cm,
			},
		long/.style={
				minimum width=2cm,
				minimum height=0.6cm,
				rounded corners=0.05cm,
			},
		container/.style={
				minimum height=10.5cm,
				minimum width=4.9cm,
        draw=darkgray,
        fill=verylightgray,
			},
		orange/.style={
				draw=darkorange,
				fill=lightorange,
			},
		blue/.style={
				draw=darkblue,
				fill=lightblue,
			},
		red/.style={
				draw=darkred,
				fill=lightred,
			},
		purple/.style={
				draw=darkpurple,
				fill=lightpurple,
			},
		yellow/.style={
				draw=darkyellow,
				fill=lightyellow,
			},
		greenblue/.style={
				draw=darkgreenblue,
				fill=lightgreenblue,
			},
		gray/.style={
				draw=darkgray,
				fill=lightgray,
			},
		shadowed/.style={
				drop shadow={
						shadow xshift=0.1cm,
						shadow yshift=-0.1cm,
						fill=black,
						opacity=0.2,
					}
			},
		arrow/.style={
				>=triangle 45,
				->,
				rounded corners=0.1cm,
				draw=darkgray,
			},
		line/.style={
				rounded corners=0.1cm,
				draw=darkgray,
			},
		bigarrow/.style ={
				draw,
				single arrow,
				minimum height=0.5cm,
				minimum width=0.15cm,
				single arrow head extend=0.15cm
			},
	]

	\node[] at(5.0, 3.6) {$T$};
	\node[box,long,greenblue,fill opacity=0.3] at (5.0,3.0) {};
	\node[] at (5.0,3.0) {[CLS]};
	\node[box,long,yellow,fill opacity=0.3] at (5.0,2.4) {};
	\node[box,long,yellow,fill opacity=0.3] at (5.0,1.8) {};
	\node[box,long,yellow,fill opacity=0.3] at (5.0,1.2) {};
	\node[box,long,yellow,fill opacity=0.3] at (5.0,0.6) {};
	\node[box,long,yellow,fill opacity=0.3] at (5.0,0.0) {};

	\node[bigarrow,gray,fill=white,shape border rotate=180] at (3.0, 0.0){};

	\node[] at(5.0, 5.9) {$T+1$};
	\node[box,long,greenblue] at (5.0,9.5) {[CLS]};
	\node[box,long,yellow] at (5.0,8.9) {};
	\node[box,long,yellow] at (5.0,8.3) {};
	\node[box,long,yellow] at (5.0,7.7) {};
	\node[box,long,yellow] at (5.0,7.1) {};
	\node[box,long,yellow] at (5.0,6.5) {};

	\node[bigarrow,gray,fill=white] at (3.0, 9.5){};

	\node[box,container,shadowed] at (-0.05,4.95) {};
	\node[box,container,shadowed] at (-0.15,4.85) {};
	\node[box,container,shadowed] at (-0.25,4.75) {};

	\node[box,orange,medium,minimum height=0.6cm] at (0,0) {\large Input};
	\draw[arrow] (0,0.4) -- (0,0.9);
	\node[box,gray,small] at (0,1.3) {\large LayerNorm};
	\draw[arrow] (0,1.7) -- (0,2.2);
	\draw[arrow] (0,1.85) -| (1.5,2.2);
	\draw[arrow] (0,1.85) -| (-1.5,2.2);
	\node[box,red,medium] at (0,2.7) {\large MHA};

	\draw[line] (0,0.55) -- (-2.5,0.55) |- (0,8.75);
	\node[] at (0.5,8.85) {Add};
	\draw[line] (-2.5,4.65) -- (0,4.65);
	\node[] at (0.5,4.75) {Add};
	% \node[] at () {\large Add};

	\draw[arrow] (0,3.2) -- (0,3.7);
	\node[box,gray,small] at (0,4.1) {\large Dropout};
	\draw[arrow] (0,4.5) -- (0,5.0);
	\node[box,gray,small] at (0,5.4) {\large LayerNorm};
	\draw[arrow] (0,5.8) -- (0,6.3);
	\node[box,red,medium] at (0,6.8) {\large FFN};
	\draw[arrow] (0,7.3) -- (0,7.8);
	\node[box,gray,small] at (0,8.2) {\large Dropout};
	\draw[arrow] (0,8.6) -- (0,9.1);
	\node[box,blue,medium,minimum height=0.6cm] at (0,9.5) {\large Output};

	\node[rectangle] at (0,-0.5) {};
\end{tikzpicture}
		\captionof{figure}{TF encoder block.}  \label{fig:enc:tfblock}
	\end{minipage}
	% \caption{Latent space encoders $\phi:\{0,1\}^D \to \mathbb R^d$}
	% \label{fig:enc}
\end{figure}

\subsubsection{Optimization function}

For the decoder $\psi:\mathbb R^d \to \mathbb R^D$, we consider a multi-layered fully-connected feed-forward network.
Given the encoder $\phi$ and the decoder $\psi$, we use a group-wise softmax operator $\sigma$ to map the output of the decoder to a per-input-feature probability simplex: given an initial binary vector $b \in \{0, 1\}^D$ constituting per-input-feature one-hot encodings $b^i$ (that is $b = [b^1 \oplus \ldots \oplus b^{c+r}]$), and a decoder output $B \in \mathbb R^D$ with per-input-feature constituents $B^i$ (that is $B = [B^1 \oplus \ldots \oplus B^{c+r}]$, we apply the softmax operation to each per-input-feature constituent to get $\hat b = [\hat b^1 \oplus \ldots \oplus \hat b^{c+r}] \in [0,1]^D$, where $\hat b^i = \textsf{softmax}(B^i)$.
We utilize the following per-sample reconstruction loss:
\begin{equation}
	\ell(b, \hat b) = \tfrac{1}{c+r}\textstyle\sum_{i = 1}^{c+r} \frac{1}{\log_2 |b^i|}\textsf{CE}(b^i, \hat b^i),
\end{equation}
where $\textsf{CE}$ is the standard cross-entropy loss between a one-hot vector and a softmax output, and $|b^i|$ is the length of the $i$-th constituent one-hot encoding in $b$, corresponding to the number of categories (or bins) in the $i$-th categorical (or numerical) feature.
This loss is a weighted average of the per-input-feature cross-entropy loss, with weights $(\nicefrac{1}{\log_2 |b^i|})$ to normalize the loss across all features with varying number of categories or bins.

The encoder and decoder are then learned by optimizing the following unsupervised loss:
\begin{equation}
	\label{eq:enc:recon_loss}
	\mathcal L_{R}(\phi, \psi) = \tfrac{1}{N} \textstyle\sum_{(x, y) \in S} \ell\left(P(x), \sigma(\psi(\phi(P(x))))\right),
\end{equation}
where $P$ is the data homogenizer, and $\sigma$ is the aforementioned group-wise softmax operator.
Learning the latent representation in such an unsupervised manner makes this distillation pipeline agnostic to the choice of downstream model. Another advantage of this choice is that the decoder allows us to map the distilled artificial samples in the latent space to the original features, which might be necessary in some applications (for interpretability reasons).

\subsubsection{Supervised latent space fine-tuning}
\label{apdx:subsec:ae-sft}
% \todo{instead of X-MH, can we have X-SFT or X-FT?}
%
Given the already learned encoder and decoder, we consider a supervised fine-tuning (FT) step where we utilize a classifier $f:\mathbb R^d \to Y$ that utilizes the latent representation.
The classifier is learned, and the encoder and decoder are fine-tuned by minimizing the following loss to ensure that the latent space is quite predictive while the reconstruction loss stays low:
\begin{equation}
	\mathcal L_R(\phi, \psi) + \textstyle \frac{\alpha}{N} \sum_{(x, y) \in S} \textsf{CE}(y, f(\phi(P(x)))),
\end{equation}
where $\alpha > 0$ is penalty parameter to balance the two losses, and $\textsf{CE}$ is the cross-entropy loss.
% \todo[author=inwon]{What is the $N$ in $\frac{\alpha}{N}$?}
We consider multi-layer FFN architecture as the classifier $f$.

\subsubsection{Encoder architectures}
\label{apdx:subsec:enc_arch}

\noindent\textbf{Fully-connected feed-forward network (FFN).}
This encoder first selects the column embeddings corresponding to nonzero entries in the binary representation $b$, concatenates them to get a $(c+r)M$-dimensional dense vectors (recall that $b$ will only have $c+r$ nonzeros out of the $D$ dimensions), and inputs them to a fully-connected feed-forward network $\mu: \mathbb R^{(c+r)M} \to \mathbb R^d$. The encoder $\phi: \{0, 1\}^D \to \mathbb R^d$ can be written as:
\begin{equation}
	z = \phi(b) = \mu(\oplus([w_i, i \in \{1, \ldots, D\}: b[i] = 1])),
\end{equation}
where $b[i]$ is the $i$-th entry of the $D$-dimensional vector, and $\oplus$ is the concatenation operator. The FFN $\mu$ and the column embeddings $\{w_i, i \in \{1, \ldots, D\}\}$ constitute the parameters of the encoder $\phi$. For a FFN with $H$ hidden layers, each of width $W$, the total number of parameters in this encoder is $O(DM + (c+r)MW + HW^2 + Wd)$.
\Cref{fig:enc:ffn} shows a simplified architecture of the FFN encoder.

\noindent\textbf{Graph neural network (GNN) encoder.}
We also consider a more recent encoder for tabular data proposed in~\citet{wuOpenWorldFeatureExtrapolation2021}. A bipartite graph is constructed between the column embeddings $\{w_i, i \in \{1, \ldots, D\}\}$ and the (zero-initialized) row (sample) embeddings $\{z_j \in \mathbb R^d, j \in \{1, \ldots, N\}\}$, with a bidirectional edge between $w_i$ and $z_j$ if the $b_j[i] = 1$, where $b_j \in \{0, 1\}^D$ is the binary representation of the $j$-th sample. Given the (learned) column embeddings, the row embeddings are obtained via multiple rounds of message passing through multiple GNN layers. This can be written as:
\begin{equation}
	\begin{split}
		 & z_j^{h} = \mu_h(z_j^{h-1}, \textsf{Agg}(w_i^{h-1}, i \in \mathcal N_j)), \\
		 & w_i^{h} = \mu_h(w_i^{h-1}, \textsf{Agg}(z_j^h, j \in \mathcal N_i)),
	\end{split}
\end{equation}
where $\mu_h$ is the $h$-th GNN layer, $\textsf{Agg}$ is an aggregation, $\mathcal N_i$ (or $\mathcal N_j$) is the neighbor set of the $i$-th column embedding (or $j$-th row embedding). We set the initial $z_j^0 = 0$ (zero-initialized row embeddings), $w_i^0 = w_i$, and utilize $z_j^H$ as the latent representation for distillation after $H$ GNN layers.
While~\citet{wuOpenWorldFeatureExtrapolation2021} only considered Graph Convolutional Networks~\cite{kipfSemiSupervisedClassificationGraph2016} as GNN modules, we extend it to GraphSage~\cite{hamiltonInductiveRepresentationLearning2017} and Graph Attention Networks~\cite{velickovicGraphAttentionNetworks2018}. An important aspect of the GNN encoder is that the desired row embedding size $d$ must match the column embedding size $M$, thus $d = M$. With $H$ GNN layers, the total number of parameters in this encoder is usually $O(Dd + Hd^2)$, which can be significantly smaller than the FFN encoder with moderately sized FFN (large enough $M$, $W$).
\Cref{fig:enc:gnn} shows the graph formulation (left) and the GNN encoder architeture (right).

\noindent\textbf{Transformer encoder.}
Finally, we consider a transformer-based autoencoder inspired by the architecture of FT-Transformer~\cite{gorishniyRevisitingDeepLearning2021}.
This encoder uses the same embedding layer as the FFN encoder, which is then followed by transformer blocks.
We learn an additional $cls$ embedding, which is placed before all other tokens in every sequence.
Each block takes in a sequence (one row) of $d$ embeddings, and is composed of a multihead-attention (MHA) module and a feed-forward network (FFN) module.

For a MHA module with $m$ attention heads, we modify the architecture seen in~\citep{gorishniyRevisitingDeepLearning2021} by allowing the dimension of the attention head to be separately configurable
-- i.e. instead of using $d/m$ as the dimension of a single attention head, we allow the module to compute the attention in $d_{qkv}$.
This choice is motivated by the fact that our encoders were trained with a latent size of $16$, which may not be wide enough for the TF encoder.
We then project the resulting embedding in $d_{qkv}m$-dimension back to $d$-dimension with $W_o$.
For an input $w_i$ at the $i$th transformer block, the computation for the MHA module is as follows:
\begin{equation}
	a_{i} = W_o^i(\textsf{softmax}(\frac{W_q^i(w_i)W_k^i(w_i)}{\sqrt{d_{qkv}}})W_v^i(w_i))
\end{equation}
The resulting attention score $a_i$ is then added with the original embedding and passed through an FFN module.
% \begin{equation}
% 	\begin{split}
% 		r_{i} = W_{f}^i(a_i) \\
% 		w_{i+1} = w_i + r_i
% 	\end{split}
% \end{equation}
% where $W_{\{q,k,v,o,f\}}^i$ are parameters of the $i$th transformer block.
Similarly to~\citet{gorishniyRevisitingDeepLearning2021}, the {\tt [cls]} embedding is used as the final output of the encoder.
\Cref{fig:enc:mha} shows our modified MHA component, and~\cref{fig:enc:tfblock} shows the TF encoder block.

\subsection{Distill Methods}

\subsubsection{Choice of Distill Methods (KIP, GM)}
\label{apdx:subsec:choices}
The clustering-based distillation schemes and KIP are not explicitly tied to a specific model and thus satisfy our desiderata of model-agnosticity. In contrast, the Gradient Matching or GM distillation scheme heavily relies on the choice of the backbone model $M_\theta$ (as well as the learning algorithm parameters such as the learning rate), and there is no guarantee that the distilled samples $R$ would be useful for any other model.
Thus, this scheme is not model-agnostic. However, we consider GM to be representative of the model-specific distillation schemes for the sake of completeness of our evaluations.
% Thus, we consider GM to be representative of the non model-agnostic distillation schemes for the sake of completeness of our evaluations.
For our table distillation, we choose $M_\theta$ to be a multi-layered perceptron with a single hidden layer. This will pose a mismatch when we evaluate the quality of the distilled data $R$ on standard tabular models such as decision tree ensembles and nearest-neighbor models, highlighting the need for model-agnosticity in tabular data distillation.

\subsubsection{Distill Method Implementaion}
\label{apdx:subsec:dd_params}
% \todo[author=inwon,inline]{need to do}

\begin{table}[t!]
	\centering
	\caption{Parameters of distillation methods.}
	\label{apdx:tab:dd_params}
	{
		\small
		\begin{tabular}{lllp{3in}}
			\hline
			\toprule
			Method                  & Hyperparameter                     & Value  & Description                                                                                                                                                                  \\
			\midrule
			\multirow{4}{*}{Common} & \texttt{distill space}             & -      & Whether to use the encoder latent space or the raw binary representation.                                                                                                    \\
			                        & \texttt{use\_closest}*             & -      & Whether to use \textit{median} points instead of the euclidean center. Only applicable to clustering methods.                                                                \\
			                        & \texttt{output\_space}$^{\dagger}$ & -      & Whether to keep the encoder latent/ decode or use the raw binary space. The binary space is only applicable to clustering methods when \texttt{use\_closest} is set to True. \\
			                        & \texttt{random\_seed}$^{\ddagger}$ & -      & Random seed for distillation algorithm. Not applicable to agglomerative.                                                                                                     \\
			\midrule
			\multirow{2}{*}{KIP}    & \texttt{n\_epochs}                 & $1000$ & Number of epochs to train the \textit{distilled data}.                                                                                                                       \\
			                        & \texttt{mlp\_dim}                  & $1024$ & Width of the neural network to compute the NTK of.                                                                                                                           \\
			\midrule
			\multirow{5}{*}{GM}     & \texttt{n\_epochs}                 & $500$  & Number of epochs to train the \textit{distilled data}.                                                                                                                       \\
			                        & \texttt{mlp\_dim}                  & $1024$ & Size of the hidden layer of the target model.                                                                                                                                \\
			                        & \texttt{n\_layers}                 & $2$    & Number of hidden layers in the target model.                                                                                                                                 \\
			                        & \texttt{lr\_mlp}                   & $0.01$ & Learning rate for the target model.                                                                                                                                          \\
			                        & \texttt{lr\_data}                  & $0.1$  & Learning rate for the \textit{distilled data}.                                                                                                                               \\
			                        & \texttt{mom\_data}                 & $0.5$  & Momentum for \textit{distilled data}.                                                                                                                                        \\
			\bottomrule
		\end{tabular}
	}
\end{table}

\paragraph{$k$-means}
We use the \texttt{sklearn.cluster.KMeans} from~\cite{scikit-learn} with the \texttt{n\_init} set to \texttt{"auto"}.

\paragraph{Agglomerative}
We use \texttt{sklearn.cluster.AgglomerativeClustering} from~\cite{scikit-learn} with the \texttt{linkage} set to \texttt{"ward"}.
Because agglomerative clustering does not have a ``centroid'', we manually calculate a euclidean centroid for each cluster by using \texttt{sklearn.neighbors.NearestCentroid} to compute the centroid or the closest \textit{real} point.

\paragraph{KIP}
We use the implementation provided by~\cite{nguyenDatasetMetaLearningKernel2021} available at~\url{https://github.com/google-research/google-research/tree/master/kip}.

\paragraph{GM}
We use the implementation provided by~\cite{zhao2021dataset} available at~\url{https://github.com/VICO-UoE/DatasetCondensation}.

\Cref{apdx:tab:dd_params} shows the parameters available for each distillation methods.
The common parameters are used for every algorithm, with the exceptions marked on the right-most column.
The method-specific parameters for KIP and GM are for the original algorithms as proposed in~\citet{nguyenDatasetMetaLearningKernel2021,zhao2021dataset}.

\subsection{Downstream Classifier Hyperparameters}
\label{apdx:subsec:clf_hpo}

\begin{table}[t]
	\centering
	\caption{Hyperparameters of downstream classifiers.}
	\label{apdx:tab:clf_hpo}
	{
		\small
		\begin{tabular}{llc}
			\hline
			\toprule
			Classifier                             & Hyperparameter                      & Value          \\
			\midrule
			\multirow{10}{*}{FT-Transformer}       & \texttt{d\_token}                   & $128$          \\
			                                       & \texttt{n\_blocks}                  & $2$            \\
			                                       & \texttt{attention\_n\_heads}        & $8$            \\
			                                       & \texttt{attention\_dropout}         & $0.15$         \\
			                                       & \texttt{ffn\_d\_hidden\_multiplier} & $1.25$         \\
			                                       & \texttt{ffn\_dropout}               & $0.05$         \\
			                                       & \texttt{residual\_dropout}          & $0$            \\
			                                       & \texttt{learning\_rate}             & $10^{-4}$      \\
			                                       & \texttt{weight\_decay}              & $10^{-5}$      \\
			                                       & \texttt{early\_stopping}            & \texttt{True}  \\
			\midrule
			\multirow{1}{*}{Naive Bayes}           & \texttt{var\_smoothing}             & $10^{-9}$      \\
			\midrule
			\multirow{3}{*}{$K$-Nearest-Neighbors} & \texttt{n\_neighbors}               & $5$            \\
			                                       & \texttt{leaf\_size}                 & $30$           \\
			                                       & \texttt{p}                          & $2$            \\
			\midrule
			\multirow{4}{*}{Logistic Regression}   & \texttt{penalty}                    & \texttt{l2}    \\
			                                       & \texttt{tol}                        & $10^{-4}$      \\
			                                       & \texttt{C}                          & $1$            \\
			                                       & \texttt{solver}                     & \texttt{lbfgs} \\
			\midrule
			\multirow{4}{*}{MLP}                   & \texttt{d\_hidden}                  & $100$          \\
			                                       & \texttt{n\_hidden}                  & $1$            \\
			                                       & \texttt{learning\_rate}             & $10^{-4}$      \\
			                                       & \texttt{early\_stopping}            & \texttt{True}  \\
			\midrule
			\multirow{8}{*}{ResNet}                & \texttt{n\_blocks}                  & $4$            \\
			                                       & \texttt{d\_block}                   & $128$          \\
			                                       & \texttt{d\_hidden\_multiplier}      & $1.25$         \\
			                                       & \texttt{dropout}                    & $0.2$          \\
			                                       & \texttt{learning\_rate}             & $0.0001$       \\
			                                       & \texttt{weight\_decay}              & $0.00001$      \\
			                                       & \texttt{early\_stopping}            & \texttt{True}  \\
			                                       & \texttt{patience}                   & $16$           \\
			\bottomrule
		\end{tabular}
	}
\end{table}

\Cref{apdx:tab:clf_hpo} shows the hyperparameters used for each downstream classifier.
We use scikit-learn~\cite{scikit-learn}'s implementation of Naive Bayes, $K$-Nearest-Neighbors, Logistic Regression, and MLP, and~\citet{gorishniyRevisitingDeepLearning2021}'s implementation of FT-Transformer and ResNet.

\subsection{ResNet and FT-Transformer performance}
\label{apdx:subsec:resnet_ft}

\begin{table}
	\centering
	\caption{Average train/test times and test performance comparison for all downstream classifiers.}
	\label{tab:apdx:tab:resnet_ft_times}
	{\small
		\begin{tabular}[c]{lrrr}
			\toprule
			Classifier    & Train Time & Test Time & Test Perf. \\ \midrule
			FTTransformer & 281.3431   & 0.17934   & 0.7879     \\
			NB            & 0.0030     & 0.00232   & 0.6624     \\
			KNN           & 0.0007     & 0.54309   & 0.7474     \\
			LR            & 0.4901     & 0.00646   & 0.7709     \\
			MLP           & 2.4444     & 0.00554   & 0.7826     \\
			ResNet        & 154.9824   & 0.08508   & 0.7833     \\
			XGB           & 11.4055    & 0.01439   & 0.8180     \\
			\bottomrule
		\end{tabular}
	}
\end{table}

% As discussed in~\cref{sec:datasets_and_models}, 
We test ResNet and FT-Transformer for 5 datasets.
We found that even with early stopping, the two classifiers take significantly longer to train given the same computing resources.
On average, we find that ResNet takes around 10 times longer to finish training, while FT-Transformer takes around 28 times when compared to XGBoost.
We also find that the performance of resnet and FT-Transformer does not stand out -- in fact, the average test performance when trained on the full dataset shows that both ResNet and FTTransformer show a similar performance to MLP, and are outperformed by XGBoost.

\subsection{Determining the best overall performance}
\label{apdx:subsec:best_overall_perf}

We describe the best overall pipeline in~\cref{sec:results:best_overall} and~\cref{tab:results:best_performer_by_count}.
Here, we provide a more detailed explanation of how we determined the best overall pipeline.
The runs are grouped by their classifier, dataset and distill size $n$.
Similar to other parts of analysis, the grouping is done in order to ensure that the comparisons are \textit{fair}.
In this instance, we are interested in only the pipeline components that lead to the best classifier performance, regardless of the exact classifier kind.
Thus, we group every run by their non-pipeline-specific parameters, which are the classifier, dataset and distill size $n$.
In each group, we then count the instances the pipeline places on the top 3 in terms of the regret score and sum up the counts for each pipeline.

Following the previous findings,~\cref{tab:results:best_performer_by_count} shows that $k$-means based methods have the best performance, placing in the top 3 with all SFT encoder variants.
Surprisingly, we also find pipelines that use KIP and GM as the 4th and 5th best performers.
While we were not able to determine any specific conditions that cause KIP and GM to place on top, this result shows that there are exist some conditions which leads the pipelines using gradient-based methods (KIP, GM) to be the top performer.
On the other hand, the consistent rank placement of pipelines that use the autoencoder latent space shows that fine-tuned autoencoders can indeed boost the performance of distillation methods significantly.
% while $k$-means is the \textit{overall best}, 

\section{Additional Analysis}

\subsection{Full results of distillation methods by downstream classifiers}
\begin{figure}
	\centering
	\includegraphics[width=0.8\linewidth]{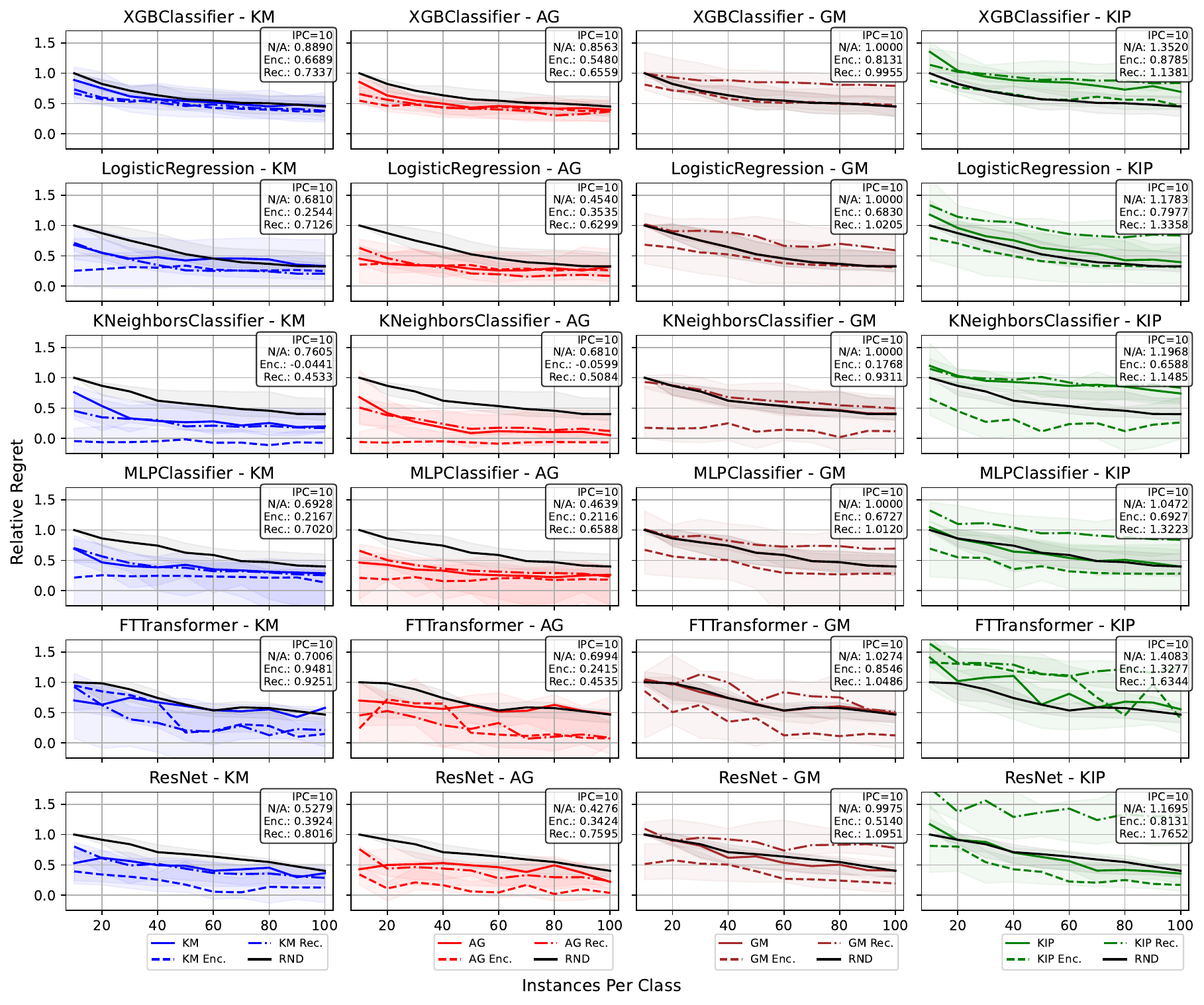}
	\caption{Full results of distillation methods by downstream classifiers.}
	\label{fig:apdx:dm-per-clf-per-dm}
\end{figure}

\subsection{Effect of column embedding scheme on downstream performance}
\label{apdx:sec:col_emb_effect}

\begin{table}[t]
	\centering
	\captionof{table}{
		A comparison of relative regret scores of distillation pipelines that use the encoded space of autoencoders trained with different column embeddings, tested on 5 datasets (Adult, Amazon Employee Access, Credit, House, Phishing Websites).
		The center value shows the median relative regret, and smaller values on each side refers to the first and third quantile, respectively.
		In general, PLE embeddings show the strongest performance.
		However, it is worth noting that PLE embeddings are not applicable to GNN encoders, and that binary embeddings also show superior performance to scaled embeddings.
		% Binary column embeddings yield consistently lower regret scores. 
	}
	\label{tab:results:comare-col-emb}
	{\small \begin{tabular}{lcccc}
\toprule
Col. Emb. & KM & AG & GM & KIP \\
\midrule
Binary & $_{0.1082}$ 0.5645 $_{0.7886}$ & $_{0.0976}$ 0.4633 $_{0.7181}$ & $_{0.5504}$ 0.9038 $_{1.0063}$ & $_{0.6551}$ 0.9254 $_{1.1918}$ \\
Scaled & $_{0.7214}$ 0.8613 $_{1.0671}$ & $_{0.4908}$ 0.6939 $_{1.0249}$ & $_{1.0092}$ 1.4412 $_{1.8658}$ & $_{1.3304}$ 1.6137 $_{2.2985}$ \\
PLE & $_{-0.2428}$ 0.1976 $_{0.9305}$ & $_{-0.2698}$ 0.2173 $_{0.6752}$ & $_{-0.0865}$ 0.2747 $_{1.0524}$ & $_{-0.0263}$ 0.7398 $_{1.3923}$ \\
\bottomrule
\end{tabular}
}
\end{table}

While column embeddings are standard for categorical columns -- each category is represented with a vector, there are various ways of embedding numerical columns:
(i)~A numerical feature can be binned, and each bin treated as a category with an embedding $\rvw \in \R^m$ corresponding to each bin.
(ii)~With linearly scaled column embeddings, a single column embedding $\rvw \in \R^m$ is used for each numerical column, and the column embedding for a particular numerical value $v \in \R$ is obtained by scaling $\rvw$ to $v \cdot \rvw$.
(iii)~Piecewise linear encoding or PLE~\citep{gorishniyEmbeddingsNumericalFeatures2022} also bin the numerical feature but use a more sophisticated way of generating the column embeddings for a given numerical value.
% Here we consider on the binned numerical features for a couple of reasons:
We considered binned numerical features in the main paper for a couple of reasons:
(a)~Binned numerical features naturally handle missing values (quite prevalent in tabular data) by maintaining a ``missing'' bin instead of relying on a heuristic intermediate imputation step; sometimes, the fact that a value is missing is in itself a signal, and heuristic imputation schemes often lose this information.
(b)~The binned features can be used for all architectures we consider here -- FFN, Transformer, and GNN -- and using a common embedding scheme allows us to ablate the effect of the different architectures. The other numerical embedding schemes do not apply to GNNs.

% One question that remains is the how the data is transformed in the process of latent projection.
% In our experiments, we mainly consider column embeddings to convert the original tabular data for the autoencoders.
To understand the effect of different kinds of column embeddings schemes, we conduct a smaller scale experiment on 5 datasets.
Specifically, we compare scaled embeddings as seen in~\cite{gorishniyRevisitingDeepLearning2021}, piecewise linear encoding (PLE) as seen in~\cite{gorishniyEmbeddingsNumericalFeatures2022}, against using binary column embeddings where continuous features are binarized by binning, and examine the downstream performance of distillation pipelines that use the latent space of the autoencoders trained with the corresponding column embedding scheme.
% training autoencoders using binary and scaled column embeddings.
\Cref{tab:results:comare-col-emb} shows that using the \textbf{both binary column embeddings and PLE consistently leads to lower regret scores compared to scaled column embeddings}.
% \todo{fix this to weaker argument based on medians..}
While PLE embeddings show the strongest performance, they are not applicable to the GNN autoencoder architecture.
Thus, we conduct most of our experiments using binary column embeddings for a fair comparison across different autoencoder architectures for a fair comparison.

\begin{figure}
	\centering
	\includegraphics[width=\linewidth]{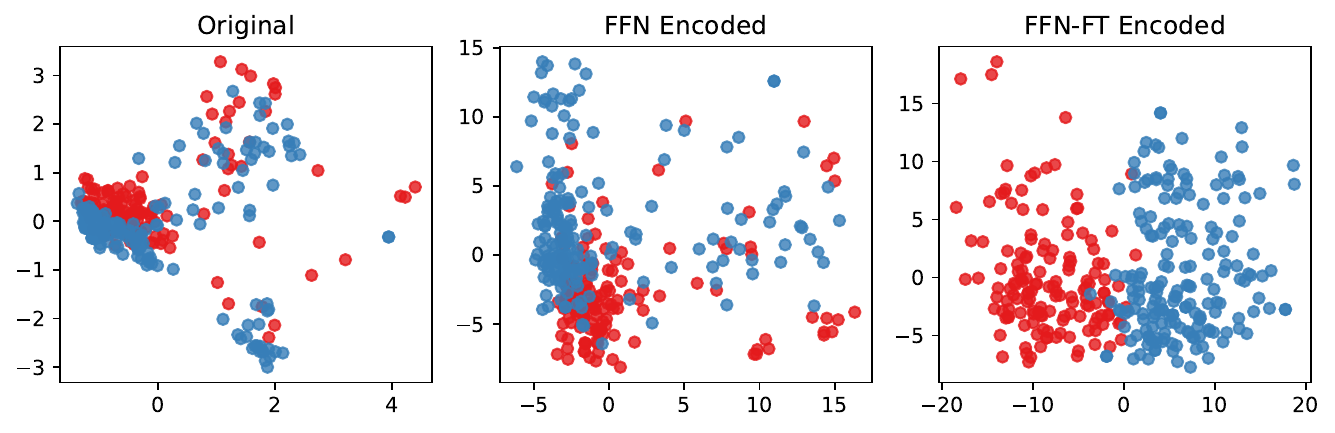}
	\caption{PCA visualization of Phishing Websites dataset.}
	\label{fig:phishing_pca}
\end{figure}

\begin{figure}
	\centering
	\includegraphics[width=\linewidth]{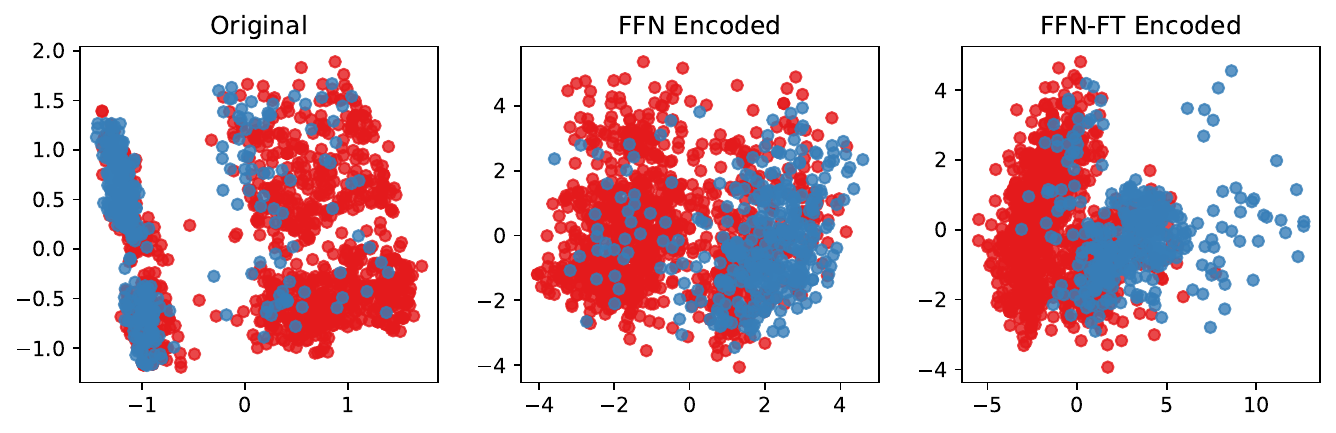}
	\caption{PCA visualization of the Adult dataset.}
	\label{fig:adult_pca}
\end{figure}

\subsection{Effect of supervised fine-tuning.}
~\Cref{fig:phishing_pca,fig:adult_pca} show the PCA visualizations the adult and tencent CTR datasets in the original, FFN-encoded, FFN-SFT encoded representations.
Both figures show that while the distribution inside the vanilla FFN's latent space does not look significantly different from the original space, adding supervised fine-tuning leads to a clearer separation between different classes.

\begin{figure}
	\centering
	\includegraphics[width=\linewidth]{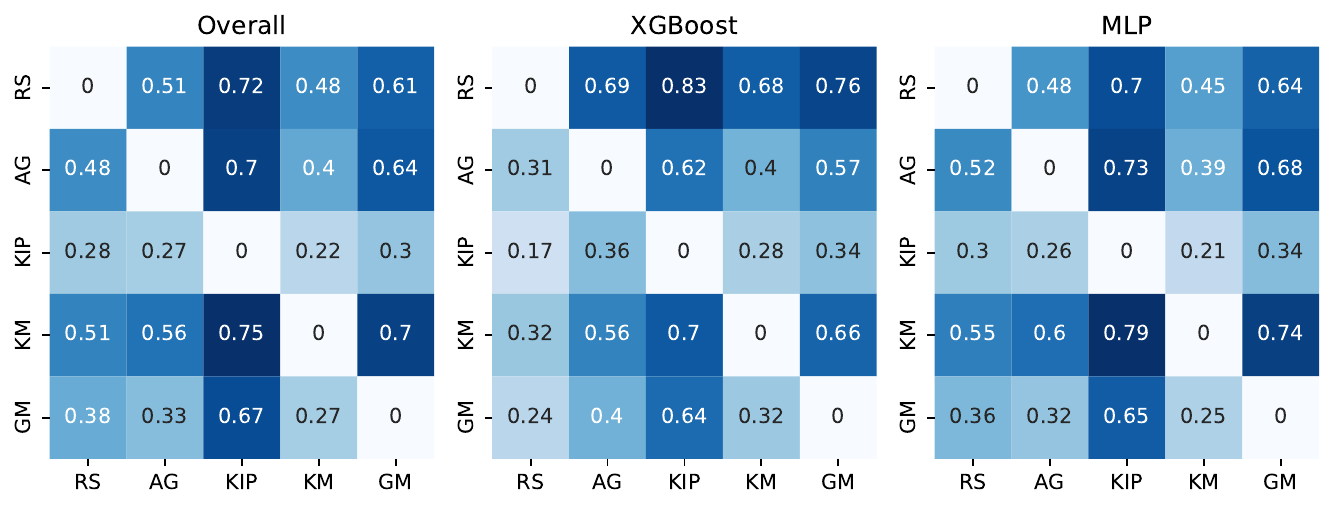}
	\caption{\textit{Pairwise comparision of distillation methods.} The relative performances of distillation methods under otherwise equal sttings. Rows denote \textit{win} ratio, columns denote \textit{loss} ratio.}
	\label{fig:dm_vs_matrix}
\end{figure}

\subsection{Pairwise comparision of distillation methods.}
In addition, we compare the downstream classifier performance with every pair of pipelines that use different distillation methods under otherwise equal settings.
The left table of~\cref{fig:dm_vs_matrix} reveals that KIP had the highest tendency to underperform other distillation methods, while \textit{k}-means had the highest tendency to outperform other distillation methods.
This is consistent with our previous findings, where \textit{k}-means outranked other distillation methods most frequently.
% We suspect the reason behind KIP's underperformance is the fact that KIP's kernel is still based on a neural network.
In order to gain further insights behind the performance lag of graident-based distillation methods, we conduct a pairwise comparison of the distillation methods for different classifiers as well.
The center and right tables of~\cref{fig:dm_vs_matrix} shows the pairwise comparison of distillation methods for XGBoost and MLP as downstream models.
% We find that KIP continues to underperform every other distillation methods considered.
This suggests that gradient-based methods' underperformance is not solely due to its kernel, but that tabular data itself may pose a unique challenge in distillation that is not seen in image data.
It is also worth noting that while the clustering-based approaches had the best overall rank, random sampling proved to be a strong baseline with a near 50\% win ratio against them.

\section{Documentation of TDBench}
\label{apdx:sec:tdbench}
The information in this section is also available in a markdown format in the \texttt{README.md} file of the supplementary material.

\subsection{Reproducing Results}
\label{apdx:sec:tdbench:reproduce}

Every plot and table in the main paper can be reconstructed using the following scripts:

\begin{itemize}
	\item \texttt{Q0\_experiment\_scale.py}
	\item \texttt{Q1\_1\_col\_embeds.py}
	\item \texttt{Q1\_encoding.py}
	\item \texttt{Q2\_distill\_methods.py}
	\item \texttt{Q3\_autoencoders.py}
	\item \texttt{Q4\_1\_runtime.py}
	\item \texttt{Q4\_2\_get\_hpo\_dirs.py}
	\item \texttt{Q4\_2\_hpo.py}
	\item \texttt{Q4\_combinations.py}
	\item \texttt{Q5\_class\_imbal.py}
\end{itemize}

The scripts are organized in order of the question addressed in~\cref{sec:results} and will be populated in \texttt{iclr-figures} directory.
These can be simply ran by calling \texttt{python SCRIPT\_NAME}.

The following files are included in the supplementary material and contain all the necessary information for the scripts:

\begin{itemize}
	\item \texttt{dataset\_stats.csv}
	\item \texttt{enc\_stats.csv}
	\item \texttt{*data\_mode\_switch\_results.csv}
	\item \texttt{hpo-measure/}
	\item \texttt{*mixed\_tf\_results.csv}
  \item \texttt{*ple\_tf\_results.csv}
\end{itemize}

The files marked with an asterisk (*) are not included in the repository, but can be downloaded from this url: \url{https://drive.google.com/drive/folders/1tJ5e1iCvaz-UbxEgpmuCPj-58crgYRJW?usp=share_link}

\subsection{Description of the workflow}
\label{apdx:sec:tdbench:workflow}

The \texttt{\#\# Running the Code} section of \texttt{README.md} file discusses the actual commands and available options for running each stage in detail.

The procedure is as follows:
\begin{itemize}
	\item Train the autoencoder with the desired configuration.
	\item (\textit{Optional}) Fine-tune the autoencoder with a classifier head.
	\item Run distillation methods against specified downstream classifiers.
\end{itemize}

\subsection{Constructing a new pipeline}
\label{apdx:sec:tdbench:newpipeline}

\paragraph{Changing default parameters}

The configurations for this project are managed by hydra and can be modified by adding new files/directories under the `config` directory.

\paragraph{Adding new datasets}

Adding new datasets is as simple as adding a new \texttt{config/data/datasets/DATASET\_NAME.yaml} file.
Currently, only openml datasets are supported.

\begin{table}[h]
	\centering
	{\small
		\begin{tabular}[c]{ll}
			\toprule
			\textbf{Field}         & \textbf{Type} \\ \midrule
			\texttt{dataset\_name} & string        \\
			\texttt{download\_url} & string        \\
			\texttt{label}         & string        \\
			\texttt{n\_classes}    & int           \\
			\texttt{source\_type}  & string        \\
			\bottomrule
		\end{tabular}
	}
	\caption{Configuration details for datasets}
	\label{tab:docs:dset_conf}
\end{table}

The following flags must be specified for the dataset to be correctly loaded as seen in~\cref{tab:docs:dset_conf}.

\paragraph{Adding new preprocessing methods}

\begin{table}[h]
	\centering
	{\small
		\begin{tabular}[c]{ll}
			\toprule
			\textbf{Field}       & \textbf{Type} \\ \midrule
			\texttt{parse\_mode} & string        \\
			\texttt{scale\_mode} & string        \\
			\texttt{bin\_strat}  & string        \\
			\texttt{n\_bins}     & int           \\
			\bottomrule
		\end{tabular}
	}
	\caption{Configuration details for data preprocessing}
	\label{tab:docs:dmode_conf}
\end{table}

The preprocessing is handled by the \texttt{TabularDataModule} object that lives in \texttt{tabdd/data/tabulardatamodule.py}.
The preprocessing strategies are identified by a string, and can be configured under \texttt{config/data/mode}.
The fields seen in~\cref{tab:docs:dmode_conf} must be specified for the preprocessing to work correctly.
One can additionally define any type of \texttt{scale\_mode} or \texttt{bin\_strat,} which will be consumed by the \texttt{TabularDataModule.}

This object is configured with \texttt{DatasetConfig} and \texttt{DataModeConfig.}
The \texttt{DatasetConfig} is the configuration for the dataset, and the \texttt{DataModeConfig} is the configuration for the preprocessing method.

It's \texttt{TabularDataModule.prepare\_data} is the method that will parse the data accordingly and save to cache.
One can add arbitrary preprocessing methods in this file by adding new flags to \texttt{DataModeConfig} and handling it inside the \texttt{prepare\_data} method.

\paragraph{Adding new distillation methods}

\begin{table}[h]
	\centering
	{\small
		\begin{tabular}[c]{ll}
			\toprule
			\textbf{Field}             & \textbf{Type} \\ \midrule
			\texttt{is\_random}        & string        \\
			\texttt{is\_cluster}       & string        \\
			\texttt{can\_use\_encoder} & string        \\
			\texttt{args}              & int           \\
			\bottomrule
		\end{tabular}
	}
	\caption{Configuration details for distillation methods}
	\label{tab:docs:distill_conf}
\end{table}

The distillation methods are identified by a string, which should have a configuration with the same name under \texttt{config/distill/methods}.
Once can characterize the method the following fields seen in~\cref{tab:docs:distill_conf}.

\begin{itemize}
	\item \texttt{is\_random}: Whether there is randomness in the method. If true, the pipeline will be ran multiple times.
	\item \texttt{is\_cluster}: Whether the method is a clustering method. If true, an option that uses the nearest-to-center method will be included.
	\item \texttt{can\_use\_encoder}: Whether the method can be applied in the latent space.
	\item \texttt{args}: any additional arguments to the actual function.
\end{itemize}

Once the configuration is created, it will be consumed by \texttt{load\_distilled\_data} method of \texttt{tabdd/distill/load\_distilled\_data.py.}
This method can then be modified to include the new distillation method.

\paragraph{Adding new encoders}

All encoders used in the benchmark are subclasses \texttt{BaseEncoder} from \texttt{tabdd/models/encoder/base\_encoder.py}.
A simple example of how to implement can be seen in \texttt{tabdd/models/encoder/mlp\_autoencoder.py}.
The module needs to encoder the following methods: \texttt{\_\_init\_\_()}, \texttt{encode}, \texttt{decode} and \texttt{forward}.

The autoencoders are specified by the configuration files in \texttt{config/encoder/models/}.
The class of the encoder is specified by \texttt{cls}, and the hyperparameters are specified by \texttt{tune\_params}.

\section{Additional Analysis}
\subsection{Additional Distillation Methods}
\label{apdx:sec:additional_distill_methods}

% \begin{figure}
% 	\centering
% 	\begin{subfigure}[t]{0.3\textwidth}
% 		\centering
% 		\includegraphics[width=\textwidth]{./figures/iclr-rebuttal/rq2-distill-method-crit-diff-n10.pdf}
% 		\vspace{-12px}
% 		\caption{IPC=10}
% 	\end{subfigure}
% 	\begin{subfigure}[t]{0.3\textwidth}
% 		\centering
% 		\includegraphics[width=\textwidth]{./figures/iclr-rebuttal/rq2-distill-method-crit-diff-n50.pdf}
% 		\vspace{-12px}
% 		\caption{IPC=50}
% 	\end{subfigure}
% 	\begin{subfigure}[t]{0.3\textwidth}
% 		\centering
% 		\includegraphics[width=\textwidth]{./figures/iclr-rebuttal/rq2-distill-method-crit-diff-n100.pdf}
% 		\vspace{-12px}
% 		\caption{IPC=100}
% 	\end{subfigure}
% 	\vspace{-8px}
% 	\caption{
% 		Critical difference plot comparing ranks of distillation methods across datasets per IPC value when applied with TF-SFT encoder for XGBoost classifier with additional baselines
% 		The x-axis denotes the average rank, and a black horizontal line connects groups of methods that are \textit{not significantly different} in the rank distribution.
% 		$k$-means and agglomerative are indistinguishable from each other in $\text{IPC} \in \{10, 50\}$, but $k$-means gains an edge in IPC=100.
% 		(FG: Forgetting, GN: GraNd, GL: Glister, GC: Graph Cut)
% 	}
% 	\label{apdx:fig:results:distill_method_crit}
% 	% \vspace{-12px}
% \end{figure}

\begin{table}[t]
	\centering
	\caption{
		Relative regret of pipelines that use different combinations of distill methods and encoders at IPC=10, aggregated over classifiers. The best value for each column is marked with \textbf{bold}, and the second best is marked with \underline{underline}.
		(FG: Forgetting, GN: GraNd, GL: Glister, GC: Graph Cut)
	}
	\label{apdx:tab:results:distill_method}
	\vspace{-6px}
	{\footnotesize\begin{tabular}{llcccccc}
	\toprule
	Distill Method & \multicolumn{6}{c}{Regret}                                                                                                          \\
	               & Min                        & Q1                 & Mean               & Median             & Q3                 & Max                \\
	\midrule
	Random Sample  & 1.0000                     & 1.0000             & 1.0000             & 1.0000             & 1.0000             & 1.0000             \\
	\midrule
	KM             & \underline{0.0597}         & \underline{0.5256} & \underline{0.6682} & \underline{0.6654} & \textbf{0.8186}    & \underline{1.1094} \\
	AG             & \textbf{0.0000}            & \textbf{0.5177}    & \textbf{0.6301}    & \textbf{0.6036}    & \underline{0.8914} & \textbf{0.9965}    \\
	KIP            & 0.6728                     & 0.8483             & 1.1109             & 1.0544             & 1.2523             & 2.2713             \\
	GM             & 0.4175                     & 0.7707             & 0.9858             & 0.9377             & 1.1461             & 1.7292             \\
	FG             & 0.8705                     & 1.1400             & 2.3837             & 1.4465             & 1.8731             & 16.0146            \\
	GN             & 0.7748                     & 1.1498             & 2.0530             & 1.3704             & 2.2624             & 10.6670            \\
	GL             & 0.8376                     & 1.1000             & 2.0907             & 1.3146             & 1.6823             & 14.1625            \\
	GC             & 0.6361                     & 0.9077             & 1.5084             & 1.1031             & 1.5998             & 6.8392             \\
	MTT            & 0.4175                     & 0.7707             & 1.0340             & 0.9699             & 1.2176             & 2.3026             \\
	DATM           & 0.4175                     & 0.7707             & 1.0340             & 0.9699             & 1.2176             & 2.3026             \\
	\bottomrule
\end{tabular}
}
\end{table}

% We conduct a further comparison of more recent distillation methods against the methods compared in~\cref{sec:results} to verify whether these methods will show superior performance. Specifically, we incorporate four representative NN-based coreset selection methods examined in Deepcore~\citep{guoDeepCoreComprehensiveLibrary2022} -- Forgetting~\citep{toneva2018empirical}, GraNd~\citep{paul2021deep}, Glister~\citep{killamsetty2021glister}, Graph Cut~\citep{iyer2013submodular}) and MTT~\citep{cazenavetteDatasetDistillationMatching2022b} and DATM~\citep{guoLosslessDatasetDistillation2023}. The results are presented in Table~\ref{apdx:tab:results:distill_method} and Figure~\ref{apdx:fig:results:distill_method_crit}. Consistent to our findings in~\cref{sec:results}, we find that more recent distillation methods that rely on NNs do not fair well on non-differentiable downstream classifier (XGBoost), and that clustering methods still show dominance. It is also interesting to note that GM shows superior performance to MTT and DATM, suggesting that the latter two methods may actually be overfitting to the teacher network's architecture.

\subsection{Dataset Feature Correlation}

\begin{figure}
	\centering
	% \begin{subfigure}[t]{\textwidth}
	% 	\centering
	% 	\includegraphics[width=0.95\textwidth]{./figures/iclr-rebuttal/rq6-feature-corr-credit.pdf}
	% 	\vspace{-8px}
	% 	\caption{Credit}
	% \end{subfigure}
	\begin{subfigure}[t]{\textwidth}
		\centering
		\includegraphics[width=0.95\textwidth]{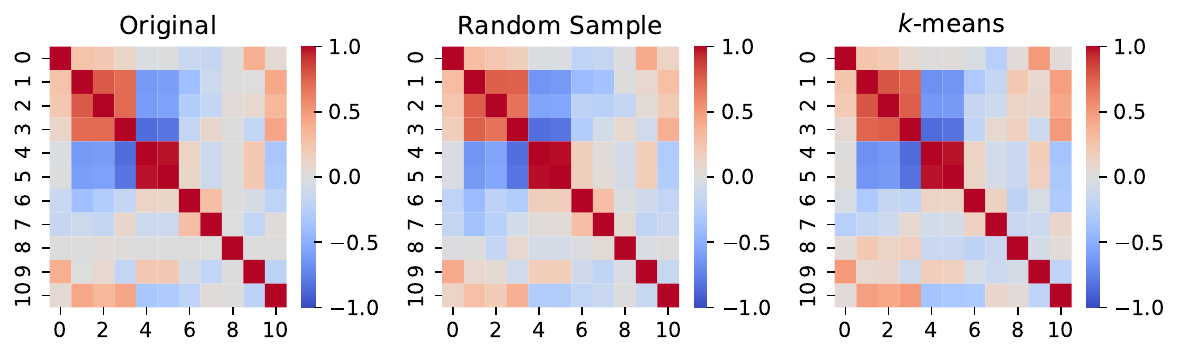}
		\vspace{-8px}
		\caption{Magic Telescope}
	\end{subfigure}
	\begin{subfigure}[t]{\textwidth}
		\centering
		\includegraphics[width=0.95\textwidth]{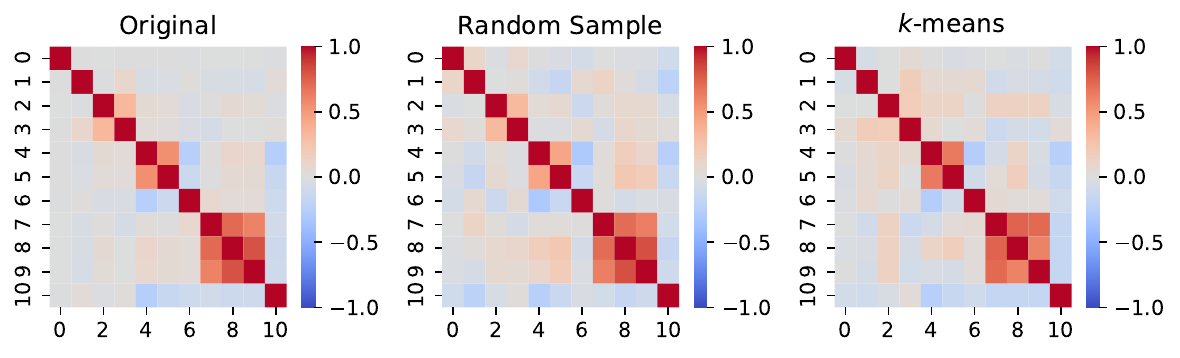}
		\vspace{-8px}
		\caption{Tencent CTR}
	\end{subfigure}
	\caption{A side-by-side comparison of correlation of numerical features in the training data before distillation, after random sampling@IPC=100, and after distillation@ICP=100. While some weaker correlations are not entirely accurately portrayed, the distilled data preserves the stronger correlations remarkably well.}
	\label{apdx:fig:results:feature_corr}
\end{figure}

We further investigate the presevation of feature correlation in the distilled data. \Cref{apdx:fig:results:feature_corr} shows the change in feature correlation in the original, randomly sampled and distilled with $k$-means in the latent space of TF-SFT in 3 datasets -- Credit, Magic Telescope and Tencent CTR.

\subsection{Relation to Previous Work}

\cite{kang2024effective} presented a preliminary abstract on work that explores data distillation for tabular data. The authors utilize an MLP and GNN based autoencoder networks to transform the data to before distilling and show that simple clusetering-based methods can outperform competetive distillation algorithms proposed in computer vision (KIP~\citep{nguyenDatasetMetaLearningKernel2021})

Building upon this work, our work provides a comprehensive analysis of distillation methods on tabular data, and provides a detailed comparison of distillation methods across a wide range of datasets and classifiers. We also provide a detailed analysis of the effect of IPC on the performance of distillation methods, and provide insights into the effect of distillation on the feature correlation of the data.

Below, we provide a detailed comparison of our work with the preliminary abstract presented by~\cite{kang2024effective}:

\begin{itemize}
	\item We conduct a comprehensive comparison of different binning methods and their effect on downstream performance.
	\item We test with a transformer-based autoencoder, and show that it outperforms MLP and GNN based autoencoders.
	\item We additionally consider gradient matching~\cite{zhao2021dataset} as an additional baseline to represent the gradient-based family of distillation methods~\cite{cazenavetteDatasetDistillationMatching2022b,zhao2023dataset,guoLosslessDatasetDistillation2023}
	\item We provide a complete python package, TDBench, that can be used and extended by anyone in the community.
	\item We explore a realistic use case for data distillation in the context of HPO and show the trade-offs in utility and cost saving.
	\item We introduce a relative regret metric to compare the performance of different distillation methods across datasets and classifiers.
\end{itemize}

\subsection{Raw balanced Accuracy score}

Below is a comparison of the raw balanced accuracy of each distillation pipelines averaged over random iterations.~\Cref{apdx:tab:results:raw_ba_xgb} shows a comparison of all 10 distillation methods that were ran with TF-SFT encoder and tested on XGB downstream classifier, and~\cref{apdx:tab:results:raw_ba_xgb_none} shows the performance of the baseline methods when applied without the encoders. ~\Cref{apdx:tab:results:raw_ba_knn,apdx:tab:results:raw_ba_knn_none} show the same comparison that with and without TF-SFT encoder for the 4 baselines methods ($k$-means, aggloermative, KIP, GM) on KNN classifier, and~\cref{apdx:tab:results:raw_ba_mlp,apdx:tab:results:raw_ba_mlp_none} show the same for MLP classifier.

The last two rows of the tables each denote the number of instances that the pipeline ranked at the top, and the number of times it outperformed random sampling. The results show that random sampling is not a trivial baseline for many methods, and that both clustering methods, AG and KM, show the strongest performance. We also see that adding the encoder to the pipeline significantly increases the downstream performer of all 3 representative models.

\clearpage

\begin{table}
	\begin{adjustbox}{angle=90}
		\begin{minipage}{\textheight}
			\centering
			{
				\tiny
				\begin{tabular}{lllllllllll}
\toprule
Dataset & AG & DATM & FG & GM & GL & GN & GC & KIP & KM & MTT \\
\midrule
AD & \underline{0.7409$\pm$0.0128} & 0.7304$\pm$0.0316 & 0.6779$\pm$0.0458 & 0.7304$\pm$0.0316 & 0.5241$\pm$0.0000 & 0.5241$\pm$0.0000 & 0.6491$\pm$0.0545 & 0.7174$\pm$0.0517 & \textbf{0.7425$\pm$0.0220} & 0.7304$\pm$0.0316 \\
AE & 0.5072$\pm$0.0017 & 0.5233$\pm$0.0250 & 0.4960$\pm$0.0355 & 0.5253$\pm$0.0308 & 0.4950$\pm$0.0086 & 0.5149$\pm$0.0348 & \textbf{0.5418$\pm$0.0119} & 0.5107$\pm$0.0228 & \underline{0.5292$\pm$0.0210} & 0.5233$\pm$0.0250 \\
BM & 0.7519$\pm$0.0258 & \textbf{0.7619$\pm$0.0124} & 0.6247$\pm$0.0878 & \textbf{0.7619$\pm$0.0124} & 0.3536$\pm$0.0325 & 0.5063$\pm$0.0007 & 0.7204$\pm$0.0387 & 0.7113$\pm$0.0425 & 0.7418$\pm$0.0110 & \textbf{0.7619$\pm$0.0124} \\
CR & 0.5215$\pm$0.0079 & \underline{0.5463$\pm$0.0119} & 0.4606$\pm$0.0078 & \underline{0.5463$\pm$0.0119} & 0.5051$\pm$0.0097 & 0.4990$\pm$0.0057 & 0.4872$\pm$0.0483 & \textbf{0.5504$\pm$0.0089} & 0.5400$\pm$0.0137 & \underline{0.5463$\pm$0.0119} \\
CD & \underline{0.5932$\pm$0.0348} & 0.5900$\pm$0.0430 & 0.3826$\pm$0.0299 & 0.5900$\pm$0.0430 & 0.5334$\pm$0.0141 & 0.5088$\pm$0.0556 & 0.5703$\pm$0.0243 & 0.5761$\pm$0.0484 & \textbf{0.6083$\pm$0.0173} & 0.5900$\pm$0.0430 \\
DB & 0.5147$\pm$0.0023 & \textbf{0.5288$\pm$0.0366} & 0.4327$\pm$0.0089 & \textbf{0.5288$\pm$0.0366} & 0.4909$\pm$0.0226 & 0.4909$\pm$0.0226 & 0.5256$\pm$0.0454 & 0.5185$\pm$0.0485 & 0.5275$\pm$0.0205 & \textbf{0.5288$\pm$0.0366} \\
EL & \textbf{0.6305$\pm$0.0015} & 0.5464$\pm$0.0279 & 0.5155$\pm$0.0392 & 0.5464$\pm$0.0279 & 0.5049$\pm$0.0015 & 0.5044$\pm$0.0017 & 0.4855$\pm$0.0377 & 0.5671$\pm$0.0432 & \underline{0.6148$\pm$0.0160} & 0.5464$\pm$0.0279 \\
EV & \underline{0.6432$\pm$0.0100} & 0.6094$\pm$0.0385 & 0.5640$\pm$0.0621 & 0.6094$\pm$0.0385 & 0.5544$\pm$0.0258 & 0.5544$\pm$0.0258 & 0.5664$\pm$0.0211 & 0.6139$\pm$0.0535 & \textbf{0.6504$\pm$0.0328} & 0.6094$\pm$0.0385 \\
HG & \textbf{0.5894$\pm$0.0069} & 0.5270$\pm$0.0213 & 0.4666$\pm$0.0281 & 0.5270$\pm$0.0213 & 0.4992$\pm$0.0145 & 0.4728$\pm$0.0333 & 0.5064$\pm$0.0176 & 0.5174$\pm$0.0415 & \underline{0.5818$\pm$0.0090} & 0.5270$\pm$0.0213 \\
HE & \textbf{0.6510$\pm$0.0222} & 0.6239$\pm$0.0276 & 0.3396$\pm$0.0264 & 0.6239$\pm$0.0276 & 0.5216$\pm$0.0358 & 0.4571$\pm$0.1091 & 0.6108$\pm$0.0226 & 0.6115$\pm$0.0590 & \underline{0.6377$\pm$0.0158} & 0.6239$\pm$0.0276 \\
HS & \underline{0.7387$\pm$0.0086} & 0.6865$\pm$0.0168 & 0.6366$\pm$0.0490 & 0.6865$\pm$0.0168 & 0.5718$\pm$0.0408 & 0.5718$\pm$0.0408 & 0.5814$\pm$0.0661 & 0.6725$\pm$0.0228 & \textbf{0.7417$\pm$0.0136} & 0.6865$\pm$0.0168 \\
JN & \underline{0.6818$\pm$0.0176} & 0.6609$\pm$0.0371 & 0.3574$\pm$0.0702 & 0.6609$\pm$0.0371 & 0.4908$\pm$0.0569 & 0.4213$\pm$0.1246 & 0.6051$\pm$0.0691 & 0.6773$\pm$0.0164 & \textbf{0.6883$\pm$0.0232} & 0.6609$\pm$0.0371 \\
LA & \textbf{0.8821$\pm$0.0363} & \underline{0.8800$\pm$0.0199} & 0.8246$\pm$0.0606 & \underline{0.8800$\pm$0.0199} & 0.5769$\pm$0.1078 & 0.5769$\pm$0.1078 & 0.6681$\pm$0.0566 & 0.8692$\pm$0.0485 & 0.8504$\pm$0.0403 & \underline{0.8800$\pm$0.0199} \\
MT & 0.9303$\pm$0.0194 & \underline{0.9423$\pm$0.0063} & 0.5985$\pm$0.4075 & \underline{0.9423$\pm$0.0063} & 0.6450$\pm$0.2563 & 0.7326$\pm$0.0261 & 0.8285$\pm$0.0853 & \textbf{0.9620$\pm$0.0108} & 0.9358$\pm$0.0135 & \underline{0.9423$\pm$0.0063} \\
MA & 0.5457$\pm$0.0007 & 0.5264$\pm$0.0153 & 0.5086$\pm$0.0347 & \textbf{0.5524$\pm$0.0273} & 0.5098$\pm$0.0150 & 0.5098$\pm$0.0150 & \underline{0.5483$\pm$0.0147} & 0.5354$\pm$0.0073 & 0.5396$\pm$0.0163 & 0.5264$\pm$0.0153 \\
MB & 0.6482$\pm$0.0196 & 0.6649$\pm$0.0406 & 0.4087$\pm$0.1227 & 0.6649$\pm$0.0406 & 0.5201$\pm$0.0278 & 0.5172$\pm$0.0492 & 0.5955$\pm$0.0767 & \underline{0.6665$\pm$0.0311} & \textbf{0.6793$\pm$0.0147} & 0.6649$\pm$0.0406 \\
NU & \textbf{0.5037$\pm$0.0058} & 0.5012$\pm$0.0040 & 0.5009$\pm$0.0061 & 0.5012$\pm$0.0040 & 0.4985$\pm$0.0042 & 0.4978$\pm$0.0049 & \underline{0.5032$\pm$0.0039} & 0.5022$\pm$0.0052 & 0.5025$\pm$0.0042 & 0.5012$\pm$0.0040 \\
NS & 0.8964$\pm$0.0237 & \underline{0.9126$\pm$0.0412} & 0.8325$\pm$0.0980 & \underline{0.9126$\pm$0.0412} & 0.6211$\pm$0.0601 & 0.6211$\pm$0.0601 & 0.8335$\pm$0.0142 & \textbf{0.9272$\pm$0.0161} & 0.9023$\pm$0.0225 & \underline{0.9126$\pm$0.0412} \\
PW & 0.7976$\pm$0.0044 & \textbf{0.8198$\pm$0.0131} & 0.5660$\pm$0.1908 & 0.7698$\pm$0.0321 & 0.6195$\pm$0.0669 & 0.6311$\pm$0.0485 & 0.6609$\pm$0.1233 & 0.6843$\pm$0.0690 & 0.7812$\pm$0.0388 & \textbf{0.8198$\pm$0.0131} \\
PL & \underline{0.7954$\pm$0.0533} & 0.7718$\pm$0.0735 & 0.6692$\pm$0.1175 & 0.7718$\pm$0.0735 & 0.6462$\pm$0.0636 & 0.6462$\pm$0.0636 & 0.6713$\pm$0.0753 & 0.7373$\pm$0.0813 & \textbf{0.8128$\pm$0.0264} & 0.7718$\pm$0.0735 \\
RS & \underline{0.6528$\pm$0.0048} & 0.6247$\pm$0.0217 & 0.3369$\pm$0.0384 & 0.6247$\pm$0.0217 & 0.5454$\pm$0.0446 & 0.5456$\pm$0.0444 & 0.5881$\pm$0.0591 & 0.5697$\pm$0.0666 & \textbf{0.6616$\pm$0.0226} & 0.6247$\pm$0.0217 \\
TC & \underline{0.5536$\pm$0.0037} & 0.5110$\pm$0.0251 & 0.5219$\pm$0.0236 & 0.5108$\pm$0.0153 & 0.5262$\pm$0.0116 & 0.5342$\pm$0.0155 & 0.5298$\pm$0.0082 & 0.5307$\pm$0.0278 & \textbf{0.5623$\pm$0.0153} & 0.5110$\pm$0.0251 \\
TD & 0.8543$\pm$0.0033 & \underline{0.8568$\pm$0.0224} & 0.2951$\pm$0.1540 & \textbf{0.8670$\pm$0.0205} & 0.7513$\pm$0.1152 & 0.3936$\pm$0.3638 & 0.7600$\pm$0.1364 & 0.8055$\pm$0.0451 & 0.8563$\pm$0.0080 & \underline{0.8568$\pm$0.0224} \\
\midrule
\# Best & 5/23 & 3/23 & 0/23 & 4/23 & 0/23 & 0/23 & 1/23 & 3/23 & 9/23 & 3/23 \\
vs RND & 18/23 & 18/23 & 6/23 & 18/23 & 2/23 & 2/23 & 11/23 & 18/23 & 18/23 & 18/23 \\
\bottomrule
\end{tabular}

			}
			\captionsetup{type=table}
      \caption{Comparison of raw balanced accuracy scores of distillation methods applied with TF-SFT on XGB classifier. Last two rows of the tables each denote the number of instances that the pipeline ranked at the top, and the number of times it outperformed random sampling. Best performance at for each dataset is marked in bold, and second-best performance is marked with underline.}
			\label{apdx:tab:results:raw_ba_xgb}
		\end{minipage}
	\end{adjustbox}
\end{table}

\begin{table}
			\centering
			{
				\scriptsize
				\begin{tabular}{lllll}
\toprule
Dataset & AG & GM & KIP & KM \\
\midrule
AD & \underline{0.6078$\pm$0.0354} & \textbf{0.7175$\pm$0.0423} & 0.5353$\pm$0.0562 & 0.5949$\pm$0.0644 \\
AE & \underline{0.5252$\pm$0.0195} & \textbf{0.5304$\pm$0.0167} & 0.5060$\pm$0.0066 & 0.5007$\pm$0.0126 \\
BM & 0.5599$\pm$0.0525 & \underline{0.5816$\pm$0.0448} & 0.5114$\pm$0.0179 & \textbf{0.5842$\pm$0.0914} \\
CR & \textbf{0.5667$\pm$0.0491} & 0.5394$\pm$0.0334 & 0.5106$\pm$0.0383 & \underline{0.5612$\pm$0.0578} \\
CD & \textbf{0.5887$\pm$0.0310} & \underline{0.5607$\pm$0.0448} & 0.5309$\pm$0.0466 & 0.5526$\pm$0.0382 \\
DB & \underline{0.5133$\pm$0.0157} & 0.5053$\pm$0.0265 & 0.5008$\pm$0.0140 & \textbf{0.5146$\pm$0.0137} \\
EL & \textbf{0.5929$\pm$0.0782} & 0.5617$\pm$0.0545 & 0.5093$\pm$0.0290 & \underline{0.5866$\pm$0.0741} \\
EV & \textbf{0.6120$\pm$0.0658} & 0.5968$\pm$0.0567 & 0.5730$\pm$0.0590 & \underline{0.6009$\pm$0.0754} \\
HG & \textbf{0.5159$\pm$0.0107} & 0.5130$\pm$0.0143 & 0.5028$\pm$0.0094 & \underline{0.5141$\pm$0.0196} \\
HE & \underline{0.5909$\pm$0.0372} & \textbf{0.5918$\pm$0.0378} & 0.5112$\pm$0.0396 & 0.5790$\pm$0.0608 \\
HS & \textbf{0.6770$\pm$0.0526} & 0.6257$\pm$0.0567 & 0.5288$\pm$0.0671 & \underline{0.6484$\pm$0.0956} \\
JN & \underline{0.6076$\pm$0.0176} & \textbf{0.6111$\pm$0.0262} & 0.5759$\pm$0.0654 & 0.5755$\pm$0.0511 \\
LA & \underline{0.8079$\pm$0.1752} & 0.8006$\pm$0.1236 & 0.7352$\pm$0.1598 & \textbf{0.8101$\pm$0.1533} \\
MT & \underline{0.8217$\pm$0.1813} & \textbf{0.9581$\pm$0.0285} & 0.8029$\pm$0.1473 & 0.8082$\pm$0.1785 \\
MA & \underline{0.5146$\pm$0.0185} & \textbf{0.5585$\pm$0.0324} & 0.4991$\pm$0.0108 & 0.5112$\pm$0.0222 \\
MB & \textbf{0.6715$\pm$0.0942} & \underline{0.6480$\pm$0.0710} & 0.5559$\pm$0.0732 & 0.6476$\pm$0.1124 \\
NU & \textbf{0.5047$\pm$0.0079} & 0.5005$\pm$0.0060 & 0.5004$\pm$0.0041 & \underline{0.5022$\pm$0.0050} \\
NS & \textbf{1.0000$\pm$0.0000} & \textbf{1.0000$\pm$0.0000} & \textbf{1.0000$\pm$0.0000} & \textbf{1.0000$\pm$0.0000} \\
PW & 0.7665$\pm$0.1429 & \textbf{0.8466$\pm$0.0613} & 0.6758$\pm$0.1216 & \underline{0.7918$\pm$0.1242} \\
PL & 0.5966$\pm$0.0441 & \underline{0.6813$\pm$0.0515} & 0.6045$\pm$0.1043 & \textbf{0.6834$\pm$0.0898} \\
RS & \textbf{0.6469$\pm$0.0546} & 0.5810$\pm$0.0373 & 0.5200$\pm$0.0350 & \underline{0.6469$\pm$0.0541} \\
TC & \textbf{0.5343$\pm$0.0295} & 0.5118$\pm$0.0357 & 0.5031$\pm$0.0228 & \underline{0.5301$\pm$0.0245} \\
TD & \textbf{0.8162$\pm$0.0100} & \underline{0.7790$\pm$0.0270} & 0.6355$\pm$0.0884 & 0.7736$\pm$0.0584 \\
\midrule
\# Best & 12/23 & 8/23 & 1/23 & 5/23 \\
vs RND & 15/23 & 16/23 & 3/23 & 15/23 \\
\bottomrule
\end{tabular}

			}
      \caption{Comparison of raw balanced accuracy scores of distillation methods applied  in the original space (no encoder) on XGB classifier. Last two rows of the tables each denote the number of instances that the pipeline ranked at the top, and the number of times it outperformed random sampling. Best performance at for each dataset is marked in bold, and second-best performance is marked with underline.}
			\label{apdx:tab:results:raw_ba_xgb_none}
\end{table}

\clearpage

\begin{table}
	\centering
	{
		\scriptsize
		\begin{tabular}{lllll}
\toprule
Dataset & AG & GM & KIP & KM \\
\midrule
AD & \underline{0.7904$\pm$0.0171} & 0.7609$\pm$0.0170 & 0.6645$\pm$0.0662 & \textbf{0.7940$\pm$0.0078} \\
AE & \textbf{0.5371$\pm$0.0022} & 0.5246$\pm$0.0244 & 0.5129$\pm$0.0130 & \underline{0.5365$\pm$0.0192} \\
BM & \textbf{0.7898$\pm$0.0052} & 0.7546$\pm$0.0317 & 0.6997$\pm$0.0556 & \underline{0.7897$\pm$0.0083} \\
CR & \underline{0.5437$\pm$0.0127} & 0.5337$\pm$0.0260 & \textbf{0.5500$\pm$0.0170} & 0.5219$\pm$0.0199 \\
CD & \underline{0.6490$\pm$0.0302} & 0.6449$\pm$0.0471 & 0.5819$\pm$0.0483 & \textbf{0.6674$\pm$0.0112} \\
DB & \textbf{0.5607$\pm$0.0019} & 0.5054$\pm$0.0474 & 0.5408$\pm$0.0335 & \underline{0.5565$\pm$0.0064} \\
EL & \underline{0.6163$\pm$0.0173} & 0.5758$\pm$0.0423 & 0.5655$\pm$0.0241 & \textbf{0.6276$\pm$0.0126} \\
EV & \textbf{0.7152$\pm$0.0017} & 0.6621$\pm$0.0319 & 0.6205$\pm$0.0448 & \underline{0.7130$\pm$0.0193} \\
HG & \textbf{0.5796$\pm$0.0338} & 0.5239$\pm$0.0128 & 0.5205$\pm$0.0106 & \underline{0.5792$\pm$0.0130} \\
HE & \textbf{0.6870$\pm$0.0061} & 0.6588$\pm$0.0174 & 0.6325$\pm$0.0600 & \underline{0.6786$\pm$0.0103} \\
HS & \textbf{0.7759$\pm$0.0119} & 0.7211$\pm$0.0279 & 0.6575$\pm$0.0831 & \underline{0.7721$\pm$0.0128} \\
JN & \textbf{0.7383$\pm$0.0050} & 0.6972$\pm$0.0111 & 0.6795$\pm$0.0142 & \underline{0.7308$\pm$0.0035} \\
LA & \textbf{0.9979$\pm$0.0000} & 0.9654$\pm$0.0255 & 0.9395$\pm$0.0760 & \underline{0.9935$\pm$0.0055} \\
MT & \textbf{0.9717$\pm$0.0002} & 0.9674$\pm$0.0040 & 0.9714$\pm$0.0065 & \underline{0.9715$\pm$0.0026} \\
MA & 0.5570$\pm$0.0096 & \underline{0.5587$\pm$0.0252} & 0.5063$\pm$0.0160 & \textbf{0.5683$\pm$0.0083} \\
MB & 0.6478$\pm$0.0156 & \underline{0.6871$\pm$0.0152} & 0.6307$\pm$0.0494 & \textbf{0.6939$\pm$0.0241} \\
NU & 0.4971$\pm$0.0083 & \underline{0.4994$\pm$0.0060} & 0.4967$\pm$0.0058 & \textbf{0.5075$\pm$0.0020} \\
NS & \textbf{0.9944$\pm$0.0063} & 0.9573$\pm$0.0063 & 0.9716$\pm$0.0095 & \underline{0.9941$\pm$0.0056} \\
PW & \underline{0.8964$\pm$0.0158} & 0.8620$\pm$0.0170 & 0.6696$\pm$0.0283 & \textbf{0.9016$\pm$0.0158} \\
PL & \underline{0.7829$\pm$0.0206} & 0.7505$\pm$0.0500 & 0.6717$\pm$0.0584 & \textbf{0.8277$\pm$0.0313} \\
RS & \underline{0.7154$\pm$0.0216} & 0.6357$\pm$0.0386 & 0.6679$\pm$0.0459 & \textbf{0.7208$\pm$0.0134} \\
TC & \underline{0.5530$\pm$0.0256} & 0.5261$\pm$0.0222 & 0.5173$\pm$0.0106 & \textbf{0.5609$\pm$0.0138} \\
TD & \underline{0.9230$\pm$0.0012} & 0.9117$\pm$0.0167 & 0.8204$\pm$0.0502 & \textbf{0.9242$\pm$0.0052} \\
\midrule
\# Best & 11/23 & 0/23 & 1/23 & 11/23 \\
vs RND & 22/23 & 21/23 & 15/23 & 22/23 \\
\bottomrule
\end{tabular}

	}
	\captionsetup{type=table}
	\caption{Comparison of raw balanced accuracy scores of distillation methods with TF-SFT and KNN downstream classifier. Best performance at for each dataset is marked in bold, and second-best performance is marked with underline.}
	\label{apdx:tab:results:raw_ba_knn}
\end{table}

\begin{table}
	\centering
	{
		\scriptsize
		\begin{tabular}{lllll}
\toprule
Dataset & AG & GM & KIP & KM \\
\midrule
AD & \textbf{0.7352$\pm$0.0212} & \underline{0.7292$\pm$0.0136} & 0.5600$\pm$0.0599 & 0.7246$\pm$0.0309 \\
AE & \textbf{0.5309$\pm$0.0252} & \underline{0.5204$\pm$0.0109} & 0.5131$\pm$0.0235 & 0.5143$\pm$0.0162 \\
BM & \underline{0.7111$\pm$0.0136} & 0.6352$\pm$0.0335 & 0.5374$\pm$0.0504 & \textbf{0.7210$\pm$0.0277} \\
CR & 0.5364$\pm$0.0178 & \underline{0.5393$\pm$0.0256} & 0.5161$\pm$0.0210 & \textbf{0.5508$\pm$0.0180} \\
CD & \textbf{0.6082$\pm$0.0252} & \underline{0.6005$\pm$0.0292} & 0.5525$\pm$0.0275 & 0.6005$\pm$0.0312 \\
DB & \underline{0.5154$\pm$0.0162} & 0.5139$\pm$0.0261 & 0.5049$\pm$0.0213 & \textbf{0.5288$\pm$0.0178} \\
EL & \textbf{0.6300$\pm$0.0314} & 0.5630$\pm$0.0346 & 0.5273$\pm$0.0463 & \underline{0.6210$\pm$0.0331} \\
EV & \underline{0.6840$\pm$0.0477} & 0.6380$\pm$0.0361 & 0.5907$\pm$0.0423 & \textbf{0.6949$\pm$0.0317} \\
HG & \textbf{0.5397$\pm$0.0181} & 0.5121$\pm$0.0124 & 0.5118$\pm$0.0064 & \underline{0.5281$\pm$0.0128} \\
HE & \underline{0.6442$\pm$0.0253} & 0.5837$\pm$0.0354 & 0.5194$\pm$0.0449 & \textbf{0.6546$\pm$0.0183} \\
HS & \underline{0.6954$\pm$0.0497} & 0.6340$\pm$0.0532 & 0.5274$\pm$0.0498 & \textbf{0.7115$\pm$0.0275} \\
JN & \underline{0.6555$\pm$0.0197} & 0.6320$\pm$0.0185 & 0.5891$\pm$0.0515 & \textbf{0.6597$\pm$0.0211} \\
LA & \textbf{0.8267$\pm$0.0424} & 0.7451$\pm$0.0585 & \underline{0.8233$\pm$0.0684} & 0.8039$\pm$0.0672 \\
MT & \underline{0.8070$\pm$0.0590} & 0.7332$\pm$0.0815 & 0.7098$\pm$0.0900 & \textbf{0.8236$\pm$0.0855} \\
MA & \textbf{0.5756$\pm$0.0098} & 0.5632$\pm$0.0237 & 0.5111$\pm$0.0340 & \underline{0.5676$\pm$0.0186} \\
MB & \underline{0.6712$\pm$0.0703} & 0.6122$\pm$0.0577 & 0.5565$\pm$0.0573 & \textbf{0.6731$\pm$0.0627} \\
NU & \textbf{0.5065$\pm$0.0033} & 0.5019$\pm$0.0050 & 0.5005$\pm$0.0054 & \underline{0.5035$\pm$0.0049} \\
NS & \underline{0.9278$\pm$0.0753} & 0.8064$\pm$0.0162 & \textbf{0.9775$\pm$0.0140} & 0.8876$\pm$0.0842 \\
PW & \textbf{0.8700$\pm$0.0175} & 0.8128$\pm$0.0311 & 0.6291$\pm$0.0598 & \underline{0.8678$\pm$0.0240} \\
PL & \underline{0.6327$\pm$0.0557} & 0.5675$\pm$0.0262 & 0.5634$\pm$0.0362 & \textbf{0.6554$\pm$0.0705} \\
RS & \textbf{0.6350$\pm$0.0324} & 0.5440$\pm$0.0214 & 0.5213$\pm$0.0200 & \underline{0.6261$\pm$0.0304} \\
TC & 0.5129$\pm$0.0285 & \underline{0.5152$\pm$0.0240} & 0.4953$\pm$0.0155 & \textbf{0.5205$\pm$0.0195} \\
TD & \underline{0.7632$\pm$0.0386} & 0.7125$\pm$0.0293 & 0.6139$\pm$0.0481 & \textbf{0.7814$\pm$0.0377} \\
\midrule
\# Best & 10/23 & 0/23 & 1/23 & 12/23 \\
vs RND & 22/23 & 18/23 & 3/23 & 22/23 \\
\bottomrule
\end{tabular}

	}
	\captionsetup{type=table}
	\caption{Comparison of raw balanced accuracy scores of distillation methods in the original space (no encoder) KNN downstream classifier. Best performance at for each dataset is marked in bold, and second-best performance is marked with underline.}
	\label{apdx:tab:results:raw_ba_knn_none}
\end{table}

\begin{table}
	\centering
	{
		\scriptsize
		\begin{tabular}{lllll}
\toprule
Dataset & AG & GM & KIP & KM \\
\midrule
AD & \underline{0.7627$\pm$0.0039} & 0.7406$\pm$0.0168 & 0.7318$\pm$0.0212 & \textbf{0.7628$\pm$0.0210} \\
AE & \underline{0.5467$\pm$0.0250} & 0.5324$\pm$0.0090 & 0.5189$\pm$0.0063 & \textbf{0.5630$\pm$0.0212} \\
BM & \textbf{0.7894$\pm$0.0319} & 0.7685$\pm$0.0210 & 0.7632$\pm$0.0312 & \underline{0.7887$\pm$0.0142} \\
CR & 0.5299$\pm$0.0242 & \underline{0.5443$\pm$0.0228} & \textbf{0.5525$\pm$0.0185} & 0.5349$\pm$0.0134 \\
CD & 0.6323$\pm$0.0845 & \underline{0.6358$\pm$0.0411} & 0.6138$\pm$0.0261 & \textbf{0.6542$\pm$0.0414} \\
DB & \underline{0.5364$\pm$0.0051} & 0.5211$\pm$0.0278 & 0.5348$\pm$0.0349 & \textbf{0.5405$\pm$0.0139} \\
EL & \underline{0.6432$\pm$0.0317} & 0.5690$\pm$0.0326 & 0.6131$\pm$0.0309 & \textbf{0.6543$\pm$0.0187} \\
EV & \textbf{0.7310$\pm$0.0019} & 0.6792$\pm$0.0421 & 0.6742$\pm$0.0347 & \underline{0.7202$\pm$0.0350} \\
HG & \textbf{0.6058$\pm$0.0126} & 0.5302$\pm$0.0068 & 0.5477$\pm$0.0254 & \underline{0.5997$\pm$0.0152} \\
HE & \underline{0.6540$\pm$0.0057} & 0.6364$\pm$0.0225 & 0.6256$\pm$0.0370 & \textbf{0.6580$\pm$0.0157} \\
HS & \textbf{0.7801$\pm$0.0029} & 0.7257$\pm$0.0336 & 0.7478$\pm$0.0149 & \underline{0.7768$\pm$0.0141} \\
JN & \textbf{0.7192$\pm$0.0036} & 0.6911$\pm$0.0130 & 0.6952$\pm$0.0315 & \underline{0.7153$\pm$0.0102} \\
LA & \textbf{0.9983$\pm$0.0010} & 0.9893$\pm$0.0192 & 0.9883$\pm$0.0238 & \underline{0.9980$\pm$0.0017} \\
MT & \underline{0.9698$\pm$0.0055} & 0.9627$\pm$0.0056 & 0.9697$\pm$0.0050 & \textbf{0.9733$\pm$0.0032} \\
MA & \underline{0.5694$\pm$0.0127} & 0.5571$\pm$0.0283 & 0.5160$\pm$0.0107 & \textbf{0.5878$\pm$0.0149} \\
MB & \textbf{0.6818$\pm$0.0092} & 0.6555$\pm$0.0568 & 0.6697$\pm$0.0217 & \underline{0.6707$\pm$0.0169} \\
NU & 0.4958$\pm$0.0047 & \underline{0.5012$\pm$0.0057} & 0.4987$\pm$0.0060 & \textbf{0.5071$\pm$0.0045} \\
NS & 0.9749$\pm$0.0153 & 0.9731$\pm$0.0179 & \underline{0.9838$\pm$0.0139} & \textbf{0.9842$\pm$0.0129} \\
PW & \underline{0.8804$\pm$0.0107} & 0.8466$\pm$0.0383 & 0.7921$\pm$0.0523 & \textbf{0.9046$\pm$0.0108} \\
PL & \textbf{0.9010$\pm$0.0198} & 0.8502$\pm$0.0175 & 0.8198$\pm$0.0426 & \underline{0.9000$\pm$0.0059} \\
RS & \underline{0.6842$\pm$0.0019} & 0.6210$\pm$0.0500 & 0.6627$\pm$0.0714 & \textbf{0.6877$\pm$0.0205} \\
TC & \textbf{0.5785$\pm$0.0231} & 0.5150$\pm$0.0307 & 0.5366$\pm$0.0251 & \underline{0.5734$\pm$0.0197} \\
TD & \underline{0.9191$\pm$0.0104} & 0.9010$\pm$0.0260 & 0.8999$\pm$0.0213 & \textbf{0.9200$\pm$0.0037} \\
\midrule
\# Best & 9/23 & 0/23 & 1/23 & 13/23 \\
vs RND & 21/23 & 19/23 & 18/23 & 23/23 \\
\bottomrule
\end{tabular}

	}
	\captionsetup{type=table}
	\caption{Comparison of raw balanced accuracy scores of distillation methods with TF-SFT and MLP downstream classifier. Best performance at for each dataset is marked in bold, and second-best performance is marked with underline.}
	\label{apdx:tab:results:raw_ba_mlp}
\end{table}

\begin{table}
	\centering
	{
		\scriptsize
		\begin{tabular}{lllll}
\toprule
Dataset & AG & GM & KIP & KM \\
\midrule
AD & 0.7183$\pm$0.0392 & \textbf{0.7576$\pm$0.0148} & 0.6756$\pm$0.0604 & \underline{0.7385$\pm$0.0276} \\
AE & \textbf{0.5743$\pm$0.0265} & 0.5444$\pm$0.0153 & 0.5267$\pm$0.0241 & \underline{0.5618$\pm$0.0410} \\
BM & \textbf{0.7406$\pm$0.0224} & 0.6573$\pm$0.0312 & 0.5776$\pm$0.0569 & \underline{0.7351$\pm$0.0311} \\
CR & \underline{0.5607$\pm$0.0216} & 0.5388$\pm$0.0272 & 0.5037$\pm$0.0440 & \textbf{0.5618$\pm$0.0277} \\
CD & \textbf{0.6146$\pm$0.0276} & 0.5920$\pm$0.0524 & 0.5706$\pm$0.0564 & \underline{0.6040$\pm$0.0332} \\
DB & 0.5168$\pm$0.0171 & \underline{0.5203$\pm$0.0207} & 0.5052$\pm$0.0281 & \textbf{0.5329$\pm$0.0203} \\
EL & \textbf{0.6713$\pm$0.0315} & 0.5904$\pm$0.0363 & 0.5568$\pm$0.0787 & \underline{0.6573$\pm$0.0347} \\
EV & \underline{0.6828$\pm$0.0270} & 0.6570$\pm$0.0289 & 0.6380$\pm$0.0651 & \textbf{0.6900$\pm$0.0255} \\
HG & \textbf{0.5463$\pm$0.0218} & 0.5184$\pm$0.0120 & 0.5190$\pm$0.0163 & \underline{0.5423$\pm$0.0221} \\
HE & \underline{0.6221$\pm$0.0210} & 0.6213$\pm$0.0397 & 0.5369$\pm$0.0574 & \textbf{0.6309$\pm$0.0227} \\
HS & \textbf{0.7514$\pm$0.0159} & 0.6746$\pm$0.0321 & 0.5970$\pm$0.0850 & \underline{0.7397$\pm$0.0396} \\
JN & \textbf{0.6352$\pm$0.0188} & 0.6328$\pm$0.0214 & 0.6209$\pm$0.0370 & \underline{0.6339$\pm$0.0137} \\
LA & \underline{0.8530$\pm$0.0389} & 0.7621$\pm$0.0419 & \textbf{0.8924$\pm$0.1000} & 0.7970$\pm$0.0443 \\
MT & \textbf{0.9008$\pm$0.0332} & 0.8068$\pm$0.0741 & \underline{0.8904$\pm$0.0269} & 0.8839$\pm$0.0584 \\
MA & \textbf{0.5710$\pm$0.0124} & 0.5591$\pm$0.0198 & 0.5294$\pm$0.0387 & \underline{0.5640$\pm$0.0151} \\
MB & \textbf{0.7411$\pm$0.0577} & 0.6768$\pm$0.0636 & 0.5995$\pm$0.0945 & \underline{0.7248$\pm$0.0656} \\
NU & \textbf{0.5076$\pm$0.0025} & 0.5009$\pm$0.0057 & 0.5004$\pm$0.0028 & \underline{0.5063$\pm$0.0059} \\
NS & \underline{0.9799$\pm$0.0208} & 0.8159$\pm$0.0102 & \textbf{0.9967$\pm$0.0038} & 0.9006$\pm$0.0746 \\
PW & \textbf{0.9018$\pm$0.0178} & 0.8248$\pm$0.0253 & 0.8084$\pm$0.0546 & \underline{0.8775$\pm$0.0323} \\
PL & \underline{0.7934$\pm$0.0946} & 0.6883$\pm$0.0491 & 0.7319$\pm$0.0239 & \textbf{0.7961$\pm$0.0719} \\
RS & \underline{0.6304$\pm$0.0134} & 0.5567$\pm$0.0191 & 0.5386$\pm$0.0350 & \textbf{0.6305$\pm$0.0228} \\
TC & \textbf{0.5404$\pm$0.0253} & \underline{0.5177$\pm$0.0141} & 0.5016$\pm$0.0358 & 0.5154$\pm$0.0328 \\
TD & \textbf{0.7924$\pm$0.0154} & 0.7164$\pm$0.0507 & 0.6930$\pm$0.0413 & \underline{0.7751$\pm$0.0265} \\
\midrule
\# Best & 14/23 & 1/23 & 2/23 & 6/23 \\
vs RND & 22/23 & 19/23 & 6/23 & 21/23 \\
\bottomrule
\end{tabular}

	}
	\captionsetup{type=table}
	\caption{Comparison of raw balanced accuracy scores of distillation methods with in the original space (no encoder) MLP downstream classifier. Best performance at for each dataset is marked in bold, and second-best performance is marked with underline.}
	\label{apdx:tab:results:raw_ba_mlp_none}
\end{table}

\end{document}